\documentclass[3p]{elsarticle}

\usepackage{
    lineno,
    hyperref,
    amsmath,
    amsthm,
    bm,
    amssymb,
    extarrows,
    mathrsfs,
    tcolorbox,
    cases,
    verbatim,
    wrapfig,
    ascmac,
    makeidx,
    appendix,
    algorithmic,
    algorithm
}
\usepackage{midpage}
\usepackage{color}
\usepackage{url}
\usepackage{mathtools}
\usepackage{booktabs}
\usepackage{multirow}
\usepackage{diagbox}
\usepackage[hang,small,bf]{caption}
\usepackage[subrefformat=parens]{subcaption}
\global\long\def\T#1{#1^{\top}}
\newcommand{\E}{\mathrm{E}}

\newcommand{\tr}{\mathrm{tr}}
\newcommand{\no}{\nonumber}
\newcommand{\B}{\boldsymbol}
\newcommand{\argmax}{\operatornamewithlimits{argmax}}

\modulolinenumbers[5]

\journal{Pattern Recognition}

\begin{document}

\begin{frontmatter}

\title{
    Time Series Clustering with an EM algorithm for Mixtures \\ of Linear Gaussian State Space Models
}

\author[1]{Ryohei Umatani\corref{cor1}}
\ead{umataniryohei@gmail.com}
\author[2]{Takashi Imai}
\author[1]{Kaoru Kawamoto}
\author[4]{Shutaro Kunimasa}

\address[1]{Department of Data Science, Shiga University, Japan}
\address[2]{Data Science and AI Innovation Research Promotion Center, Shiga University, Japan}
\address[4]{Business Analysis Center, Osaka Gas Co., Ltd., Japan}

\cortext[cor1]{Corresponding author}

\begin{abstract}
In this paper, we consider the task of clustering a set of individual time series while modeling each cluster, that is, model-based time series clustering. 
The task requires a parametric model with sufficient flexibility to describe the dynamics in various time series. 
To address this problem, we propose a novel model-based time series clustering method with mixtures of linear Gaussian state space models, which have high flexibility. 
The proposed method uses a new expectation-maximization algorithm for the mixture model to estimate the model parameters, and determines the number of clusters using the Bayesian information criterion. 
Experiments on a simulated dataset demonstrate the effectiveness of the method in clustering, parameter estimation, and model selection. 
The method is applied to real datasets commonly used to evaluate time series clustering methods.
Results showed that the proposed method produces clustering results that are as accurate or more accurate than those obtained using previous methods.
\end{abstract}

\begin{keyword}
    time series clustering \sep model-based clustering \sep state space model \sep EM algorithm \sep mixture model
\end{keyword}

\end{frontmatter}

% \linenumbers

\section{Introduction}
Analysis of time series data is a major issue in a wide range of fields, including science, engineering, business, finance, economics, health care, and government \cite{Sangeeta2012}. 
One of the more important methods of time series analysis in such fields is time series clustering, which divides a given set of time series into groups with different time-course patterns. 
Time series clustering has, for example, led to the detection of brain activities \cite{Axel2002} and the discovery of human behavior patterns \cite{Kurbalija2012}, energy consumption patterns \cite{en6020579}, and personal income patterns \cite{Kumar2002}.

There are three principal approaches to time series clustering: the shape-based approach, the feature-based approach, and the model-based approach \cite{WARRENLIAO20051857,AGHABOZORGI201516}. 
Among the three, the model-based approach has two advantages over the others. 
First, the model-based approach often provides more accurate results than the others if the adopted model can describe time-course patterns in the time series properly \cite{XIONG20041675,4782736}. 
Second, it allows us to predict the future using the estimated model. 
Thus, the model-based approach is particularly effective if an appropriate predictive model is adopted.

For accurate clustering and forecasting, we expect the adopted model to have sufficient flexibility to describe the dynamics in the time series. 
Typical examples in previous studies are an autoregressive (AR) model \cite{XIONG20041675,4782736} and a hidden Markov model (HMM) \cite{Cen2000}. 
However, the former has the limitation that it cannot adequately describe non-stationary time series. 
In addition, it is difficult to understand the underlying dynamics through the estimated AR model. 
Although the latter has latent variables to allow us to construct a rich class of models, its application is limited to cases in which the latent variables are discrete.

\begin{figure}[tbp]
    \centering{\includegraphics[width=140mm]{"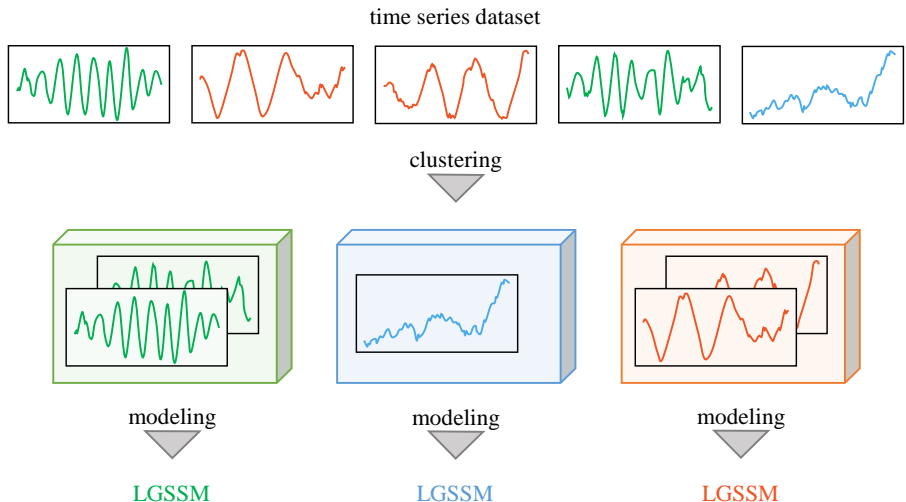"}}
    \caption{Two tasks of model-based time series clustering with MLGSSMs. 
    The first task is the clustering of a given time series dataset. The second is the modeling of each cluster with an LGSSM\@.}
    \label{fig1}
\end{figure}

In this paper, we propose a novel time series clustering method based on linear Gaussian state space models (LGSSMs) \cite{Kitagawa1996}. 
An LGSSM has three major advantages.
The first is the high flexibility of the model. 
An LGSSM consists of a state equation and an observation equation. 
While the state equation describes the signal generation mechanism, the observation equation describes the signal observation mechanism. 
These two equations enable the LGSSM to describe a non-stationary time series. 
This is the same situation as in an HMM, but in contrast to an HMM, an LGSSM uses continuous latent variables. 
The second advantage is the high interpretability of the model. 
An LGSSM directly describes the dependence among the explanatory factors of the system's dynamics. 
The third advantage is that an LGSSM can interpolate missing values in time series. 
Sometimes accidents such as machine failures and system errors can cause portions of time series data to be lost. 
To address the problem of missing values, we often remove the set of time series with missing values or fill in their values with certain statistical values. 
Such operations can result in a lack of available time series or a bias in the time series being used. 
With an LGSSM, however, we can use the Kalman filter \cite{anderson1979} to interpolate the missing values exactly.

Our method is based on finite mixtures of LGSSMs (MLGSSMs), a parametric family of mixture models \cite{mclachlan1988mixture} whose components correspond to LGSSMs. 
As shown in Fig.~\ref*{fig1}, this model not only classifies a time series dataset into a finite number of groups, but it also associates each group with an LGSSM\@.

The main goal of this paper is to derive an expectation-maximization (EM) algorithm \cite{Dempster1977,Richard1984} for MLGSSMs. 
EM algorithms are an approach to indirect maximum likelihood estimation. 
Because in this case we cannot directly compute the likelihood of LGSSMs, the lower bound of the likelihood is increased iteratively instead. 
This iterative operation is guaranteed to increase the likelihood.

The remainder of the paper is organized as follows: We summarize related work in Sec.~2, and review the well-known EM algorithm for LGSSMs in Sec.~3. 
The EM algorithm for MLGSSMs is presented in Sec.~4. 
We demonstrate the validity of this algorithm via experiments on simulated and real datasets in Sec.~5. 
In Sec.~6, we conclude the paper and discuss future work.

\section{Related work}
In many traditional time series clustering methods, the given time series are converted into low-dimensional feature vectors using a feature extraction method, such as those proposed in Refs.~\cite{Agrawal1993,795160,Gavrilov}, to reduce computational costs for calculating the similarity between data. 
In contrast, model-based methods reduce the costs by using the similarity between the estimated models as the similarity measure. 
For example, the (dis)similarity between the parameters for a Markov chain model \cite{Sebastiani1999,Ramoni2000MultivariateCB}, of an HMM \cite{Panuccio2002,Licen1999,Li2002ApplyingTH}, and of an autoregressive integrated moving average model \cite{Piccolo1990,989529}, have been used.
More specifically, the Kullback--Leibler distance \cite{Sebastiani1999,Ramoni2000MultivariateCB,Panuccio2002}, the sequence-to-model likelihood distance \cite{Licen1999,Li2002ApplyingTH,18626}, the Euclidean distance \cite{Piccolo1990}, and the Euclidean distance of the linear predictive coding cepstrum \cite{989529} between each pair of model parameters were calculated.

While the above-mentioned model-based methods follow the traditional distance-based approach in terms of using the distance measures, fully model-based methods have also been proposed \cite{XIONG20041675,4782736,Cen2000}, which calculate the cluster assignment probabilities for each time series. 
Specifically, Refs.~\cite{XIONG20041675,4782736} and \cite{Cen2000} use mixtures of AR models (MARs) and of HMMs, respectively, to implement the probabilistic representation. 
Furthermore, Refs.~\cite{XIONG20041675,Cen2000} and \cite{4782736} present the EM \cite{Dempster1977,Richard1984} and variational Bayes (VB) \cite{Bishop2006,Beal} algorithms for the mixture models, respectively.
The fully model-based approach has been extended to more specialized models in some recent studies.
For example, a time series clustering method with mixtures of integer-valued autoregressive processes has been proposed \cite{Tyler2021}.

The novel clustering method proposed in the present study is a fully model-based method.
As described above, the EM and VB algorithms are generally used for model estimations. 
However, these algorithms have not yet been applied to MLGSSMs. 
Accordingly, in this paper, we propose an EM algorithm for MLGSSMs.

The concept of time series clustering based on state space models can be found in previous studies, but their goals are different from ours.
In some cases, it was assumed that clusters are only different in their initial state mean and process noise covariance \cite{LinBachelor,pmlr-v89-lin19b}.
Other reports considered the task of clustering components \cite{4383735} and time points \cite{NIPS2016_7d6044e9} for a single (multivariate) time series.

There are generally two algorithms to obtain the parameters for LGSSMs: an EM algorithm \cite{Ghahramani1996,Shumway1982} and a VB algorithm \cite{Beal}. 
In the next section, we review the EM algorithm for LGSSMs, which is then extended to MLGSSMs in Sec.~4.

\section{EM algorithm for single LGSSMs}
An LGSSM, also known as a linear time-invariant dynamical system, consists of the observed variables $\B Y = \bigl\{ \B y[t] \bigr\}_{t=1}^T$ and the state (latent) variables $\B X = \bigl\{\B x[t]\bigr\}_{t=1}^{T}$. 
Both the observed variables and the state variables have multivariate Gaussian distributions. 
Thus, an LGSSM takes the form
\begin{subequations}
    \renewcommand{\theequation}{
    \theparentequation-\alph{equation} 
    }
    \label{Eq:1}
    \begin{align}
        \B x[t] &= \B A \B x[t-1] + \B w[t], \\
        \B y[t] &=\B C \B x[t] + \B v[t], \\
        \B x[1] &= \B \mu + \B u ,
    \end{align}
\end{subequations}
where the noise terms $\B w$, $\B v$, and $\B u$ are zero-mean normally distributed random variables with covariance matrices $\B \Gamma$, $\B \Sigma$, and $\B P$, respectively. 
Suppose $d_y$ and $d_x$ are the dimensions of $\B y$ and $\B x$, respectively. Let $\B \theta=\bigl\{\B A,\,\B \Gamma,\,\B C,\,\B \Sigma,\,\B \mu,\,\B P\bigr\}$ be the set of parameters for the model, where $\B A$, $\B \Gamma$, and $\B P$ are  ($d_x \times d_x$) matrices, $\B C$ is a ($d_y \times d_x$) matrix, $\B \Sigma$ is a ($d_y \times d_y$) matrix, and $\B \mu$ is a $d_x$-dimensional vector. 
The parameters $\B \theta$ are estimated by the EM algorithm.

The EM algorithm for LGSSMs (\ref*{Eq:1}) is derived as follows \cite{Ghahramani1996,Shumway1982}. 
The complete-data log-likelihood function (CDLL) of an LGSSM given by
\begin{align}
    \log\,{p\bigl(\B Y,\,\B X \mid \B \theta \bigr)} 
    = \log\,{p\bigl(\B x[1] \mid \B \mu,\,\B P\bigr)} + \log\sum_{t=2}^T p\bigl(\B x[t]\mid \B x[t-1],\,\B A,\,\B \Gamma\bigr) + \log\sum_{t=1}^T p\bigl(\B y[t]\mid \B x[t],\,\B C,\,\B \Sigma\bigr) .
\end{align} 
Let $Q$ denote the expectation of the complete-data log-likelihood function (ECDLL) with the posterior distribution $p\bigl(\B X \mid \B Y,\,\B \theta(s)\bigr)$
\begin{align}
    \label{Eq:2}
    Q\bigl(\B \theta,\,\B \theta(s)\bigr) 
    = \E_{\B X \mid \B Y,\,\B \theta(s)} \Bigl[ \, \log\,{p\bigl(\B Y,\,\B X \mid \B \theta\bigr)} \, \Bigr] ,
\end{align}
where $\B \theta(s)$ means the current estimated parameters of the LGSSM\@.
Because the observed variables and state variables have Gaussian distributions, the LGSSM (\ref*{Eq:1}) is expressed in an equivalent form
\begin{subequations}
    \renewcommand{\theequation}{
    \theparentequation-\alph{equation} 
    }
    \begin{align}
        p\bigl(\B x[t]\mid \B x[t-1] ,\,\B A,\,\B \Gamma \bigr) &= \mathcal{N}\bigl(\B x[t]\mid \B A \B x[t-1],\,\B \Gamma \bigr) , \\
        p\bigl(\B y[t]\mid \B x[t],\,\B C ,\,\B \Sigma \bigr) &= \mathcal{N}\bigl(\B y[t]\mid \B C \B x[t],\,\B \Sigma \bigr) , \\
        p\bigl(\B x[1] \mid \B \mu ,\,\B P \bigr) &= \mathcal{N}\bigl(\B x[1] \mid \B \mu ,\,\B P \bigr) .
    \end{align}
\end{subequations}
Hence, the ECDLL (\ref*{Eq:2}) becomes
\begin{align}
    \label{Eq:5}
    Q\bigl(\B \theta,\,\B \theta(s)\bigr) 
    =& -\dfrac{1}{2} \log{\bigl|\B P\bigr|} - \E \Bigl[ \,\frac{1}{2} \T{\bigl(\B x[1] - \B \mu\bigr)} \B P^{-1} \bigl(\B x[1] - \B \mu\bigr) \,\Bigr] \no \\
    & -\frac{T-1}{2}\log{\bigl|\B \Gamma\bigr|} - \E \Bigl[ \,\frac{1}{2} \sum_{t=2}^{T} \T{\bigl(\B x[t] - \B A \B x[t-1]\bigr)} {\B \Gamma}^{-1} \bigl(\B x[t] - \B A \B x[t-1]\bigr) \,\Bigr] \no \\
    & -\frac{T}{2} \log{\bigl|\B \Sigma\bigr|} - \E \Bigl[ \,\frac{1}{2} \sum_{t=1}^{T} \T{\bigl(\B y[t] - \B C \B x[t]\bigr)} {\B \Sigma}^{-1} \bigl(\B y[t] - \B C \B x[t]\bigr) \,\Bigr] \no \\
    =&\, \dfrac{1}{2}\log{\bigl|{\B P}^{-1}\bigr|} - \frac{1}{2} \Bigl\{ \,\tr{ \Bigl({\B P}^{-1} \E \Bigl[ \,\B x[1]\B x\T{[1]} \,\Bigr] \Bigr)} - \tr{ \Bigl({\B P}^{-1}\B \mu \E \Bigl[ \,\B x\T{[1]} \,\Bigr] \Bigr)} \no \\
    & - \tr{ \Bigl({\B P}^{-1}\E \Bigl[ \,\B x[1]\,\Bigr] \T{\B \mu} \Bigr)} + \tr{ \Bigl({\B P}^{-1}\B \mu \T{\B \mu}\Bigr)} \,\Bigr\} \no \\
    & - \dfrac{T-1}{2}\log{\bigl|\B \Gamma\bigr|} - \frac{1}{2} \sum_{t=2}^{T} \Bigl\{ \,\tr{\Bigl( {\B \Gamma}^{-1} \E \Bigl[ \,\B x[t] \B x\T{[t]} \,\Bigr] \Bigr)} - \tr{\Bigl( {\B \Gamma}^{-1} \B A \E \Bigl[ \, \B x[t-1]\B x\T{[t]} \,\Bigr] \Bigr)} \no \\
    & - \tr{\Bigl( {\B \Gamma}^{-1} \E \Bigl[ \,\B x[t]\B x\T{[t-1]}\,\Bigr]\T{\B A} \Bigr)} + \tr{\Bigl( {\B \Gamma}^{-1} \B A \E \Bigl[ \,\B x[t-1]\B x\T{[t-1]}\,\Bigr] \T{\B A} \Bigr)} \,\Bigr\} \no \\
    & - \dfrac{T}{2}\log{\bigl|\B \Sigma\bigr|} - \frac{1}{2} \sum_{t=1}^{T} \Bigl\{ \,\tr{\Bigl( {\B \Sigma}^{-1} \B y[t] \B y\T{[t]} \Bigr)} - \tr{\Bigl( {\B \Sigma}^{-1} \B y[t] \E \Bigl[ \,\B x\T{[t]}\,\Bigr] \T{\B C} \Bigr)} \no \\
    & - \tr{\Bigl( {\B \Sigma}^{-1} \B C \E \Bigl[ \,\B x[t]\,\Bigr] \B y\T{[t]} \Bigr)} + \tr{\Bigl( {\B \Sigma}^{-1} \B C \E \Bigl[ \,\B x[t]\B x\T{[t]}\,\Bigr] \T{\B C} \Bigr)} \,\Bigr\} . 
\end{align}

In the E-step, we compute the following expectations to compute the ECDLL (\ref*{Eq:5}):
\begin{subequations}
    \renewcommand{\theequation}{
    \theparentequation-\alph{equation}  
    }
    \label{Eq:6}
    \begin{align}
        \E \Bigl[ \,\B x[t]\, \Bigr] &= \B \mu[t \mid T] , \\
        \E \Bigl[ \,\B x[t]\T{\B x[t-1]} \,\Bigr] &= \B V[t \mid T] \T{\B J[t-1]} + \B \mu[t \mid T] \T{\B \mu[t-1 \mid T]} , \\
        \E \Bigl[ \,\B x[t]\T{\B x[t]} \,\Bigr] &= \B V[t \mid T] + \B \mu[t \mid T] \T{\B \mu[t \mid T]} ,
    \end{align}
\end{subequations}
where $\B \mu$, $\B V$, and $\B J$ are defined as
\begin{subequations}
    \renewcommand{\theequation}{
    \theparentequation-\alph{equation}  
    }
    \begin{align}
        \B \mu[t \mid j] & \equiv \E_{\B x[t] \mid \{ \B y[t] \}_{t=1}^j ,\, \B \theta(s)} \Bigl[ \,\B x[t] \,\Bigr], \\
        \B V[t \mid j] & \equiv \E_{\B x[t] \mid \{ \B y[t] \}_{t=1}^j ,\, \B \theta(s)} \Bigl[ \,\bigl(\B x[t] - \B \mu[t \mid j]\bigr)\T{\bigl(\B x[t] - \B \mu[t \mid j]\bigr)} \,\Bigr], \\
        \B J[t] & \equiv \B V[t \mid t] \T{\B A} {\B V[t+1 \mid t]}^{-1} ,
    \end{align}
\end{subequations}
and are obtained from the Kalman filter \cite{anderson1979}, as described in Algorithm \ref*{kalmanfilter}.
 
In the M-step, we attempt to maximize $Q\bigl(\B \theta,\,\B \theta(s)\bigr)$ by solving
\begin{align}
    \frac{\partial}{\partial \B \theta}  Q\bigl(\B \theta,\,\B \theta(s)\bigr) = 0 .
\end{align}
This provides the following new values of the LGSSM parameters:
\begin{subequations}
    \renewcommand{\theequation}{
    \theparentequation-\alph{equation}  
    }
    \begin{align}
        \B \mu (s+1) =&\, \E \Bigl[ \,\B x[1] \,\Bigr], \\
        \B P (s+1) =&\, \E \Bigl[ \,\B x[1] \T{\B x[1]} \,\Bigr] - \E \Bigl[ \,\B x[1] \,\Bigr] \T{\E \Bigl[ \,\B x[1] \,\Bigr]} , \\
        \B A (s+1) =&\, \Bigl( \,\sum_{t=2}^{T} \E \Bigl[ \,\B x[t]\T{\B x[t-1]} \,\Bigr] \,\Bigr) \Bigl( \,\sum_{t=2}^{T} \E \Bigl[ \,\B x[t-1] \T{\B x[t-1]} \,\Bigr] \,\Bigr)^{-1} , \\
        \B \Gamma (s+1) =&\, \frac{1}{T-1} \sum_{t=2}^{T} \Bigl \{ \E \Bigl[ \,\B x[t]\T{\B x[t]} \,\Bigr] - \B A (s+1) \T{\E \Bigl[ \,\B x[t]\T{\B x[t-1]} \,\Bigr]} \no \\ 
        & - \E \Bigl[ \,\B x[t]\T{\B x[t-1]} \,\Bigr]\T{\B A (s+1)} + \B A (s+1)\E \Bigl[ \,\B x[t-1] \T{\B x[t-1]} \,\Bigr] \T{\B A (s+1)} \Bigl \} , \\
        \B C (s+1) =&\, \Bigl( \,\sum_{t=1}^{T} \B y[t]\T{\E \Bigl[ \,\B x[t]\, \Bigr]} \,\Bigr) \Bigl( \,\sum_{t=1}^{T} \E \Bigl[ \,\B x[t] \T{\B x[t]} \,\Bigr] \,\Bigr)^{-1} , \\
        \B \Sigma (s+1) =&\, \frac{1}{T} \sum_{t=1}^{T} \Bigl \{ \B y[t]\T{\B y[t]} - \B C (s+1) \E \Bigl[ \,\B x[t]\, \Bigr] \T{\B y[t]} \no \\
        & - \B y[t]\T{\E \Bigl[ \,\B x[t]\, \Bigr]} \T{\B C (s+1)} + \B C (s+1) \E \Bigl[ \,\B x[t] \T{\B x[t]} \,\Bigr] \T{\B C (s+1)} \Bigl \} .
    \end{align}
\end{subequations}
Alternately repeating the E-step and M-step until a convergence criterion is satisfied, we can obtain the optimal parameters $\B \theta^{*}$.

For the given $\B Y$, the representation of the LGSSM is not unique. 
In fact, we can obtain another representation using a nonsingular matrix, $\B T$, by replacing $\B x[t]$, $\B A$, $\B \Gamma$, $\B C$, $\B \mu$, and $\B P$ with
\begin{align*}
    \B x[t]^{'}=\B T \B x[t], \quad \B A^{'} = \B T \B A \B T^{-1}, \quad 
    \B \Gamma^{'} = \B T \B \Gamma \B T^{-1}, \quad \B C^{'} = \B C \B T^{-1}, \quad
    \B \mu^{'} = \B T \B \mu, \quad \B P^{'} = \B T \B P \B T^{-1},
\end{align*}
respectively. 
Thus, we need to place constraints on the parameter values to make the model identifiable \cite{James1994}.

\begin{figure}[H]
    \begin{algorithm}[H]
        \caption{Kalman Filter and Smoother}
        \label{kalmanfilter}
        \begin{algorithmic}[1]  
            \REQUIRE $\B \theta=\bigl\{\B A,\,\B \Gamma,\,\B C,\,\B \Sigma,\,\B \mu,\,\B P\bigr\}$
            \STATE $\B \mu[1 \mid 0] = \B \mu$, $\B V[1 \mid 0] = \B V$
            \STATE // Kalman Filter
            \FOR{$t=1,\;\dotsc\,,\;T$}
            \STATE $\B K[t] \leftarrow \B V[t \mid t-1] \T{\B C}\bigl(\B C \B V[t \mid t-1] \T{\B C} + \B \Sigma\bigr)^{-1}$
            \STATE $\B \mu[t \mid t] \leftarrow \B \mu[t \mid t-1] + \B K[t] \bigl(\B y[t] - \B C \B \mu[t \mid t-1]\bigr)$
            \STATE $\B V[t \mid t] \leftarrow \bigl(\B I - \B K[t] \B C\bigr)\B V[t \mid t-1]$
            \IF{$t<T$}
            \STATE $\B \mu[t+1 \mid t] \leftarrow \B A \B \mu[t \mid t]$
            \STATE $\B V[t+1 \mid t] \leftarrow \B A \B V[t \mid t]\T{\B A} + \B \Gamma$
            \ENDIF
            \ENDFOR 
            \STATE // Smoother
            \FOR{$t=T-1,\;\dotsc\,,\;1$}
            \STATE $\B J[t] \leftarrow \B V[t \mid t] \T{\B A}{\B V[t+1 \mid t]}^{-1}$
            \STATE $\B \mu[t \mid T] \leftarrow \B \mu[t \mid t] + \B J[t] \bigl(\B \mu[t+1 \mid T] - \B A \B \mu[t \mid t]\bigr)$
            \STATE $\B V[t \mid T] \leftarrow \B V[t \mid t] + \B J[t] \bigl(\B V[t+1 \mid T] - \B V[t+1 \mid t]\bigr)\T{\B J[t]}$
            \ENDFOR 
        \end{algorithmic}
    \end{algorithm}
\end{figure}

\section{EM algorithm for MLGSSMs}

\subsection{Definition of MLGSSMs}

Let $\B D_Y = \bigl\{\B Y_i\bigr\}_{i = 1}^{N}$ denote a dataset comprising $N$ time series $\B Y_i = \bigl\{\B y_i [t]\bigr\}_{t = 1}^{T}$ with length $T$. 
We assume that these time series are generated from $M$ different LGSSMs
\begin{subequations}
    \renewcommand{\theequation}{
    \theparentequation-\alph{equation}  
    }
    \label{Eq:10}
    \begin{align}
        \B x[t] &= \B A^{(k)} \B x[t-1] + \B w[t] , \\
        \B y[t] &=\B C^{(k)} \B x[t] + \B v[t] , \\
        \B x[1] &= \B \mu^{(k)} + \B u ,
    \end{align}
\end{subequations}
where
\begin{subequations}
    \renewcommand{\theequation}{
    \theparentequation-\alph{equation}  
    }
    \begin{align}
        \B w[t] &\sim \mathcal{N}\bigl(\B 0,\,\B \Gamma^{(k)}\bigr) , \\
        \B v[t] &\sim \mathcal{N}\bigl(\B 0,\,\B \Sigma^{(k)}\bigr) , \\
        \B u &\sim \mathcal{N}\bigl(\B 0,\,\B P^{(k)}\bigr) ,
    \end{align}
\end{subequations}
which correspond to the $M$ clusters denoted as $\omega^{(1)},\;\dotsc\,,\;\omega^{(M)}$. 
This assumption means that there exists a time series, $\B X_i = \bigl\{\B x_{i}[t]\bigr\}_{t = 1}^{T}$, of the state variables behind each time series $\B Y_i$. 
We write the whole set of these latent time series as $\B D_X$; i.e., $\B D_X = \bigl\{\B X_i\bigr\}_{i = 1}^{N}$. 
Let $p^{(k)} = p(\omega^{(k)})$, and therefore
\begin{eqnarray}
    \label{Eq:12}
    \sum_{k=1}^M p^{(k)} =1 .
\end{eqnarray}
Let $\B \theta^{(k)}=\bigl\{ \B A^{(k)},\,\B \Gamma^{(k)},\,\B C^{(k)},\,\B \Sigma^{(k)},\,\B \mu^{(k)},\,\B P^{(k)} \bigr\}$ be the LGSSM parameters of cluster $k$, and define the LGSSM mixture distribution as
\begin{eqnarray}
p\bigl(\B Y,\,\B X \mid \B \Theta\bigr) = \sum_{k=1}^{M} p\bigl(\B Y,\,\B X \mid \omega^{(k)},\,\B \theta^{(k)}\bigr)p^{(k)} ,
\end{eqnarray}
where $\B \Theta = \{(\B \theta^{(1)},\,p^{(1)}),\;\dotsc\,,\;(\B \theta^{(M)},\,p^{(M)})\}$.

\subsection{E-step of EM algorithm}

We now introduce new latent variables $\B Z = \bigl\{ \B z_1,\;\dotsc\,,\;\B z_N \bigr\}$ in which $\B z_i$ indicates the cluster of $\B Y_i$. 
For these variables, the following equality holds:
\begin{align}
    p\bigl(\B Y_i,\,\B X_i,\,\B z_i \mid \omega^{(k)},\,\B \theta^{(k)}\bigr) =
\begin{cases}
    p\bigl(\B Y_i,\,\B X_i \mid \omega^{(k)},\,\B \theta^{(k)}\bigr) & \text{if}\,\,\,k=\B z_i , \\
    0 & \text{otherwise} .
\end{cases}
\end{align}
Using this equality, we can express the log-likelihood of the complete-data $(\B D_Y,\,\B D_X,\,\B Z)$ as
\begin{align}
    \label{Eq:15}
    \log\,p\bigl(\B D_Y,\,\B D_X,\,\B Z \mid \B \Theta\bigr)
    &= \sum_{i=1}^{N} \log\,{p\bigl(\B Y_i,\,\B X_i,\,\B z_i \mid \B \Theta\bigr)} \no \\
    &= \sum_{i=1}^{N} \log{\sum_{k=1}^{M} p\bigl(\B Y_i,\,\B X_i,\,\B z_i \mid \omega^{(k)},\B \theta^{(k)}\bigr)p^{(k)}} \no \\
    &= \sum_{i=1}^{N} \log\,{p\bigl(\B Y_i,\,\B X_i \mid \omega^{(\B z_i)},\,\B \theta^{(\B z_i)}\bigr)p^{(\B z_i)}} \no \\
    &= \sum_{i=1}^{N} \log\,{p\bigl(\B Y_i,\,\B X_i \mid \omega^{(\B z_i)},\,\B \theta^{(\B z_i)}\bigr)} + \sum_{i=1}^{N} \log\,{p^{(\B z_i)}} .
\end{align}
In the E-step, we calculate the following ECDLL, $\mathcal{Q}$, of the MLGSSM using the current estimated parameters $\B \Theta(s) = \bigl\{\bigl(\B \theta^{(1)}(s),\,p^{(1)}(s)\bigr),\;\dotsc\,,\;\bigl(\B \theta^{(M)}(s),\,p^{(M)}(s)\bigr)\bigr\}$:
\begin{align}
    \mathcal{Q}\bigl(\B \Theta \mid \B \Theta(s)\bigr)
    &= \E_{\B Z,\,\B D_X \mid \B D_Y,\,\B \Theta(s)} \Bigl[ \, \log\,p\bigl(\B D_Y,\,\B D_X,\,\B Z \mid \B \Theta\bigr) \,\Bigr] \no \\
    &= \E_{\B Z \mid \B D_Y,\, \B \Theta(s)} \Bigl[ \,\E_{\B D_X \mid \B Z,\,\B D_Y,\,\B \Theta(s)} \Bigl[ \, \log\,p\bigl(\B D_Y,\,\B D_X,\,\B Z \mid \B \Theta\bigr) \,\Bigr] \,\Bigr] \no \\
    \label{Eq:16}
    &= \sum_{i=1}^{N} \sum_{k = 1}^{M} p\bigl(\omega^{(k)} \mid \B Y_i,\,\B \Theta(s)\bigr) \Bigl( Q_{i}^{(k)}\bigl(\B \theta^{(k)},\,\B \theta^{(k)}(s)\bigr) + \log\,{p^{(k)}} \Bigr) ,
\end{align}
where
\begin{align}
    \label{Eq:17}
    Q_{i}^{(k)}\bigl(\B \theta^{(k)},\,\B \theta^{(k)}(s)\bigr)
    =&\, \E_{\B X_i \mid \B Y_i,\,\B \theta^{(k)}(s)} \Bigl[ \,\log\,{p\bigl(\B Y_i,\,\B X_i \mid \B \theta^{(k)}\bigr)} \,\Bigr] \no \\
    =&\, \E_{\B X_i \mid \B Y_i ,\,\B \theta^{(k)}(s)} \Bigl[ \,\log\,{p\bigl(\B x_{i}[1] \mid \B \theta^{(k)}\bigr)} \no \\
    & + \sum_{t=2}^{T} \log\,{p\bigl(\B x_{i}[t] \mid \B x_{i}[t-1],\,\B \theta^{(k)}\bigr)} \no \\
    & + \sum_{t=1}^{T} \log\,{p\bigl(\B y_{i}[t] \mid \B x_{i}[t],\,\B \theta^{(k)}\bigr)} \,\Bigr] .
\end{align}
See Appendix A for the detailed derivation of the ECDLL (\ref*{Eq:16}). 
Equation (\ref*{Eq:17}) corresponds to the ECDLL (\ref*{Eq:2}) of the LGSSM\@.
Hence, we need to compute the expectations (\ref*{Eq:6}) using Algorithm \ref*{kalmanfilter} for each value pair of $i$ and $k$.
The posterior probability $p\bigl(\omega^{(k)} \mid \B Y_i,\,\B \Theta(s)\bigr)$ in $\mathcal{Q}\bigl(\B \Theta \mid \B \Theta(s)\bigr)$ is calculated using Bayes' theorem as follows:
\begin{align}
    \label{Eq:18}
    p\bigl(\omega^{(k)} \mid \B Y_i,\,\B \Theta(s)\bigr)
    = \dfrac{p\bigl(\B Y_i \mid \omega^{(k)},\,\B \theta^{(k)}(s)\bigr)p^{(k)}(s)}{\sum\limits_{u=1}^{M} p\bigl(\B Y_i \mid \omega^{(u)},\,\B \theta^{(u)}(s)\bigr)p^{(u)}(s)} ,
\end{align}
where $p\bigl(\B Y_i \mid \omega^{(k)},\,\B \theta^{(k)}(s)\bigr)$ can be calculated as
\begin{align}
    \label{Eq:19}
    p\bigl(\B Y_i \mid \omega^{(k)},\,\B \theta^{(k)}(s)\bigr) 
    = \prod\limits_{t=1}^{T} \mathcal{N}\bigl(\B y_{i}[t] \mid \B C^{(k)}(s) \B \mu_{i}^{(k)}[t \mid t-1],\ \B C^{(k)}(s) \B V_{i}^{(k)}[t \mid t-1]\B C^{\T{(k)}}(s) + \B \Sigma^{(k)} (s)\bigr).
\end{align}

\subsection{M-step of EM algorithm}
In the M-step, the parameter values are updated to maximize $\mathcal{Q}\bigl(\B \Theta \mid \B \Theta(s)\bigr)$. 
Unlike the case of a single LGSSM, it is not appropriate to simply calculate the parameter values at which the partial derivatives of $\mathcal{Q}$ with respect to $\B \Theta$ vanish, because we must take account of the constraint (\ref*{Eq:12}). 
Instead, we employ the Lagrangian multiplier method as follows:
\begin{align}
    \label{Eq:20}
    \frac{\partial}{\partial p^{(k)}} \left( \mathcal{Q}\bigl(\B \Theta \mid \B \Theta(s)\bigr) - \lambda \Bigl( \sum_{k=1}^M p^{(k)} - 1 \Bigr) \right)
    = \frac{\partial}{\partial p^{(k)}} \mathcal{Q}\bigl(\B \Theta \mid \B \Theta(s)\bigr) - \lambda
    =0 .
\end{align}
For the parameters other than $p^{(k)}$, which are unconstrained, their values can be obtained by maximizing $\mathcal{Q}$ directly; that is, we solve 
\begin{align}
    \label{Eq:21}
    \frac{\partial}{\partial \B \theta^{(k)}}  \mathcal{Q}\bigl(\B \Theta \mid \B \Theta(s)\bigr)
    = \sum_{i=1}^{N} p\bigl(\omega^{(k)} \mid \B Y_i,\,\B \Theta(s)\bigr) \frac{\partial}{\partial \B \theta^{(k)}} Q_{i}^{(k)}\bigl(\B \theta^{(k)},\,\B \theta^{(k)}(s)\bigr)
    = 0 .
\end{align}
Solving (\ref*{Eq:20}) and (\ref*{Eq:21}), we obtain the analytical expressions (\ref*{Eq:B6}), (\ref*{Eq:B8}), (\ref*{Eq:B9}), (\ref*{Eq:B12}), (\ref*{Eq:B13}), (\ref*{Eq:B16}), and (\ref*{Eq:B17}) of the new parameter values, as shown in Appendix B\@. 
The detailed algorithm for MLGSSMs is described in Algorithm \ref*{em_for_MLGSSM}.
We stop the EM algorithm when the sum of the absolute values of the differences between $\B \Theta(s+1)$ and $\B \Theta(s)$ becomes smaller than a small threshold $\varepsilon$.

\subsection{Clustering with posterior probability}
Once we obtain the optimal parameters $\B \Theta^{*} = \bigl\{\bigl(\B \theta^{{(1)}^{*}},\,p^{{(1)}^{*}}\bigr),\;\dotsc\,,\;\bigl(\B \theta^{{(M)}^{*}},\,p^{{(M)}^{*}}\bigr)\bigr\}$ by alternately repeating the E-step and M-step until a convergence criterion is satisfied, the cluster index $\B z_i^{*}$ of the $i$-th time series can be inferred as
\begin{align}
    \B z_i^{*} = \underset{k}{\argmax} \,p\bigl(\omega^{(k)} \mid \B Y_i,\,\B \theta^{{(k)}^{*}},\,p^{{(k)}^{*}}\bigr) .
\end{align}

\subsection{Two practical problems and their solutions}
There are two problems when estimating MLGSSM parameters. 
First, the representation of an LGSSM is not unique, as described in Sec.~3. 
To address this problem, we fix all elements in the first row of each observation matrix $\B C^{(k)}$ to 1 without loss of generality. 
This makes $\B C^{(k)} = \left[ 1 \,\dotsc\, 1 \right]$ if $d_y = 1$. 
Although this restriction does not guarantee the identifiability of the model, the parameters of the LGSSM turn out to converge in our experiments. 

Second, the EM algorithm is sensitive to the initial parameter values, which can cause the parameters to plunge into the local optimum. 
In Ref.~\cite{XIONG20041675}, to overcome this problem, a set of the AR coefficients estimated from all the time series is fed to the $k$-means clustering algorithm, and the obtained centers of the parameter values are used as the initial parameter values for an MAR\@. 
We incorporate this initialization procedure into our method. 
In our method, the EM algorithm is used for the parameter estimation of not only MLGSSMs but also single LGSSMs, as described in Sec.~3.
Therefore, the parameter initialization is performed twice to estimate an MLGSSM\@.  

\begin{figure}[tbp]
    \centering{\includegraphics[width=140mm]{"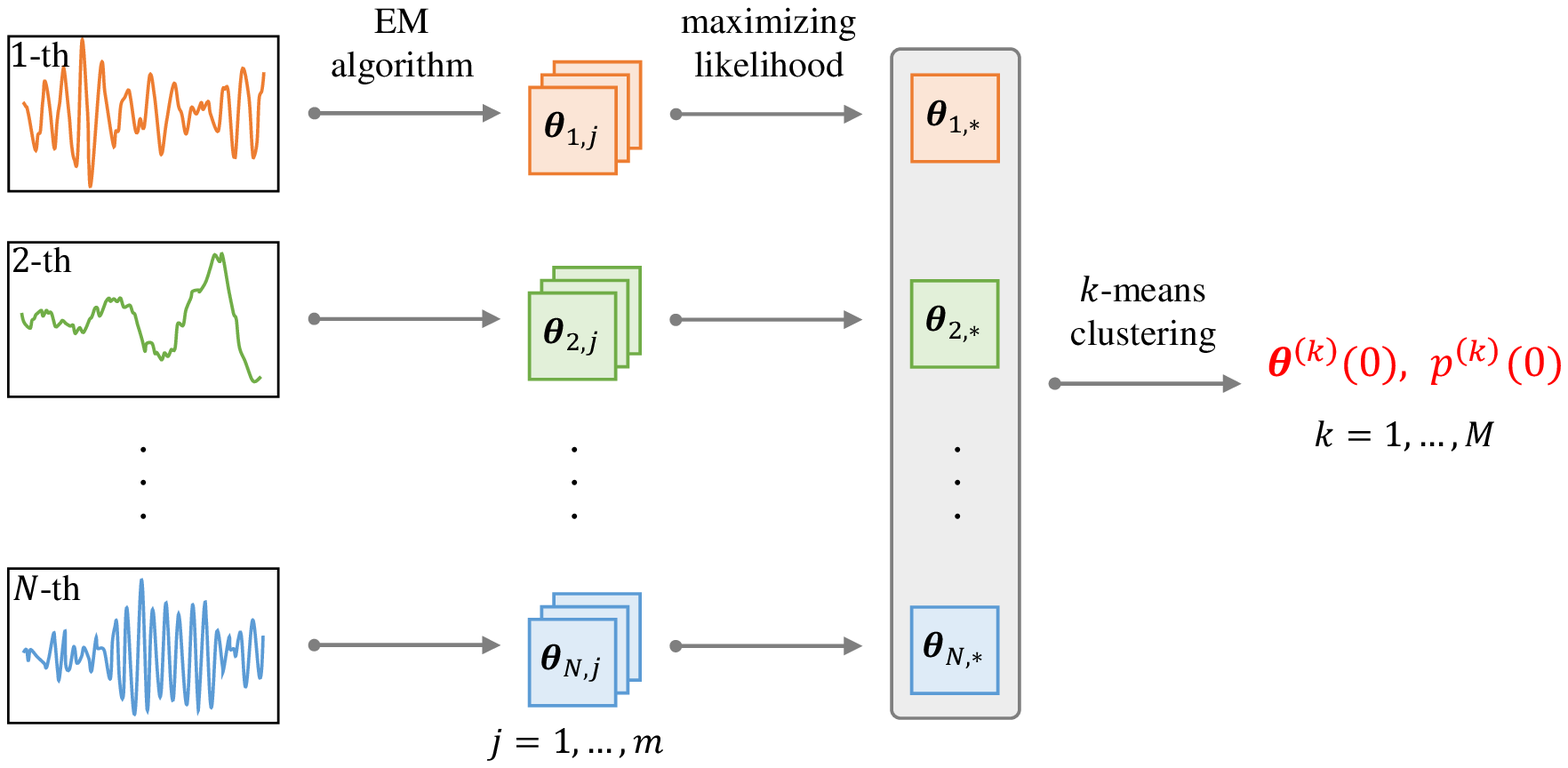"}}
    \caption{
        Parameter initialization of EM algorithm for MLGSSMs. 
        First, $m$ LGSSMs are generated for each time series, and they are trained with the same time series and $m$ different initial parameter values. 
        Second, the best parameter set for each time series is selected to maximize the likelihood of the LGSSM\@. 
        Third, the $N$ best parameter sets are fed to the $k$-means clustering algorithm to obtain the centers of the parameter values and the proportion of members in each cluster, which are used as the initial parameter values for the MLGSSM\@.
        }
    \label{fig3}
\end{figure}

The process of parameter initialization for the MLGSSM is shown in Fig.~\ref*{fig3}. 
In the first step, $m$ LGSSMs are generated for each time series, and they are trained with the same time series and $m$ different initial parameter values. 
As a result, we obtain $m$ sets, $\B \theta_{i,1}$,\, $\dotsc$\,, \,$\B \theta_{i,m}$, of estimated parameter values for the $i$-th time series. 
In the second step, the best parameter set, $\B \theta_{i,*}$, for the $i$-th time series is selected to maximize the likelihood of the LGSSM as
\begin{align}
    \B \theta_{i,*} \leftarrow \B \theta_{i,j^{*}} \quad \text{with} \quad j^{*} = \underset{j}{\argmax} \,p\bigl(\B \theta_{i,j} \mid \B M_{i,j}\bigr) ,
\end{align}
where $p\bigl(\B \theta_{i,j} \mid \B M_{i,j}\bigr)$ is the likelihood of the $j$-th LGSSM $\B M_{i,j}$ for the $i$-th time series.
Repeating these two steps for $N$ time series, we obtain a set $\bigl\{\B \theta_{i,*} \mid i=1,\;\dotsc\,,\;N\bigr\}$ of best parameter sets. 
In the third step, this set is fed to the $k$-means algorithm to obtain the centers of the LGSSM parameter values and the proportion of members in each cluster. 
These centers and proportions are used as the initial parameter values $\B \Theta(0) = \{(\B \theta^{(1)}(0),\,p^{(1)}(0)),\;\dotsc\,,\;(\B \theta^{(M)}(0),\,p^{(M)}(0))\}$ for the MLGSSM\@. 
Note that before feeding a set of parameter sets to the $k$-means algorithm, all matrices are converted into vectors, and all vectors are combined into one vector.

\begin{figure}[t]
    \begin{algorithm}[H]
        \caption{EM Algorithm for MLGSSMs}
        \label{em_for_MLGSSM}
        \begin{algorithmic}[1]
            \REQUIRE $\B \Theta(0)$, $\varepsilon_0$, and $maxiter$
            \FOR{$s=0,\;\dotsc\,,\;maxiter$}
            \STATE // E-step
            \FOR{$i=1,\;\dotsc\,,\;N$} 
            \FOR{$k=1,\;\dotsc\,,\;M$}
            \STATE Run the Kalman filter and smoother (Algorithm \ref*{kalmanfilter}) to obtain $\B \mu$, $\B V$, and $\B J$
            \STATE Compute the expectations (\ref*{Eq:6}) with $\B \mu$, $\B V$, and $\B J$
            \STATE Compute $p\bigl(\B Y_i \mid \omega^{(k)},\,\B \theta^{(k)}(s)\bigr)$ (Eq.~(\ref*{Eq:19})) with $\B \mu$ and $\B V$
            \ENDFOR
            \FOR{$k=1,\;\dotsc\,,\;M$}
            \STATE Compute $p\bigl(\omega^{(k)} \mid \B Y_i,\,\B \Theta(s)\bigr)$ (Eq.~(\ref*{Eq:18}))
            \ENDFOR
            \ENDFOR
            \STATE // M-Step
            \FOR{$k=1,\;\dotsc\,,\;M$}
            \STATE Compute (\ref*{Eq:B6}), (\ref*{Eq:B8}), (\ref*{Eq:B9}), (\ref*{Eq:B12}), (\ref*{Eq:B13}), (\ref*{Eq:B16}), and (\ref*{Eq:B17}) to obtain $\B \theta^{(k)}(s+1)$ and $p^{(k)}(s+1)$
            \ENDFOR
            \STATE // Convergence criterion
            \STATE $\varepsilon \leftarrow \bigl\| \B \Theta(s+1) - \B \Theta(s) \bigr\|_1$
            \IF {$\varepsilon < \varepsilon_0$}
            \RETURN $\B \Theta(s+1)$
            \ENDIF
            \ENDFOR
            \RETURN $\B \Theta(s+1)$
        \end{algorithmic}
    \end{algorithm}
\end{figure}

\section{Experiments}
In this section, we will demonstrate via experiments on a simulated dataset that the proposed method successfully achieves clustering, parameter estimation, and model selection. 
We will also apply our method to real datasets and compare its clustering accuracy to the accuracy of other methods reported in previous studies. 
The code is available at \url{https://github.com/ur17/em_mlgssm}.

\subsection{Simulated dataset}
We will first use a synthetic time series dataset generated from known LGSSMs. 
The dataset comprises $60$ time series with a length $T = 1000$. 
All time series are generated from different LGSSMs with two state variables and one observed variable, and we assume that their LGSSMs are divided into three groups by the parameter values. 
Specifically, the parameter matrices for the $i$-th LGSSM are
\begin{align*}
    \B \Gamma_{i} =
    \begin{bmatrix*}[r]
        0.01 & 0 \\
        0 & 0.01 \\
    \end{bmatrix*} , \quad
    \B C_{i} = \left[1 \quad 1\right] , \quad
    \B \Sigma_{i} = 0.01 , \quad
    \B \mu_{i} =
    \begin{bmatrix*}[r]
        0 \\
        0 \\
    \end{bmatrix*} , \quad
    \B P_{i} =
    \begin{bmatrix*}[r]
        0.01 & 0 \\
        0 & 0.01 \\
    \end{bmatrix*} ,
\end{align*}
and
\begin{figure}[t]
    \begin{tabular}{cc}
        \centering

        \begin{minipage}[c]{0.46\linewidth}
            \centering
            \includegraphics[width=75mm]{"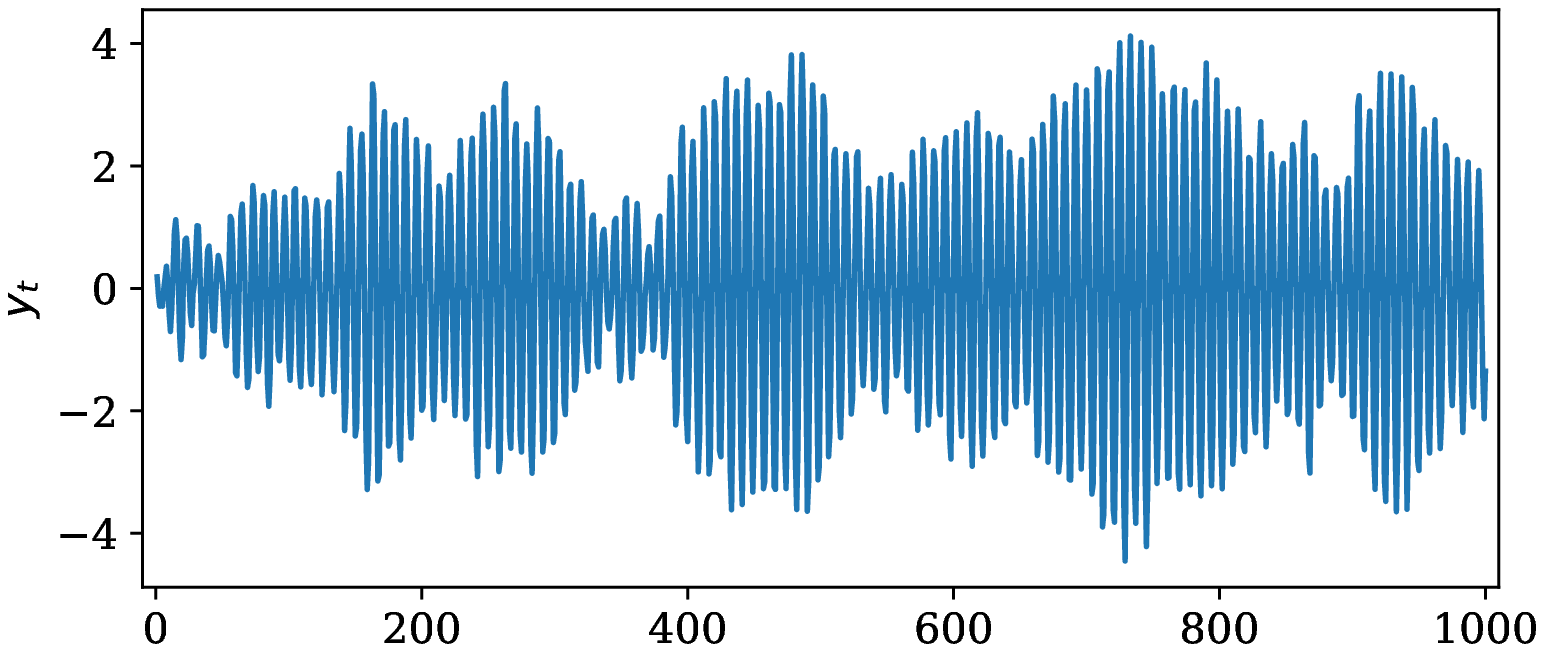"}
            \subcaption{Actual time series ($i=1$)}
            \vspace{5mm}
        \end{minipage} &

        \begin{minipage}[c]{0.46\linewidth}
            \centering
            \includegraphics[width=75mm]{"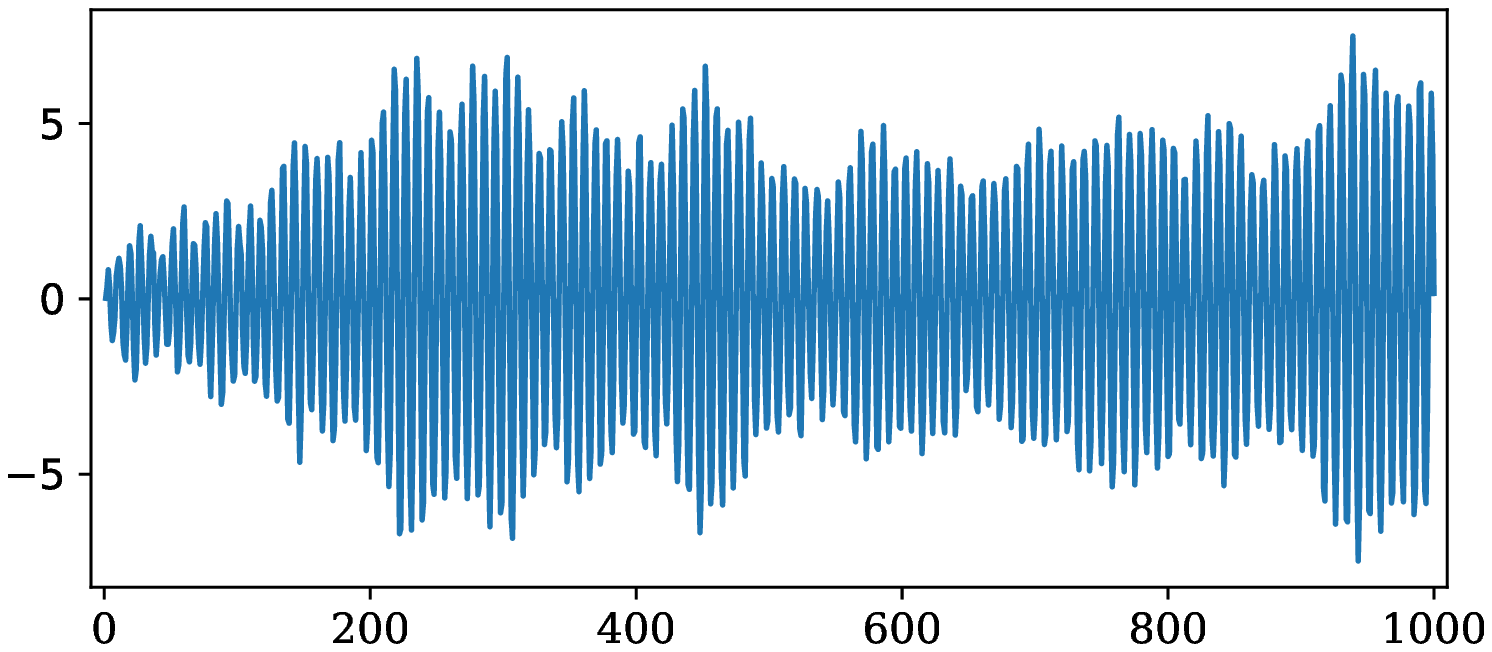"}
            \subcaption{Predicted time series (cluster $1$)}
            \vspace{5mm}
        \end{minipage} \\

        \begin{minipage}[c]{0.46\linewidth}
            \centering
            \includegraphics[width=75mm]{"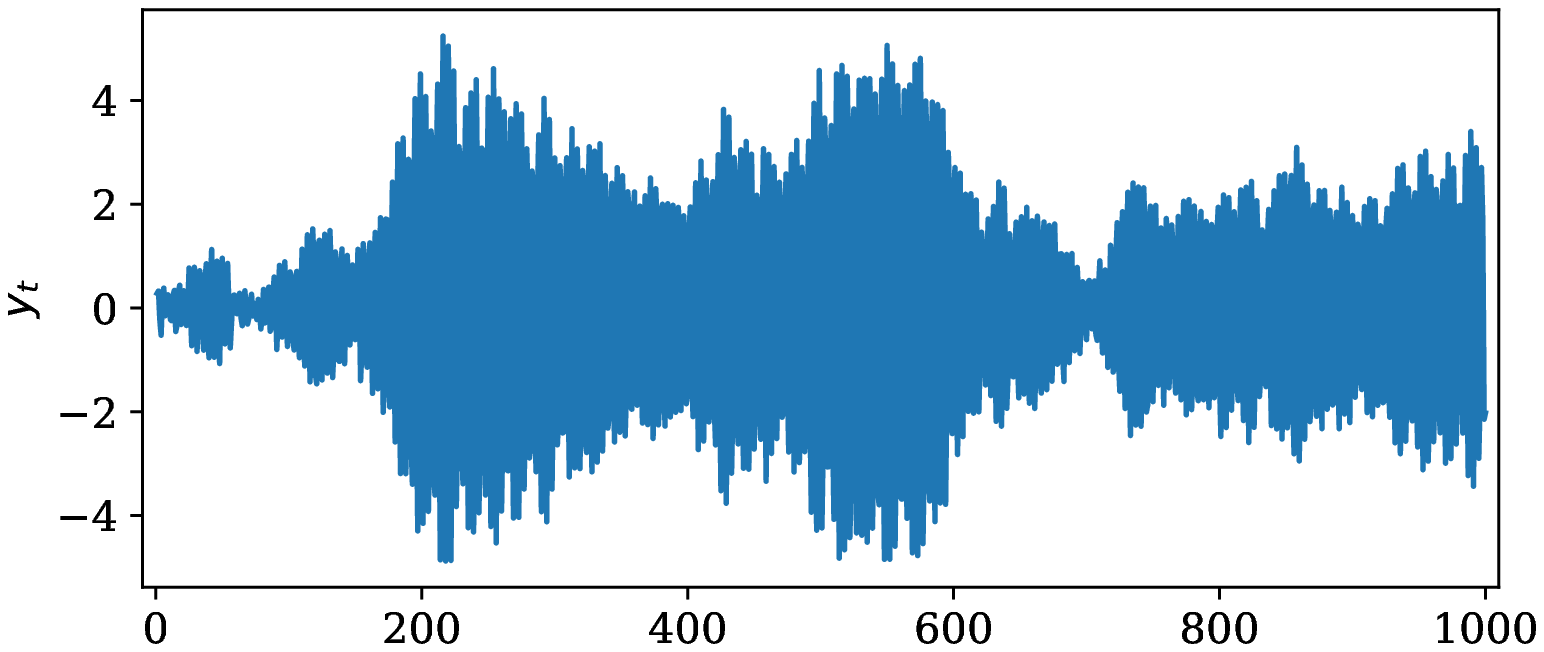"}
            \subcaption{Actual time series ($i=21$)}
            \vspace{5mm}
        \end{minipage} &

        \begin{minipage}[c]{0.46\linewidth}
            \centering
            \includegraphics[width=75mm]{"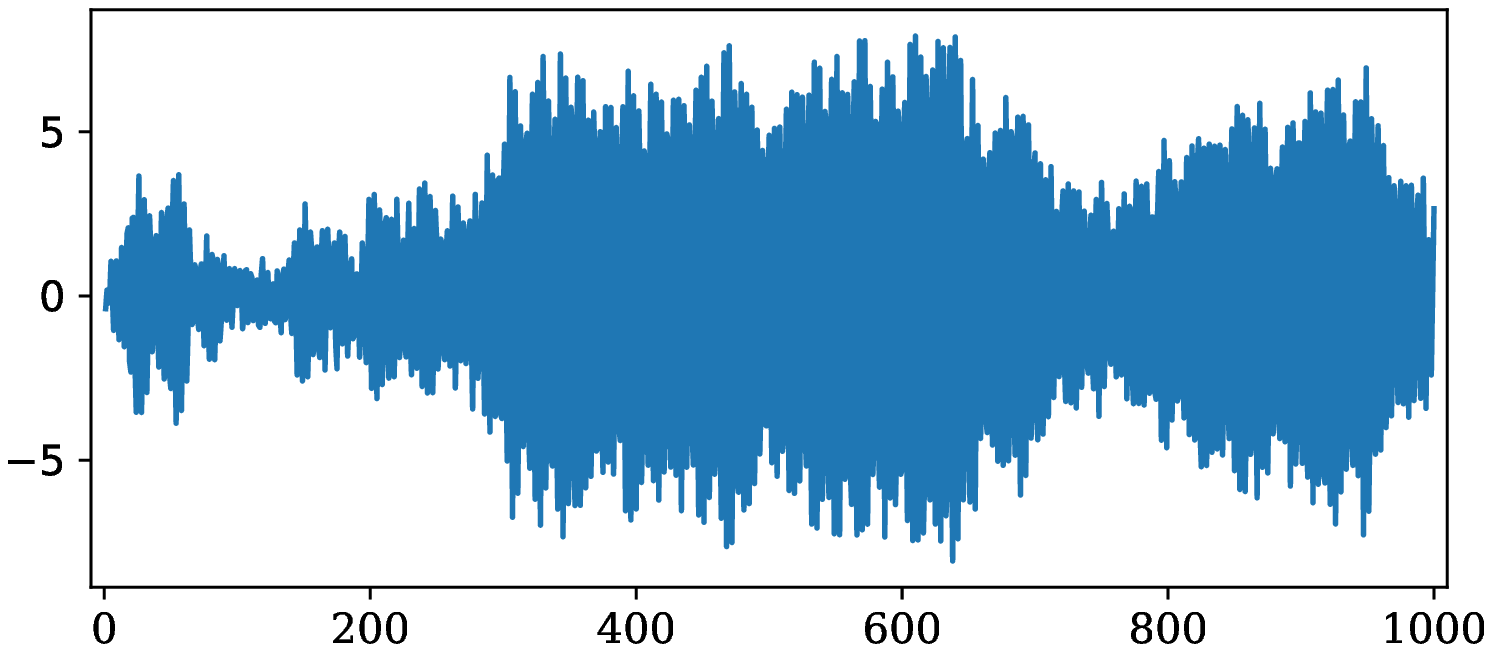"}
            \subcaption{Predicted time series (cluster $2$)}
            \vspace{5mm}
        \end{minipage} \\

        \begin{minipage}[c]{0.46\linewidth}
            \centering
            \includegraphics[width=75mm]{"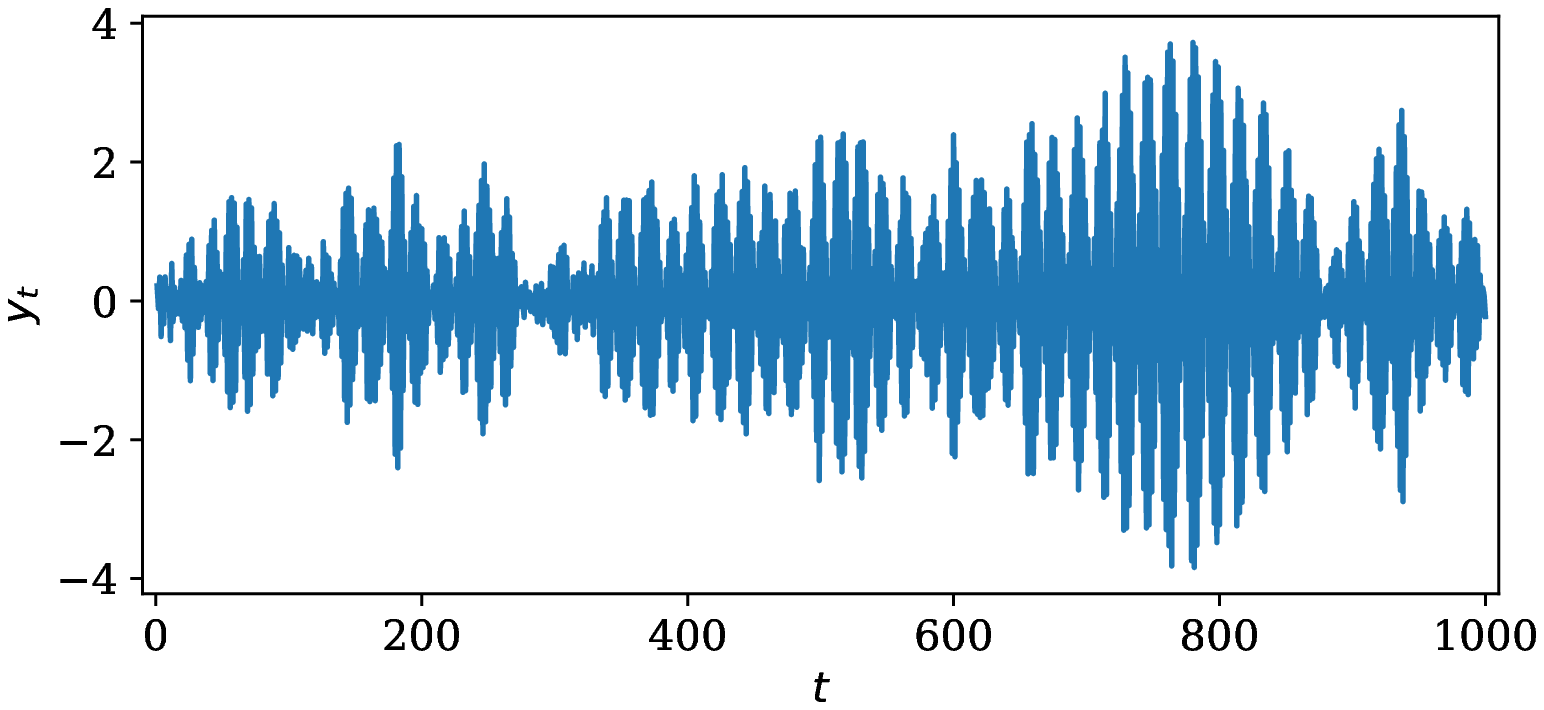"}
            \subcaption{Actual time series ($i=41$)}
        \end{minipage} &

        \begin{minipage}[c]{0.46\linewidth}
            \centering
            \includegraphics[width=75mm]{"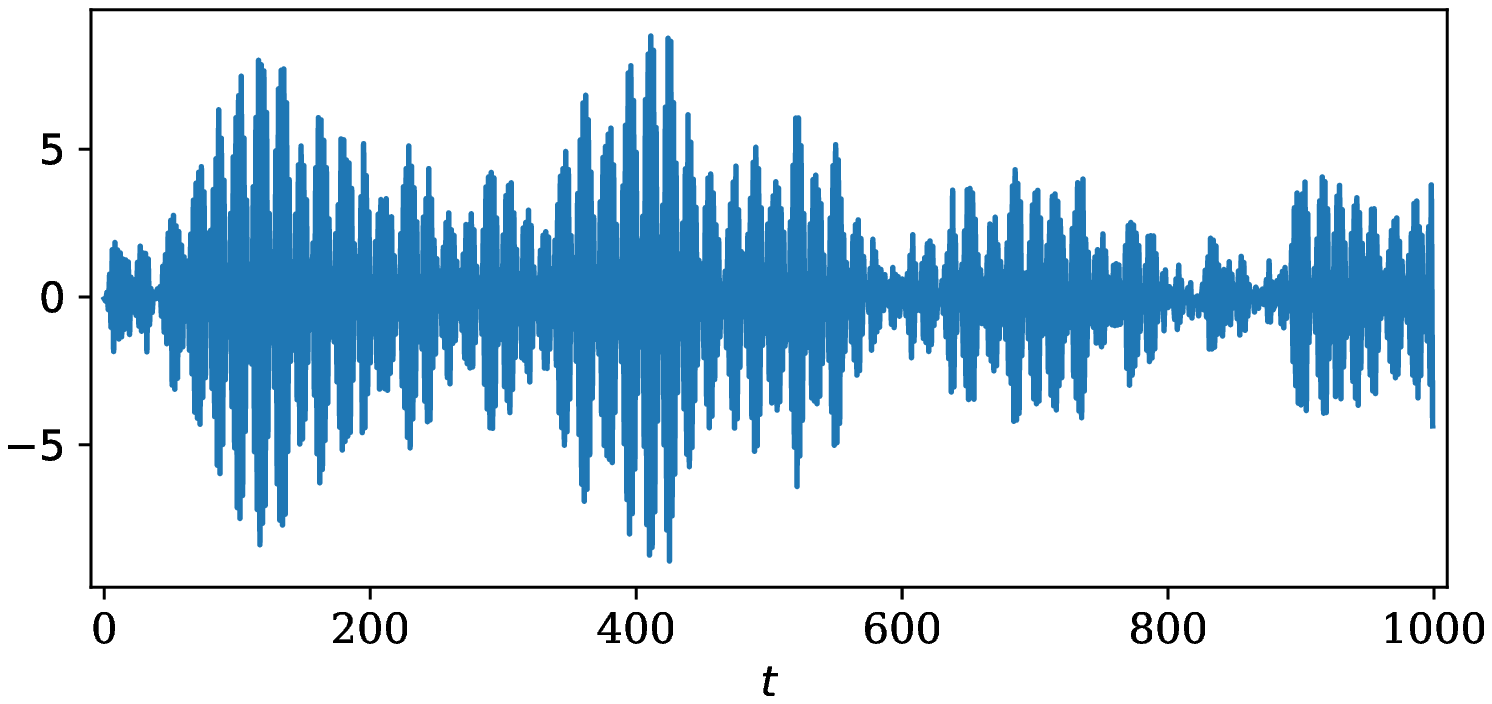"}
            \subcaption{Predicted time series (cluster $3$)}
        \end{minipage} \\

    \end{tabular}
    \caption{
        Representative samples of the training dataset and predicted time series obtained using the estimated parameters. 
        (a), (c), and (e) plot typical time series data for each group in the training dataset. 
        (b), (d), and (f) plot predicted time series obtained using the estimated LGSSM parameters for the corresponding cluster.
        }
    \label{fig2}
\end{figure}
\begin{align*}
    \B A_{i} =
    \begin{bmatrix*}[r]
        \cos \theta_{i} & - \sin \theta_{i} \\
        \sin \theta_{i} & \cos \theta_{i} \\ 
    \end{bmatrix*} ,
\end{align*}
where the values of $\theta_{i}$ are randomly sampled from the uniform distributions $\mathcal{U}\bigl((40/180)\pi,\,(45/180)\pi\bigr)$ for $i = 1,\;\dotsc\,,\;20$, $\mathcal{U}\bigl((80/180)\pi,\,(90/180)\pi\bigr)$ for $i = 21,\;\dotsc\,,\;40$, and $\mathcal{U}\bigl((160/180)\pi,\,\pi\bigr)$ for $i = 41,\;\dotsc\,,\;60$. 
For the upper and lower bounds of $\theta_i$, the values of the state matrices $\B A_i$ are calculated as
\begin{subequations}
    \renewcommand{\theequation}{
    \theparentequation-\alph{equation}  
    }
    \label{Eq:24}
    \begin{align}
        \left.
        \begin{aligned}
            \begin{bmatrix*}[r]
                \cos (40/180) \pi & -\sin (40/180) \pi\\
                \sin (40/180) \pi & \cos (40/180) \pi
            \end{bmatrix*}
            &\approx
            \begin{bmatrix*}[r]
                0.766 & -0.643\\
                0.643 & 0.766
            \end{bmatrix*}\\
            \begin{bmatrix*}[r]
                \cos (45/180) \pi & -\sin (45/180) \pi\\
                \sin (45/180) \pi & \cos (45/180) \pi
            \end{bmatrix*}
            &\approx
            \begin{bmatrix*}[r]
                0.707 & -0.707\\
                0.707 & 0.707
            \end{bmatrix*}
        \end{aligned}
        \right\}\ 
        & \text{for $i = 1,\;\dotsc\,,\;20$}, \\
        \left.
            \begin{aligned}
                \begin{bmatrix*}[r]
                    \cos (80/180) \pi & -\sin (80/180) \pi\\
                    \sin (80/180) \pi & \cos (80/180) \pi
                \end{bmatrix*}
                &\approx
                \begin{bmatrix*}[r]
                    0.174 & -0.985\\
                    0.985 & 0.174
                \end{bmatrix*}\\
                \begin{bmatrix*}[r]
                    \cos (90/180) \pi & -\sin (90/180) \pi\\
                    \sin (90/180) \pi & \cos (90/180) \pi
                \end{bmatrix*}
                &=
                \begin{bmatrix*}[r]
                    0 & -1\\
                    1 & 0
                \end{bmatrix*}
            \end{aligned}
        \right\}\ 
        & \text{for $i = 21,\;\dotsc\,,\;40$}, \\
        \left.
            \begin{aligned}
                \begin{bmatrix*}[r]
                    \cos (160/180) \pi & -\sin (160/180) \pi\\
                    \sin (160/180) \pi & \cos (160/180) \pi
                \end{bmatrix*}
                &\approx
                \begin{bmatrix*}[r]
                    -0.940 & -0.342\\
                    0.342 & -0.940
                \end{bmatrix*}\\
                \begin{bmatrix*}[r]
                    \cos \pi & -\sin \pi\\
                    \sin \pi & \cos \pi
                \end{bmatrix*}
                &=
                \begin{bmatrix*}[r]
                    -1 & 0\\
                    0 & -1
                \end{bmatrix*}
            \end{aligned}
        \right\}\ 
        & \text{for $i = 41,\;\dotsc\,,\;60$}.
    \end{align}
\end{subequations}
Figures \ref*{fig2}(a), (c), and (e) show typical examples of time series in each group.

Throughout the experiments in this paper, the initial parameter values for the single LGSSM are set as
\begin{align*}
    \B \Gamma(0) = 0.05 \times \B I_{d_x}, \quad
    \B \Sigma(0) = 0.05 , \quad
    \B \mu(0) = \B O_{d_x \times 1}, \quad
    \B P(0) = 10^4 \times \B I_{d_x} ,
\end{align*}
where $\B I_{d_x}$ is a $d_x$-dimensional identity matrix and $\B O_{a \times b}$ is a ($a \times b$) zero matrix.
Let us assume, for the moment, that $d_x = 2$ is given.
These initial parameter values have rather large diagonal entries in the initial state covariance matrix $\B P$.
This is expected to mitigate the effect of the initial guesses in the Kalman filter.
As described in Sec.~4.5, the observation matrix $\B C$ is fixed to $\left[ 1 \quad 1 \right]$. 
The initial parameter value, $\B A(0)$, for the state matrix is randomly set as $\B A (0) = \B Q$, where $\B Q$ is an orthogonal matrix obtained by QR decomposition of a ($d_x \times d_x$) standard Gaussian random matrix $\B G$.
Note that this initialization method is not deterministic because the matrix $\B G$ is randomly generated.

\subsubsection{Clustering and parameter estimation}
We evaluate the proposed method for different values of the hyperparameters associated with parameter initialization.
The number $m$ of LGSSMs is selected from $\{10,\,20,\,30,\,40,\,50\}$.
Considering the randomness of the matrix $\B G$ and the $k$-means algorithm, we perform the experiment six times for each value of $m$.
Consequently, we obtain $30$ different results for clustering and parameter estimation.

Table \ref*{total_confusion_matrix} shows the total confusion matrix obtained by clustering the simulated dataset using an MLGSSM.
This result indicates that perfect clustering is achieved.
Table \ref*{pred_param} shows the sample mean of the estimated parameter values for each cluster.
The values in parentheses are standard deviations.
All entries in the state matrix $\B A$ fall within the range (\ref*{Eq:24}) of the true values for each cluster.
Although the estimations of the covariance matrices $\B \Gamma$ and $\B P$ result in rather low accuracy for cluster $3$, this can be attributed to the particularly wide range of the true state matrices for cluster $3$ because such a wide range makes it difficult to represent the state matrices in a single matrix and consequently degrades the estimation accuracy for the covariance matrices.

Figures \ref*{fig2}(b), (d), and (f) show typical examples of predicted time series for each cluster obtained using the estimated parameter values. 
These time series are similar to the samples of the simulated dataset because the parameters are accurately estimated.

\subsubsection{Model selection}
In the above experiment, the number $M$ of clusters and the number $d_x$ of state variables were fixed, but in general they are not known a priori. 
Here, we will use the Bayesian information criterion (BIC) to determine these values, as applied in Refs.~\cite{XIONG20041675,Cen2000}. 

The BIC is defined as
\begin{align}
    \text{BIC} = \log\, L - \dfrac{1}{2} \bigl(P-1\bigr) \log\, N ,
\end{align}
where $L$ is the likelihood of a model, $P$ is the number of model parameters, and $N$ is the number of data values. 
A model with a larger BIC value is preferred. 
The BIC for an MLGSSM is expressed as
\begin{align}
    \label{Eq:BIC_MLGSSM}
    \text{BIC} = \sum_{i=1}^{N} \log\, \Bigl( \sum_{k = 1}^{M} p\bigl(\B Y_i \mid \omega^{(k)},\B \theta^{(k)}\bigr){p^{(k)}} \Bigr) 
               \,- \,\dfrac{1}{2} \Bigl(M \bigl|\B \theta^{(1)} \bigr| + M - 1\Bigr) \log\, N ,
\end{align}
where $\bigl|\B \theta^{(1)} \bigr|$ is the number of elements in $\B \theta^{(1)}$.

We choose the values of $M$ and $d_x$ from $M = 2,\,3,\,4,\,5$ and $d_x = 2,\,3,\,4$, respectively. 
The BIC value for each pair of $M$ and $d_x$ is shown in Table \ref*{bic_simulation}. 
As indicated, the BIC reaches a maximum at the correct values $M = 3$ and $d_x = 2$. 
This suggests that the proper model can be obtained using the BIC.

\subsubsection{Time complexity of proposed method}
To examine the time complexity of the proposed method, we apply this method to simulated datasets of different sizes and measure the required computational time in each case.
The datasets are generated using the same settings as above except for the number $N$ and the length $T$ of the time series.
Because of the randomness of the dataset generation process, we measure the average time over five trials for each setting.

\begin{figure}[t]
    \begin{tabular}{cc}
        \centering

        \begin{minipage}[c]{0.46\linewidth}
            \centering
            \includegraphics[width=75mm]{"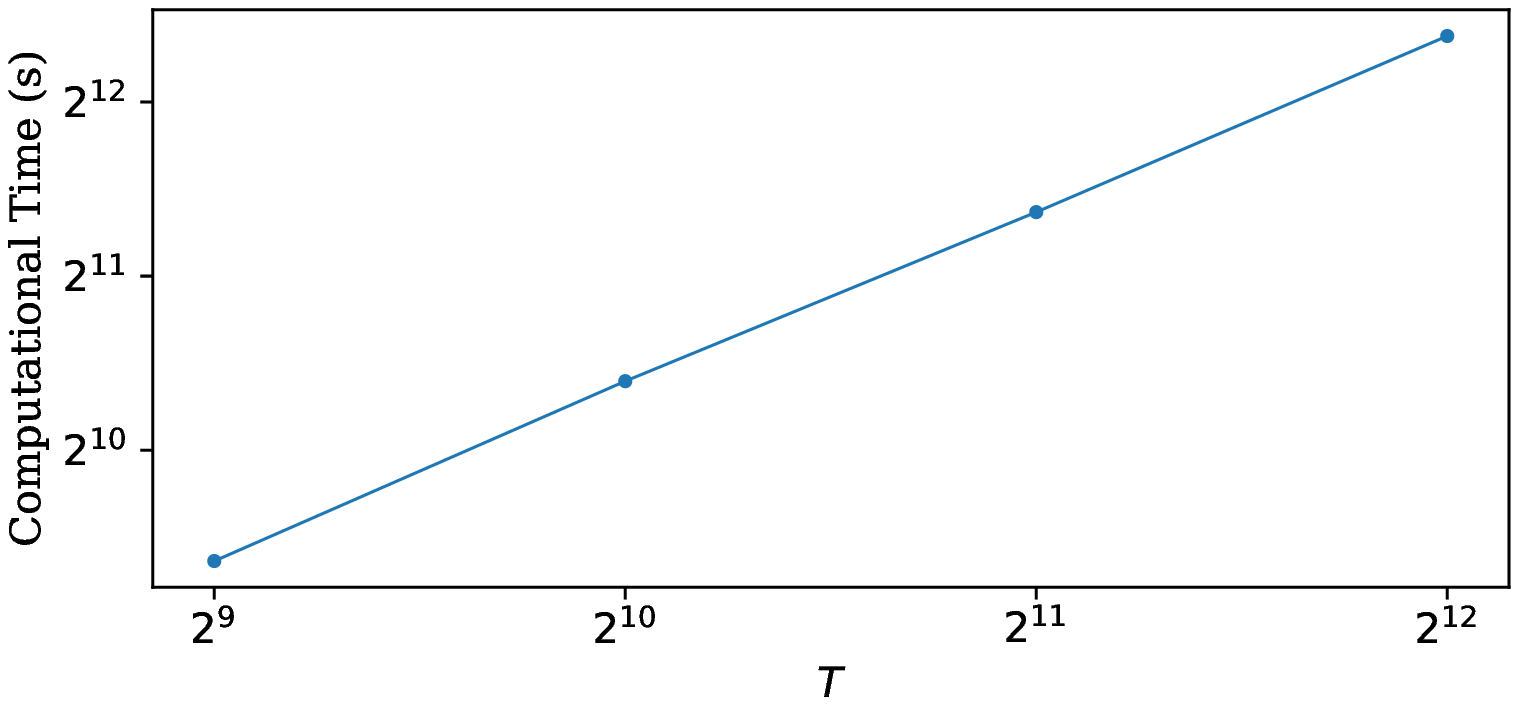"}
            \subcaption{Computational time for different values of $T$}
            \vspace{5mm}
        \end{minipage} &

        \begin{minipage}[c]{0.46\linewidth}
            \centering
            \includegraphics[width=75mm]{"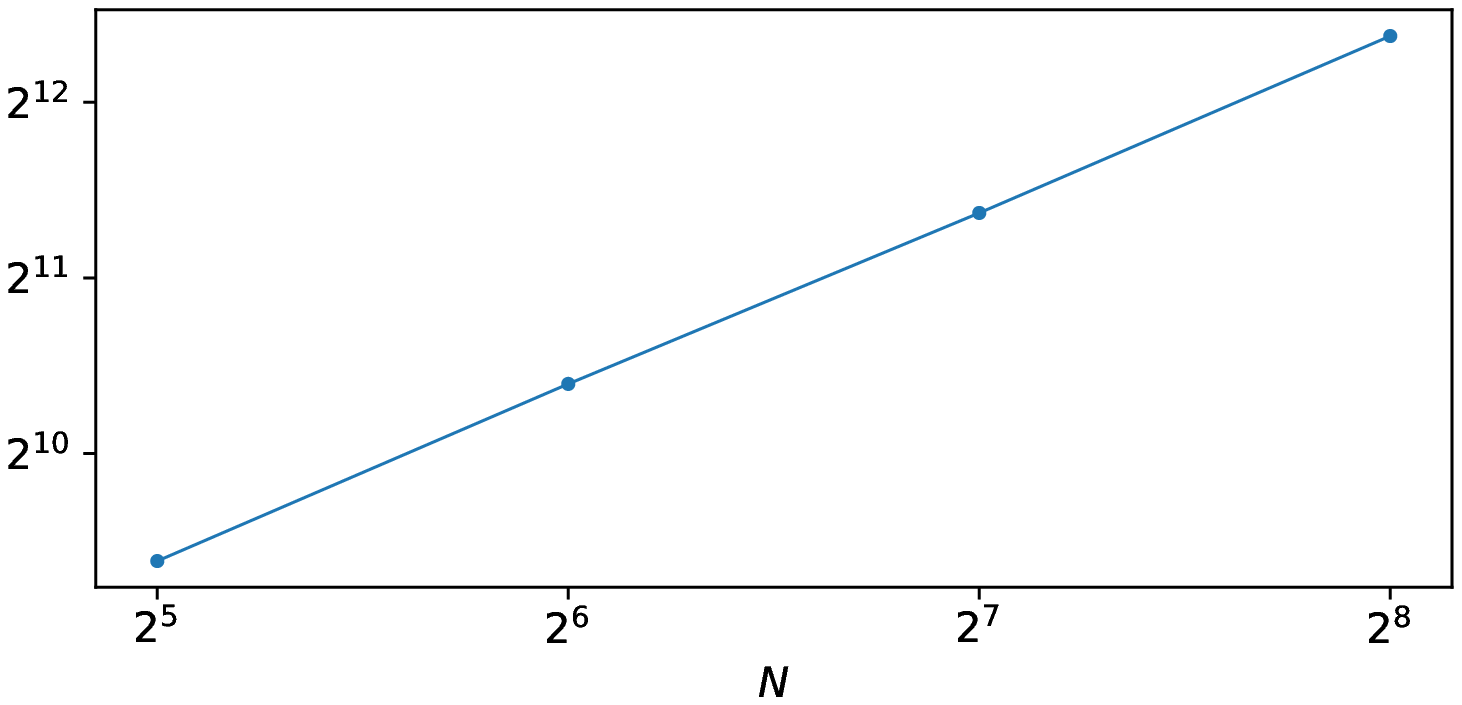"}
            \subcaption{Computational time for different values of $N$}
            \vspace{5mm}
        \end{minipage}

    \end{tabular}
    \caption{
        Time complexity of algorithm.
        (a) and (b) plot the average computational time for five runs of the proposed method on simulated datasets of different sizes.
    }
    \label{time}
\end{figure}

First, let us examine the dependence of the computational time on $T$.
We choose the value of $T$ from $\{512,\,1024,\,2048,\,4096\}$ while fixing the value of $N$ to $64$.
Figure \ref*{time}(a) shows the computational time as a function of $T$ on a log-log scale.
These results are well fitted by a linear regression line with a slope of $1.004$.

Next, we measure the computational time for $T = 1024$ and $N \in \{32,\,64,\,128,\,256\}$.
Figure \ref*{time}(b) shows the computational time as a function of $N$ on a log-log scale.
These results are well fitted by a linear regression line with a slope of $0.995$.

These results indicate that the computational time for the proposed method is linear with respect to $T$ and $N$.

\clearpage

\begin{table}[H]
    \caption{Total confusion matrix obtained by clustering simulated dataset with MLGSSM.}
    \label{total_confusion_matrix}
    \centering
    \begin{tabular}{|l|c|c|c|} \hline
        \diagbox[width=33mm]{True}{Pred} & 1 & 2 & 3 \\ \hline
        $1$ ($i = 1,\;\dotsc\,,\;20$) & 600 & 0 & 0 \\ \hline
        $2$ ($i = 21,\;\dotsc\,,\;40$) & 0 & 600 & 0 \\ \hline
        $3$ ($i = 41,\;\dotsc\,,\;60$) & 0 & 0 & 600 \\ \hline 
    \end{tabular}
\end{table}

\begin{table}[H]
    \centering
    \caption{
        Sample mean of estimated parameter values for MLGSSM with $d_x = 3$ for simulated dataset (rounded to three decimal places).
        The values in parentheses are standard deviations.
        }
        \label{pred_param}
    \begin{tabular}{|c||cccc|}
        \hline
        $M$ & \rule[0mm]{0mm}{4mm} & $\B A$ & $\B \Gamma$ & $\B \Sigma$ \\
        \hline
        \multirow{7}{*}{1}
        & \rule[-7mm]{0mm}{16mm} &
        ${
            \begin{bmatrix*}[r]
                0.730(0.002) & -0.679(0.000) \\
                0.680(0.000) & 0.733(0.002) \\
            \end{bmatrix*}
        }$ 
        &
        ${
            \begin{bmatrix*}[r]
                0.029(0.000) & -0.004(0.000) \\
                -0.004(0.000) & 0.022(0.000) \\
            \end{bmatrix*}
        }$ 
        &
        0.021(0.000) 
        \\
        \cline{2-5}
        & \rule[0mm]{0mm}{4mm} & $\B \mu$ & $\B P$ & $p(\omega)$ \\
        \cline{2-5} 
        & \rule[-7mm]{0mm}{16mm} &
        ${
            \begin{bmatrix*}[r]
                -0.017(0.001) \\
                -0.013(0.001) \\
            \end{bmatrix*}
        }$
        &
        ${
            \begin{bmatrix*}[r]
                0.016(0.000) & -0.009(0.000) \\
                -0.009(0.000) & 0.019(0.000) \\
            \end{bmatrix*}
        }$
        &
        0.333
        \\
        \hline
        $M$ & \rule[0mm]{0mm}{4mm} & $\B A$ & $\B \Gamma$ & $\B \Sigma$ \\
        \hline
        \multirow{7}{*}{2}
        & \rule[-7mm]{0mm}{16mm} &
        ${
            \begin{bmatrix*}[r]
                0.088(0.002) & -0.993(0.000) \\
                0.993(0.000) & 0.091(0.002) \\
            \end{bmatrix*}
        }$ 
        &
        ${
            \begin{bmatrix*}[r]
                0.035(0.000) & 0.000(0.000) \\
                0.000(0.000) & 0.034(0.000) \\
            \end{bmatrix*}
        }$ 
        &
        0.024(0.000) 
        \\
        \cline{2-5}
        & \rule[0mm]{0mm}{4mm} & $\B \mu$ & $\B P$ & $p(\omega)$ \\
        \cline{2-5} 
        & \rule[-7mm]{0mm}{16mm} &
        ${
            \begin{bmatrix*}[r]
                0.011(0.000) \\
                0.028(0.000) \\
            \end{bmatrix*}
        }$
        &
        ${
            \begin{bmatrix*}[r]
                0.017(0.000) & -0.008(0.000) \\
                -0.008(0.000) & 0.015(0.000) \\
            \end{bmatrix*}
        }$
        &
        0.333
        \\
        \hline
        $M$ & \rule[0mm]{0mm}{4mm} & $\B A$ & $\B \Gamma$ & $\B \Sigma$ \\
        \hline
        \multirow{7}{*}{3}
        & \rule[-7mm]{0mm}{16mm} &
        ${
            \begin{bmatrix*}[r]
                -0.963(0.009) & -0.204(0.005) \\
                0.198(0.001) & -0.986(0.008) \\
            \end{bmatrix*}
        }$ 
        &
        ${
            \begin{bmatrix*}[r]
                0.059(0.003) & -0.032(0.000) \\
                -0.032(0.000) & 0.071(0.001) \\
            \end{bmatrix*}
        }$ 
        &
        0.019(0.000) 
        \\
        \cline{2-5}
        & \rule[0mm]{0mm}{4mm} & $\B \mu$ & $\B P$ & $p(\omega)$ \\
        \cline{2-5} 
        & \rule[-7mm]{0mm}{16mm} &
        ${
            \begin{bmatrix*}[r]
                0.008(0.016) \\
                -0.022(0.015) \\
            \end{bmatrix*}
        }$
        &
        ${
            \begin{bmatrix*}[r]
                0.062(0.011) & -0.053(0.012) \\
                -0.053(0.012) & 0.056(0.012) \\
            \end{bmatrix*}
        }$
        &
        0.333
        \\
        \hline
    \end{tabular}
\end{table}

\begin{table}[H]
    \centering
    \caption{
        BIC values for MLGSSMs with different values of $M$ and $d_x$ for simulated dataset (rounded to two decimal places).
        }

    \begin{tabular}{|c|c|c|c|c|}
        \hline
        \diagbox{$d_x$}{$M$} & 2 & 3 & 4 & 5 \\ \hline
        2 & 292.94 & 316.54 & 300.64 & 294.70 \\ \hline
        3 & 296.81 & 312.84 & 301.13 & 293.24 \\ \hline
        4 & 306.80 & 315.34 & 304.79 & 294.62 \\ \hline
    \end{tabular}
    \label{bic_simulation}
\end{table}

\clearpage

\subsection{Real datasets}
Next, we apply the proposed method to real datasets and compare the clustering results with those of previous studies \cite{XIONG20041675,4782736,989529}. 
These previous studies used real datasets for electrocardiogram (ECG), personal income, temperature, and population. 
Notably, while the methods used in Refs.~\cite{XIONG20041675,4782736} provided successful results for the first three datasets, they exhibited low accuracy for the last dataset (i.e., population). 
We demonstrate the effectiveness of the proposed method for all of these datasets, including the population dataset.

To evaluate the accuracy of the clustering results, we use the following cluster similarity measure \cite{Gavrilov} employed in Refs.~\cite{XIONG20041675,4782736,989529}:
\begin{align}
    \label{Eq:34}
    \mathrm{Sim}(G,\,\hat{G}) = \dfrac{1}{M}\sum_{i=1}^{M} \max_{1 \leq j \leq M} \dfrac{2\bigl|G_i \cap \hat{G}_j\bigr|}{\bigl|G_i\bigr|+\bigl|\hat{G}_j\bigr|} ,
\end{align}
where $G = \bigl\{G_i\bigr\}_{i = 1}^{M}$ and $\hat{G} = \bigl\{\hat{G}_i\bigr\}_{i = 1}^{M}$.
The value of this measure lies in the range $0$ to $1$, where $\mathrm{Sim}(G,\,\hat{G}) = 0$ implies that $G$ and $\hat{G}$ are completely dissimilar and $1$ implies that $G$ and $\hat{G}$ are the same.

For each dataset, the BIC (\ref*{Eq:BIC_MLGSSM}) is calculated for different values of $M$ and $d_x$ such that $2 \le M \le 4$ and $2 \le d_x \le 6$, to obtain the optimal values of their hyperparameters.
Then we perform the experiment with the optimal values of $M$ and $d_x$ ten times to calculate the sample mean and standard deviation of the cluster similarity score (\ref*{Eq:34}).

\begin{figure}[t]

    \begin{tabular}{cc}

        \centering

        \begin{minipage}[c]{0.46\linewidth}
            \centering
            \includegraphics[width=75mm]{"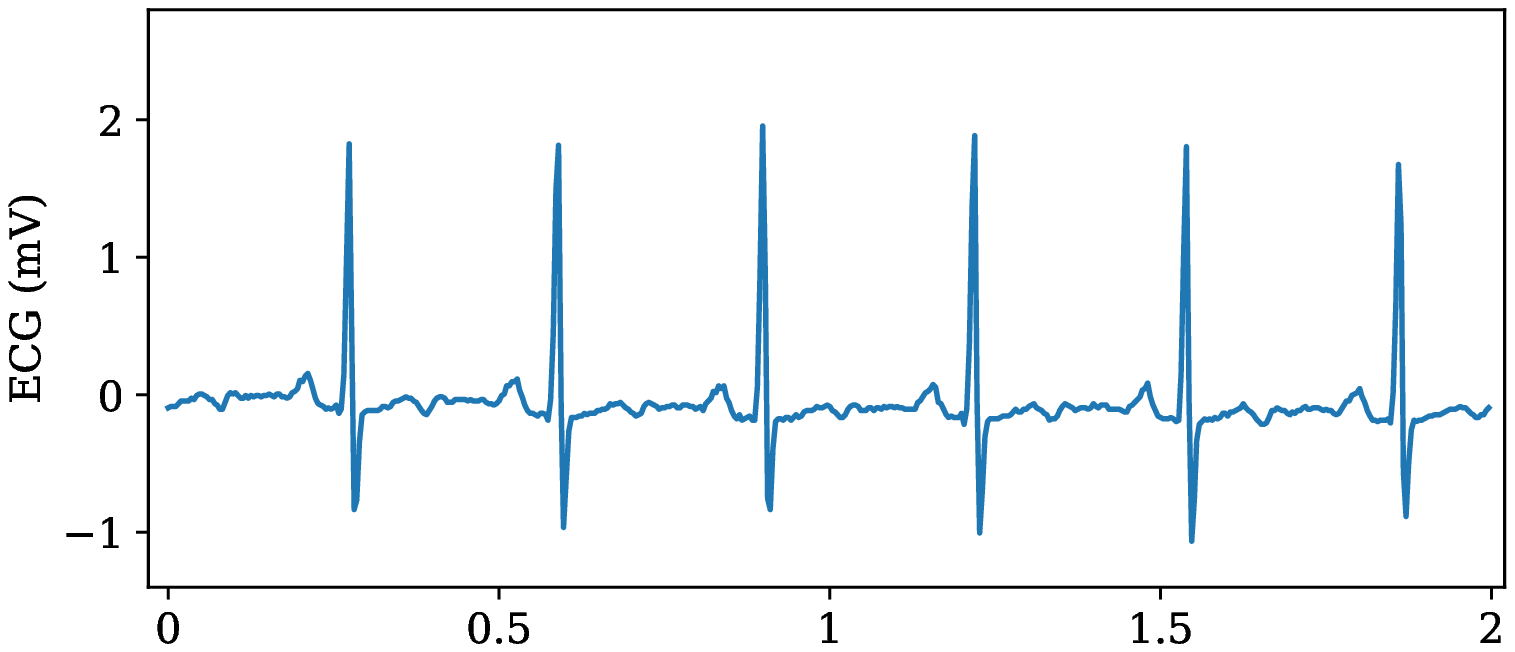"}
            \subcaption{Raw time series in group 1 (normal sinus rhythm)}
            \vspace{5mm}
        \end{minipage} &

        \begin{minipage}[c]{0.46\linewidth}
            \centering
            \includegraphics[width=75mm]{"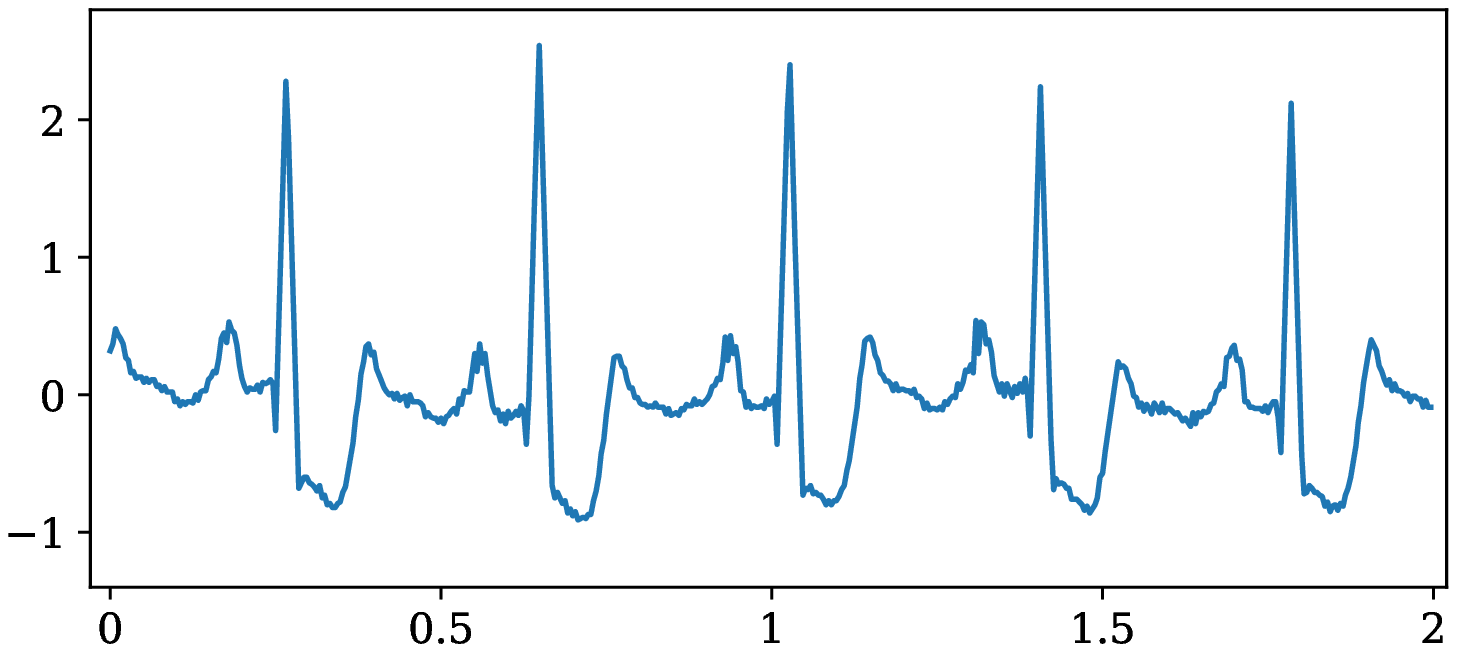"}
            \subcaption{Raw time series in group 2 (supraventricular arrhythmia)}
            \vspace{5mm}
        \end{minipage} \\

        \begin{minipage}[c]{0.46\linewidth}
            \centering
            \includegraphics[width=75mm]{"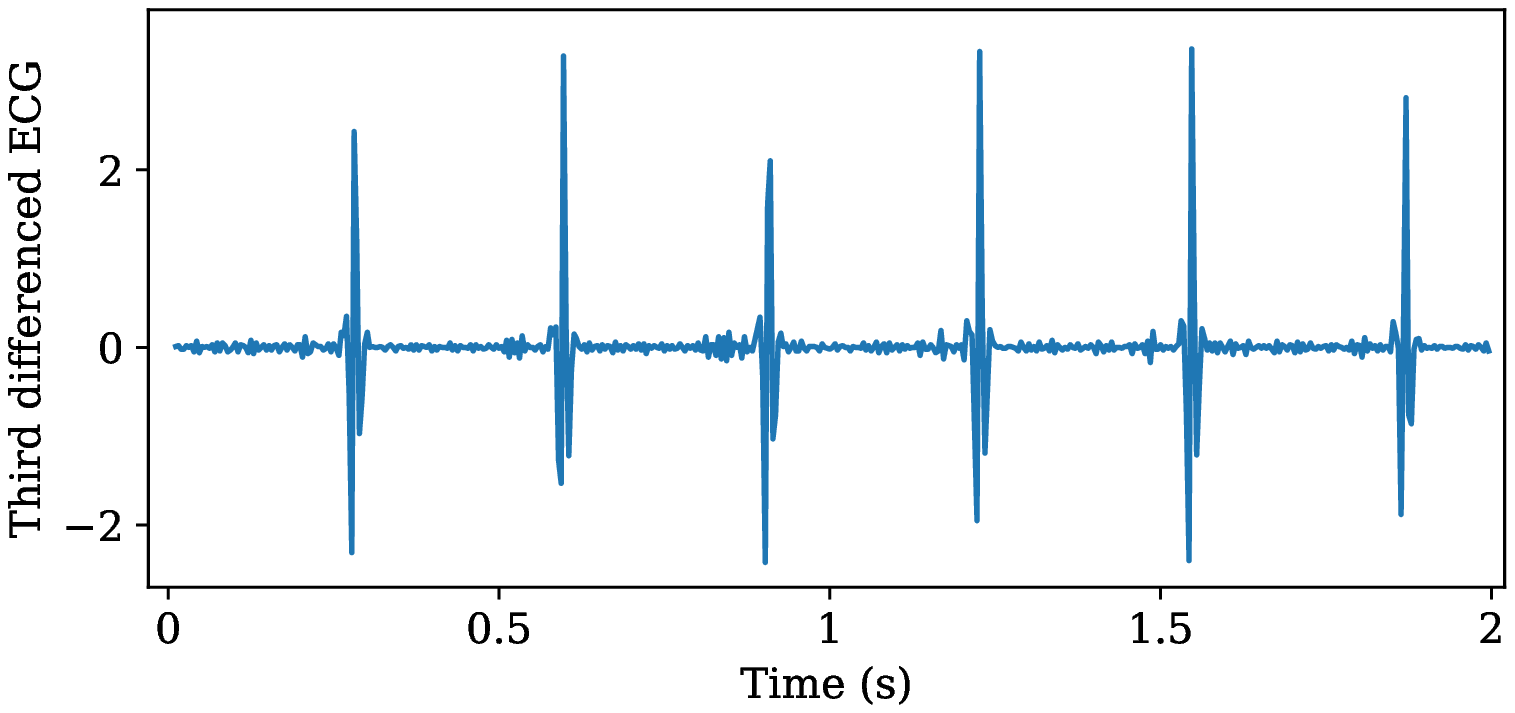"}
            \subcaption{Third-order differenced time series in group 1}
            \vspace{5mm}
        \end{minipage} &

        \begin{minipage}[c]{0.46\linewidth}
            \centering
            \includegraphics[width=75mm]{"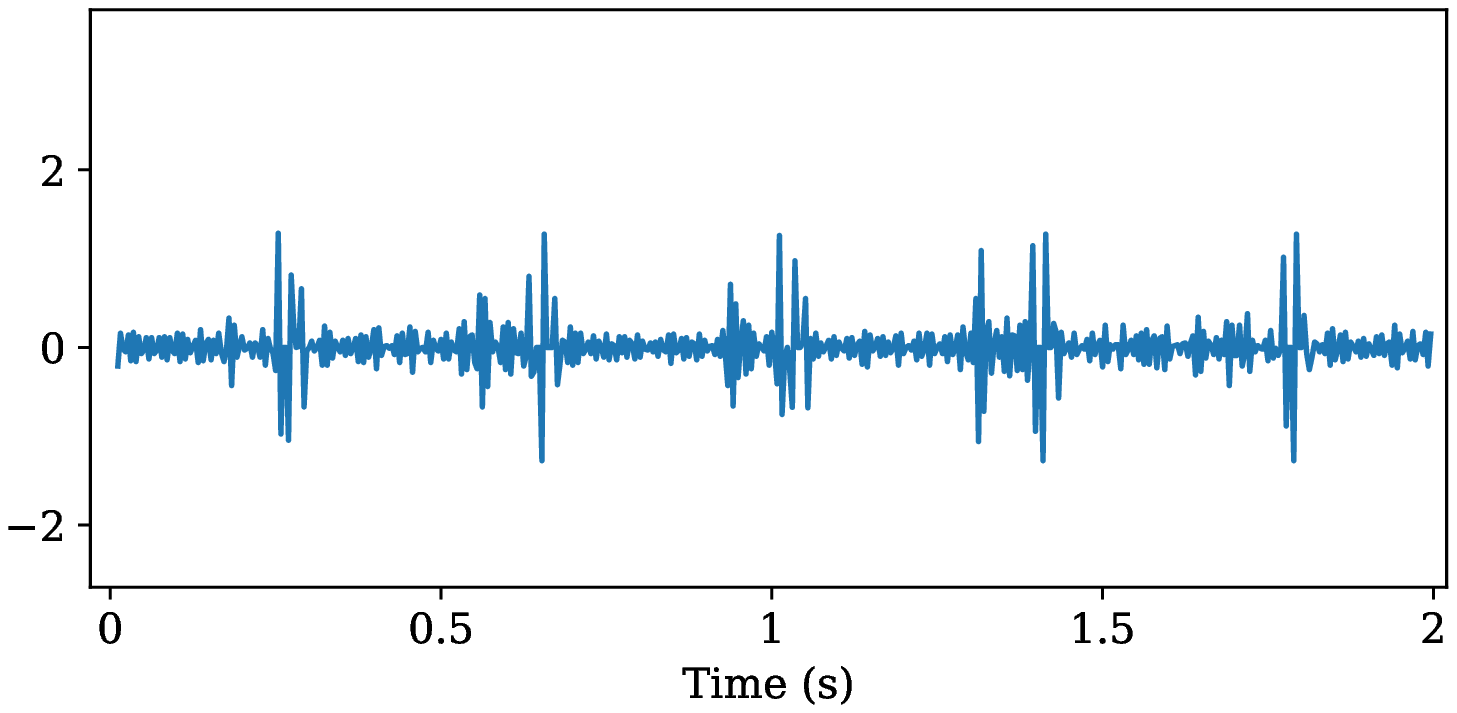"}
            \subcaption{Third-order differenced time series in group 2}
            \vspace{5mm}
        \end{minipage} \\
    \end{tabular}
    \caption{
        ECG time series. 
        (a) and (b) depict typical examples of time series in each group. 
        Third-order differencing of these time series yields (c) and (d).
        }
    \label{ECG}
\end{figure}

\subsubsection{ECG dataset}
The ECG dataset\footnote{\url{https://web.archive.org/web/20040209025257/http://www.physionet.org/physiobank/database/}} consists of $43$ time series of two-second ECG recordings.
These $43$ time series are divided into two groups: $13$ normal sinus rhythms (group 1) and $30$ supraventricular arrhythmias (group 2).
Typical examples of time series in each group are shown in Figs.~\ref*{ECG}(a) and (b).
As in Ref.~\cite{4782736}, we perform third-order differencing of the time series in order to reduce the nonstationarity.
The third-order differenced time series are shown in Figs.~\ref*{ECG}(c) and (d).

The BIC values for the MLGSSMs indicate that $M = 2$ and $d_x = 5$ are optimal for this dataset, as shown in Table \ref*{BIC}(a).
As shown in Table \ref*{CSM}, the proposed method produced clustering results that were as accurate as those for the MAR-based method used in Ref.~\cite{XIONG20041675}.
In the best case where the proposed method achieved a cluster similarity score of $0.97$, only one of the time series belonging to group 1 was incorrectly assigned to group 2, as shown in Table \ref*{confusion_matrix_population}(a).

\begin{figure}[t]

    \begin{tabular}{cc}

        \centering

        \begin{minipage}[c]{0.46\linewidth}
            \centering
            \includegraphics[width=75mm]{"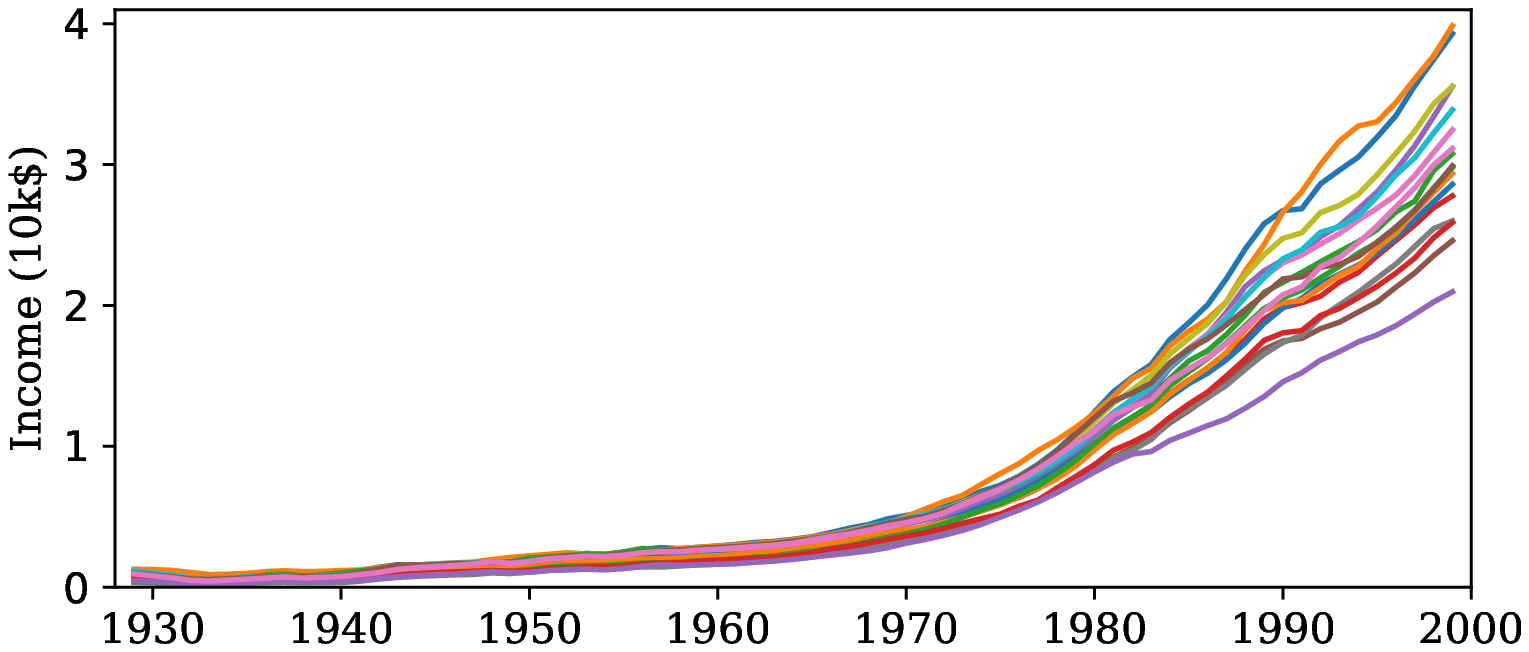"}
            \subcaption{Raw time series in group 1}
            \vspace{5mm}
        \end{minipage} &

        \begin{minipage}[c]{0.46\linewidth}
            \centering
            \includegraphics[width=75mm]{"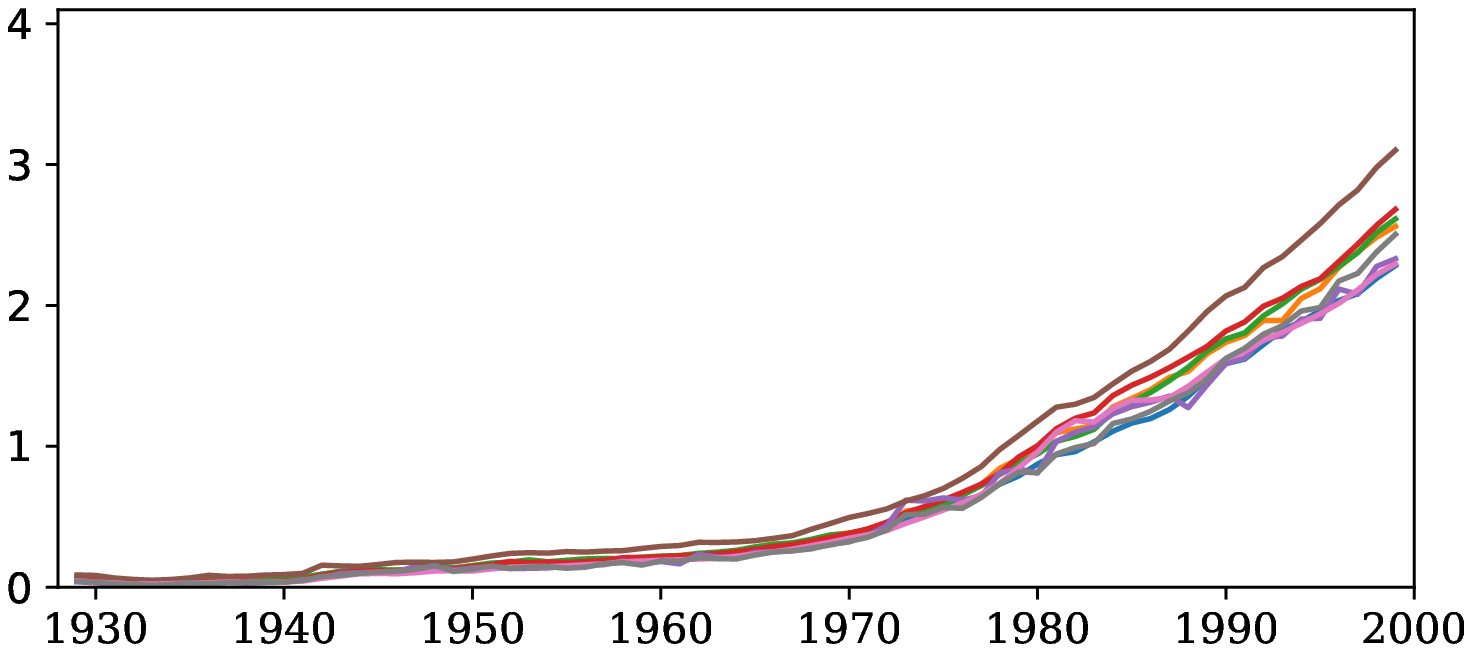"}
            \subcaption{Raw time series in group 2}
            \vspace{5mm}
        \end{minipage} \\

        \begin{minipage}[c]{0.46\linewidth}
            \centering
            \includegraphics[width=75mm]{"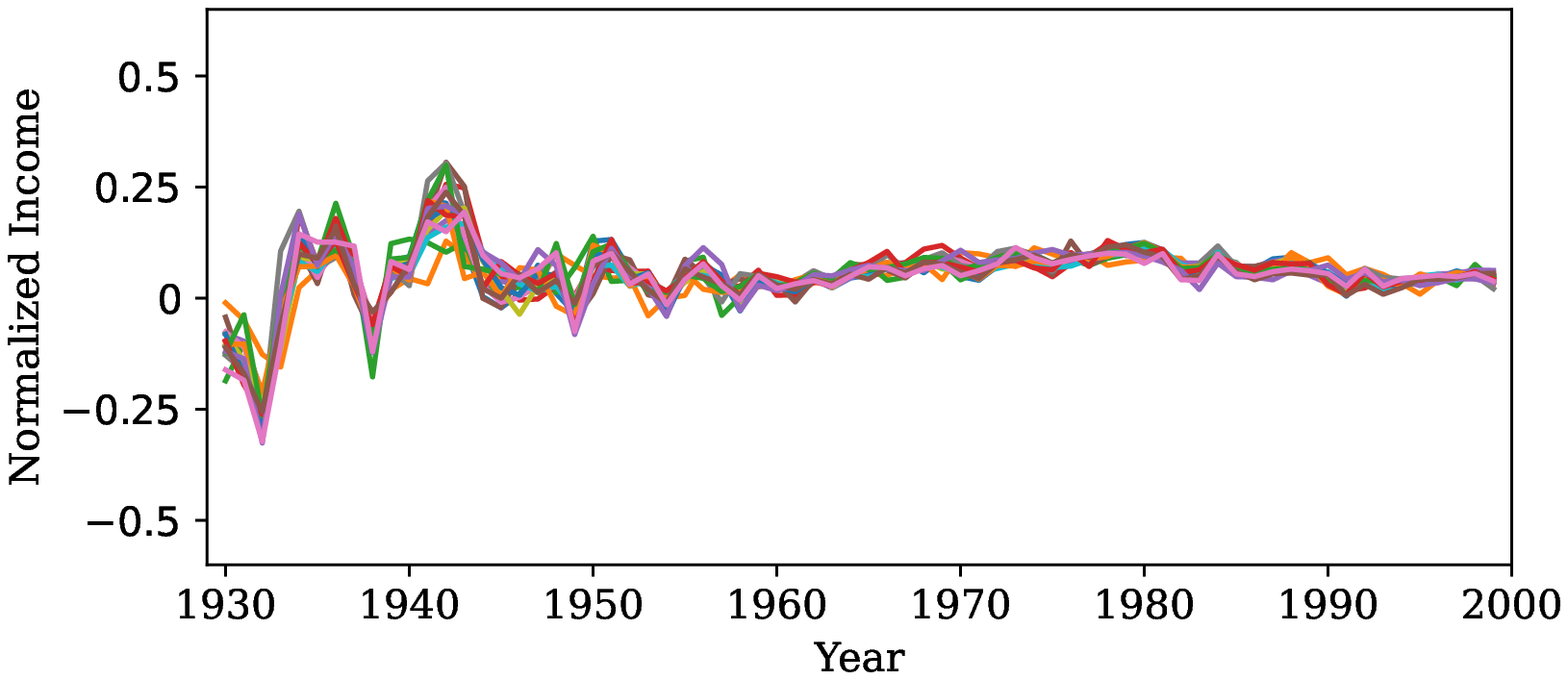"}
            \subcaption{Normalized time series in group 1}
            \vspace{5mm}
        \end{minipage} &

        \begin{minipage}[c]{0.46\linewidth}
            \centering
            \includegraphics[width=75mm]{"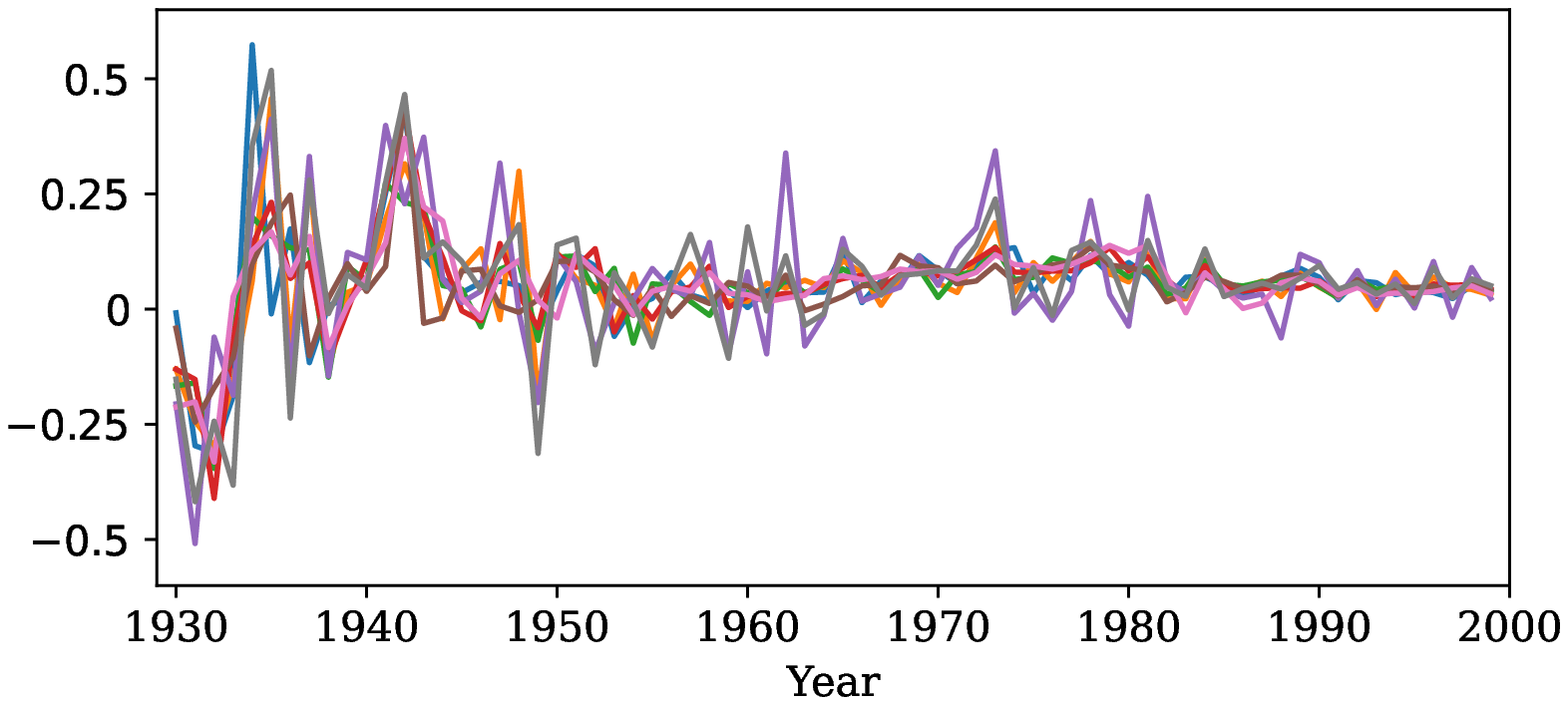"}
            \subcaption{Normalized time series in group 2}
            \vspace{5mm}
        \end{minipage} \\
    \end{tabular}
    \caption{
        Personal income time series. 
        (a) and (b) depict all time series in two groups. 
        Normalizing these time series yields (c) and (d).
        }

    \label{income}
\end{figure}

\subsubsection{Personal income dataset}
The personal income dataset\footnote{\url{https://www.bea.gov/data/income-saving/personal-income-by-state}} consists of $25$ time series of the average personal income in $25$ states of the USA during the period of $1929$--$1999$.
The $25$ states are divided into the two groups shown in Figs.~\ref*{income}(a) and (b).
Groups $1$ and $2$ consist of $17$ states with a high growth rate and $8$ states with a low growth rate, respectively.
As shown in Figs.~\ref*{income}(c) and (d), we log-transform the raw time series and then perform differencing, following Ref.~\cite{4782736}.

Table \ref*{BIC}(b) indicates that $M = 2$ and $d_x = 5$ are optimal for this dataset.
With these hyperparameter values, the proposed method produced clustering results that were more accurate than those obtained using the other methods, as shown in Table \ref*{CSM}.
As shown in Table \ref*{confusion_matrix_population}(b), the proposed method achieved perfect clustering in the best case.

\begin{figure}[t]

    \begin{tabular}{cc}

        \centering

        \begin{minipage}[c]{0.46\linewidth}
            \centering
            \includegraphics[width=75mm]{"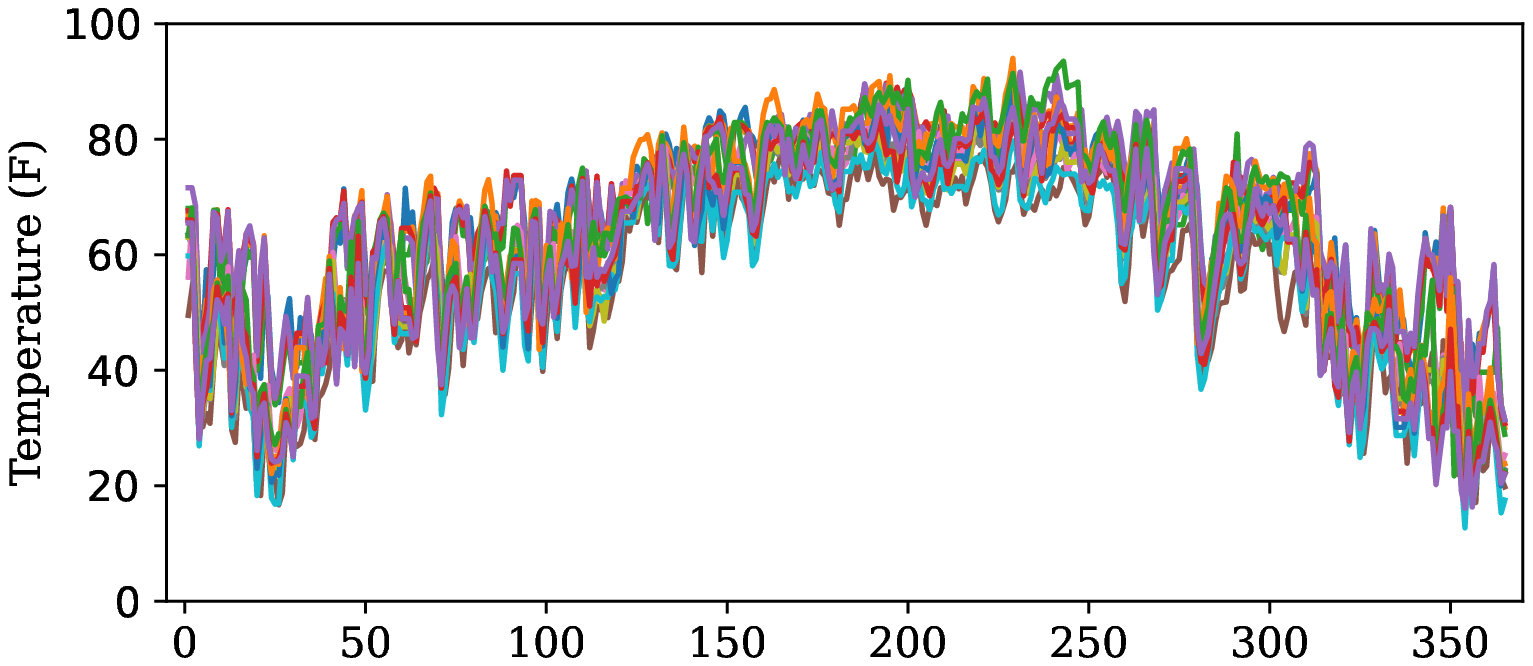"}
            \subcaption{Raw time series in group 1}
            \vspace{5mm}
        \end{minipage} &

        \begin{minipage}[c]{0.46\linewidth}
            \centering
            \includegraphics[width=75mm]{"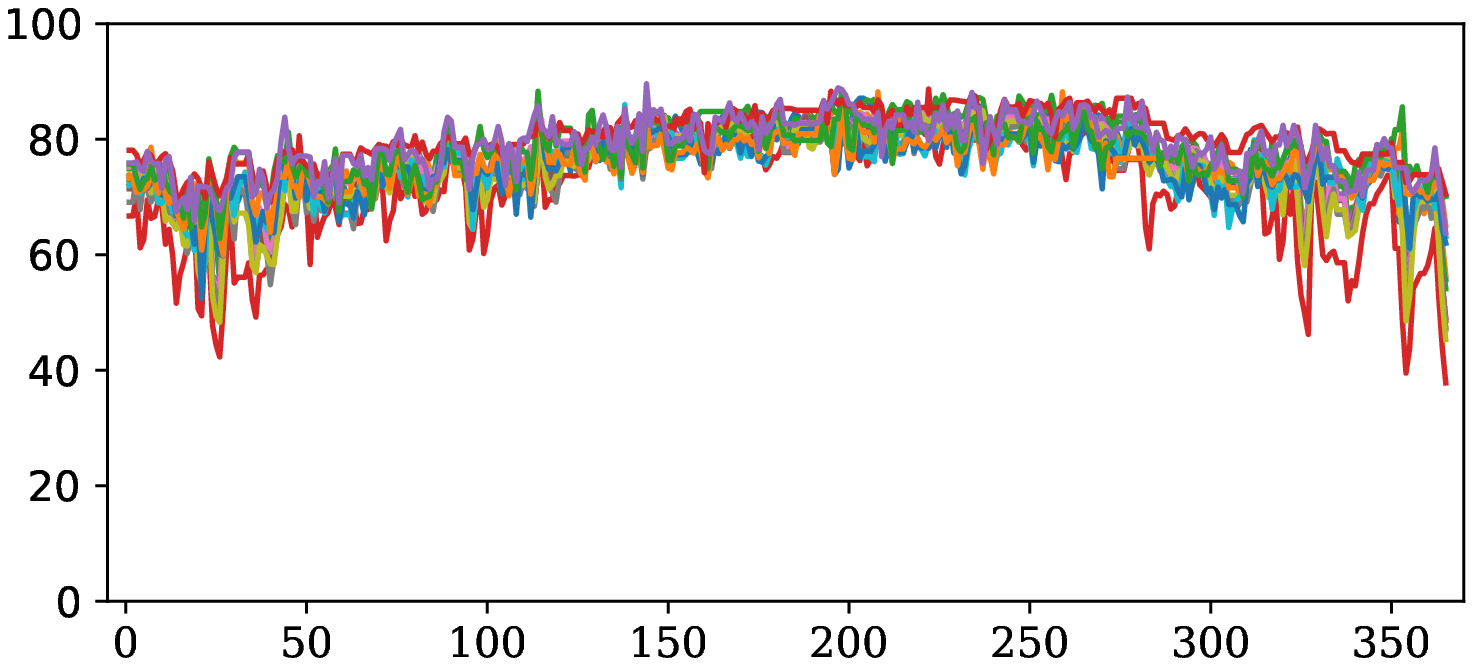"}
            \subcaption{Raw time series in group 2}
            \vspace{5mm}
        \end{minipage} \\

        \begin{minipage}[c]{0.46\linewidth}
            \centering
            \includegraphics[width=75mm]{"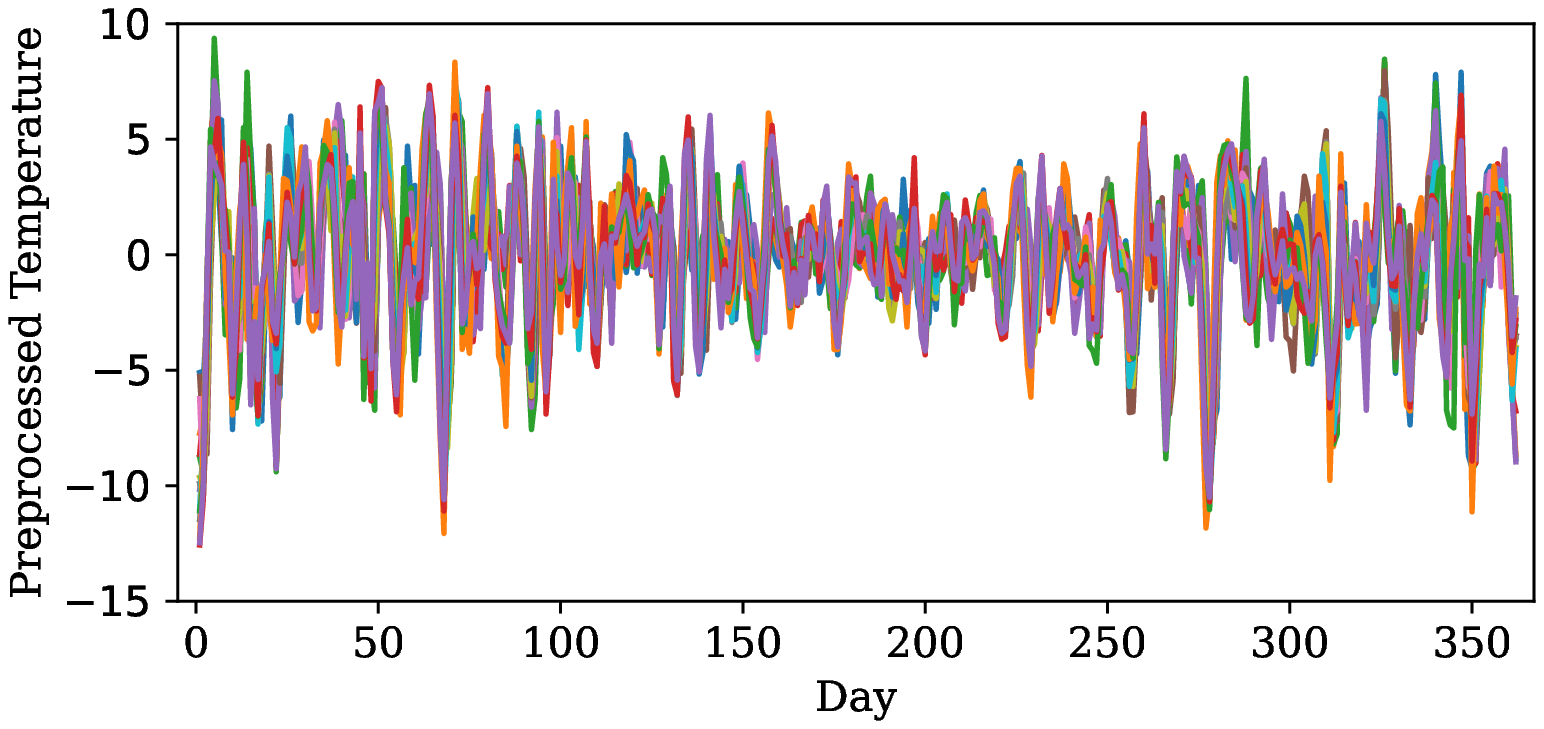"}
            \subcaption{Preprocessed time series in group 1}
            \vspace{5mm}
        \end{minipage} &

        \begin{minipage}[c]{0.46\linewidth}
            \centering
            \includegraphics[width=75mm]{"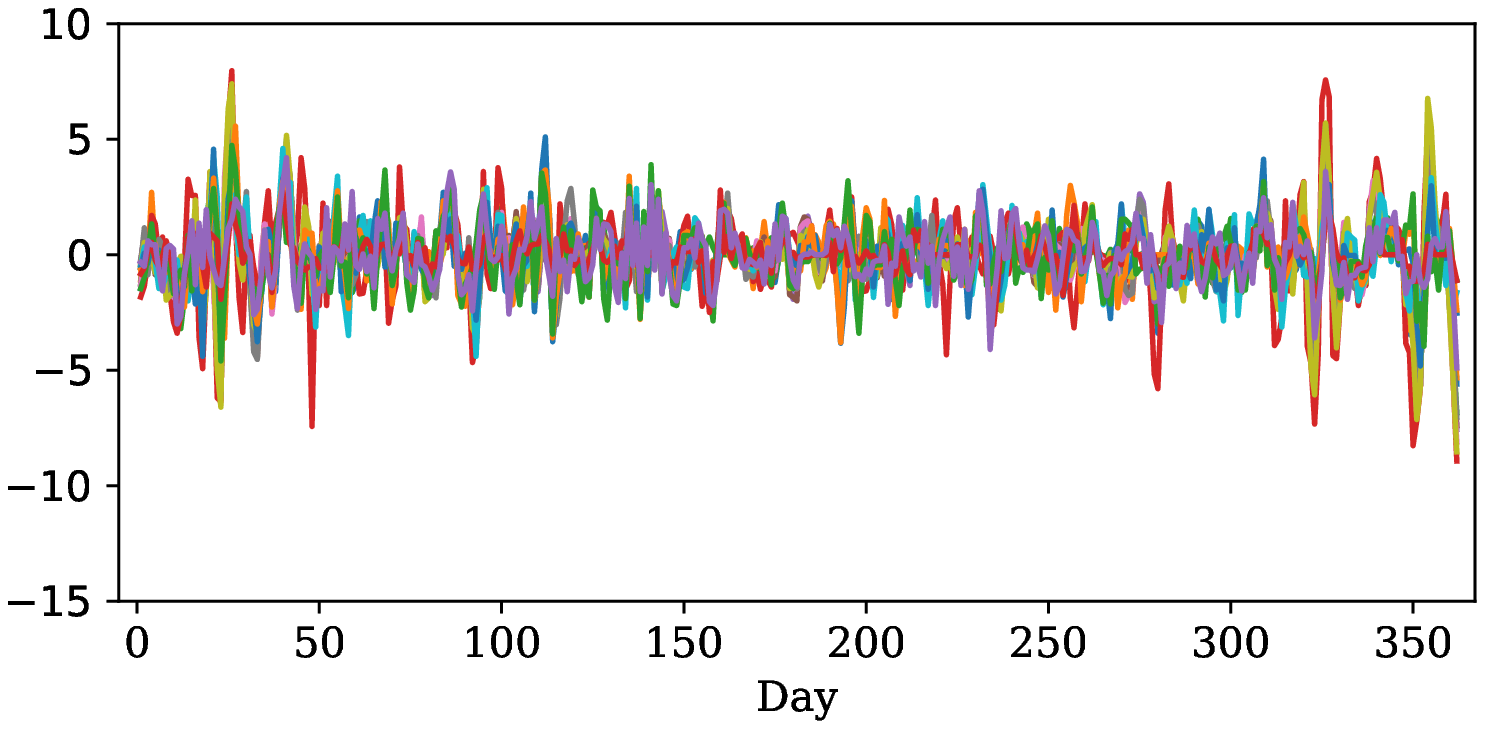"}
            \subcaption{Preprocessed time series in group 2}
            \vspace{5mm}
        \end{minipage} \\
    \end{tabular}
    \caption{
        Temperature time series. 
        (a) and (b) depict all time series in two groups. 
        Smoothing and differencing these time series yields (c) and (d).
        }

    \label{temperature}
\end{figure}

\subsubsection{Temperature dataset}
The temperature dataset\footnote{\url{https://web.archive.org/web/20040205004436/https://www.ncdc.noaa.gov/oa/climate/climatedata.html}} consists of $30$ time series of the daily average temperature recorded in Florida ($5$ locations in northern Florida and $9$ locations in southern Florida), Tennessee ($10$ locations), and Cuba ($6$ locations) in the year $2000$.
Because of the geographical proximity and similarity in temperature, northern Florida and Tennessee form group $1$, and southern Florida and Cuba form group $2$.
As in Ref.~\cite{4782736}, the time series are smoothed with a moving average window of size $3$ and are then differenced once.
Figure \ref*{temperature} shows the raw and preprocessed time series.

Table \ref*{BIC}(c) indicates that $M = 2$ and $d_x = 5$ are optimal for this dataset.
With these hyperparameter values, the proposed method produced clustering results that were as accurate or more accurate than those for the other methods, as shown in Table \ref*{CSM}.
As shown in Table \ref*{confusion_matrix_population}(c), the proposed method achieved perfect clustering in the best case.

\subsubsection{Population dataset}
The population dataset\footnote{\url{https://web.archive.org/web/20040220002039/https://eire.census.gov/popest/archives/state/st_stts.php}} consists of the $20$ population time series in $20$ states of the USA during the period of $1900$--$1999$. 
The $20$ states are divided into the two groups shown in Figs.~\ref*{fig4}(a) and (b). 
Group $1$ and group $2$ consist of $11$ states with exponential population trends and $9$ states with linear population trends, respectively. 

As shown in Figs.~\ref*{fig4}(a) and (b), the scale of the time series varies both within and between groups. 
To accurately capture the differences in trends in the populations, we normalize the time series in advance. 
Specifically, we log-transform the raw time series and then apply max-min normalization. 
Consequently, the values of the time series range from $0$ to $1$.
The normalized time series are shown in Figs.~\ref*{fig4}(c) and (d). 
As a result of normalization, the time series belonging to group 1 and those belonging to group 2 exhibit linear and logarithmic trends, respectively.

The BIC values for the MLGSSMs indicate that $M = 2$ and $d_x = 3$ are optimal for this dataset, as shown in Table \ref*{BIC}(d).
As shown in Table \ref*{CSM}, the proposed method produced clustering results that were as accurate or more accurate than those for the other methods. 
In particular, the proposed method significantly outperformed the MAR-based methods used in Refs.~\cite{XIONG20041675,4782736}.

Here we focus on the case in which the proposed method achieved the highest score.
Table \ref*{population_pred_param} shows the estimated parameter values obtained using our method for each cluster. 
Note that each observation matrix is fixed to $\B C^{(k)} = \left[ 1 \;\; 1 \;\; 1 \right]$ for $k = 1,\,2$. 
In the absence of noise (i.e., if we ignore the noise terms $\B w$, $\B v$, and $\B u$), the LGSSMs with the estimated parameter values generate the time series shown in Figs.~\ref*{fig4}(e) and (f). 
These predicted time series for cluster 1 and cluster 2 have a linear and a logarithmic trend, respectively. 
This suggests that, as expected, the proposed method performs clustering according to the trends.

As shown in Table \ref*{confusion_matrix_population}(d), two states were incorrectly assigned to group 1.
The (normalized) population time series of these two states, Michigan and New Jersey, are plotted in Figs.~\ref*{fig4}(g) and (h). 
Both time series exhibit weak logarithmic (or almost linear) trends.
It is difficult to distinguish between such a weak logarithmic trend and a linear trend, and, consequently, the proposed method fails to assign the two states to the correct group.

\begin{midpage}

\begin{table}[H]
    \caption{
        BIC values for MLGSSMs with different values of $M$ and $d_x$ for real datasets (rounded to two decimal places).
    }
    \begin{tabular}{cc}

        \begin{minipage}[c]{0.46\linewidth}
            \centering
            \subcaption{ECG}
            \begin{tabular}{|c|c|c|c|}
                \hline
                \diagbox{$d_x$}{$M$} & 2 & 3 & 4 \\ \hline
                2 & 273.72 & 264.38 & 253.45 \\ \hline
                3 & 272.30 & 263.79 & 253.40 \\ \hline
                4 & 272.24 & 261.98 & 256.37 \\ \hline
                5 & 273.99 & 262.86 & 250.01 \\ \hline
                6 & 273.09 & 266.19 & 254.82 \\ \hline
            \end{tabular}
            \vspace{5mm}
        \end{minipage}

        &

        \begin{minipage}[c]{0.46\linewidth}
            \centering
            \subcaption{Personal income}
            \begin{tabular}{|c|c|c|c|}
                \hline
                \diagbox{$d_x$}{$M$} & 2 & 3 & 4 \\ \hline
                2 & 125.27 & 115.61 & 106.51 \\ \hline
                3 & 125.57 & 117.08 & 109.99 \\ \hline
                4 & 129.24 & 119.43 & 108.74 \\ \hline
                5 & 131.66 & 120.79 & 110.95 \\ \hline
                6 & 127.75 & 117.30 & 113.08 \\ \hline
            \end{tabular}
            \vspace{5mm}
        \end{minipage}
    
        \\

        \begin{minipage}[c]{0.46\linewidth}
            \centering
            \subcaption{Temperature}
            \begin{tabular}{|c|c|c|c|}
                \hline
                \diagbox{$d_x$}{$M$} & 2 & 3 & 4 \\ \hline
                2 & 209.91 & 196.72 & 189.62 \\ \hline
                3 & 217.74 & 208.06 & 197.04 \\ \hline
                4 & 217.41 & 208.15 & 198.20 \\ \hline
                5 & 217.64 & 208.12 & 198.00 \\ \hline
                6 & 217.70 & 207.67 & 197.72 \\ \hline
            \end{tabular}
        \end{minipage}

        &

        \begin{minipage}[c]{0.46\linewidth}
            \centering
            \subcaption{Population}
            \begin{tabular}{|c|c|c|c|}
                \hline
                \diagbox{$d_x$}{$M$} & 2 & 3 & 4 \\ \hline
                2 & 101.29 & 86.29 & 118.42 \\ \hline
                3 & 143.18 & 135.23 & 125.53 \\ \hline
                4 & 140.86 & 132.04 & 125.66 \\ \hline
                5 & 141.14 & 132.43 & 125.40 \\ \hline
                6 & 139.96 & 134.59 & 127.01 \\ \hline
            \end{tabular}
        \end{minipage}
        
        \\
    \end{tabular}
    \label{BIC}
\end{table}

\begin{table}[H]
    \centering
    \caption{
        Cluster similarity scores obtained by different clustering methods for real datasets (rounded to two decimal places).
        For comparison, the following methods are cited: the EM algorithm for MARs (EMMAR) \cite{XIONG20041675}, the VB algorithm for MARs (VBMAR) \cite{4782736}, and the Euclidean distance between the linear predictive coding cepstra (CEP) \cite{989529}, the discrete Fourier transforms (DFT) \cite{Agrawal1993}, and the discrete wavelet transforms (DWT) \cite{795160} of two time series.
        For the proposed method, the mean score of $10$ trials is shown for each dataset. 
        The values in parentheses are standard deviations.
        }
    \begin{tabular}{|lcccccc|}
        \hline
        Dataset & MLGSSM (ours) & EMMAR \cite{XIONG20041675} & VBMAR \cite{4782736} & CEP \cite{989529} & DFT \cite{Agrawal1993} & DWT \cite{795160} \\
        \hline
        ECG & $0.95(0.03)$ & $0.94$ & - & - & - & - \\
        Personal income & $0.96(0.04)$ & $0.91$ & $0.92$ & $0.84$ & $0.75$ & $0.74$ \\
        Temperature & $0.98(0.02)$ & $1.00$ & $1.00$ & $0.93$ & $0.83$ & $0.82$ \\
        Population & $0.82(0.05)$ & $0.65$ & $0.76$ & $0.74$ & $0.65$ & $0.80$ \\
        \hline
    \end{tabular}
    \label{CSM}
\end{table}

\end{midpage}

\newpage

\begin{midpage}

\begin{table}[h]
    \caption{
        Confusion matrices obtained by clustering real datasets with MLGSSM at the highest cluster similarity scores.
    }
    \begin{tabular}{cc}

        \begin{minipage}[c]{0.46\linewidth}
            \centering
            \subcaption{ECG}
            \begin{tabular}{|c|c|c|} \hline
                \diagbox{True}{Pred} & group 1 & group 2 \\ \hline
                group 1 & 12 & 1 \\ \hline
                group 2 & 0 & 30 \\ \hline
            \end{tabular}
            \vspace{5mm}
        \end{minipage}
 
        &
       
        \begin{minipage}[c]{0.46\linewidth}
            \centering
            \subcaption{Personal income}
            \begin{tabular}{|c|c|c|} \hline
                \diagbox{True}{Pred} & group 1 & group 2 \\ \hline
                group 1 & 17 & 0 \\ \hline
                group 2 & 0 & 8 \\ \hline
            \end{tabular}
            \vspace{5mm}
        \end{minipage}
        
        \\
      
        \begin{minipage}[c]{0.46\linewidth}
            \centering
            \subcaption{Temperature}
            \begin{tabular}{|c|c|c|} \hline
                \diagbox{True}{Pred} & group 1 & group 2 \\ \hline
                group 1 & 15 & 0 \\ \hline
                group 2 & 0 & 15 \\ \hline
            \end{tabular}
        \end{minipage}
        
        &
   
        \begin{minipage}[c]{0.46\linewidth}
            \centering
            \subcaption{Population}
            \begin{tabular}{|c|c|c|} \hline
                \diagbox{True}{Pred} & group 1 & group 2 \\ \hline
                group 1 & 11 & 0 \\ \hline
                group 2 & 2 & 7 \\ \hline
            \end{tabular}
        \end{minipage}
        
        \\
    \end{tabular}
    \label{confusion_matrix_population}
\end{table}

\begin{table}[H]
    \centering
    \caption{
        Estimated parameter values for MLGSSM with $M = 2$ and $d_x = 3$ for population dataset (rounded to three decimal places).
        }
    \begin{tabular}{|c||cccc|}
        \hline
        $M$ & \rule[0mm]{0mm}{4mm} & $\B A$ & $\B \Gamma$ & $\B \Sigma$ \\
        \hline
        \multirow{7}{*}{1}
        & \rule[-7mm]{0mm}{16mm} &
        ${
            \begin{bmatrix*}[r]
                0.109 & 0.611 & -0.007 \\
                0.584 & 0.253 & -0.240 \\
                0.180 & 0.012 & 1.104 \\
            \end{bmatrix*}
        }$ 
        &
        ${
            \begin{bmatrix*}[r]
                0.023 & -0.006 & -0.017 \\
                -0.006 & 0.024 & -0.017 \\
                -0.017 & -0.017 & 0.034 \\
            \end{bmatrix*}
        }$ 
        &
        0.000 
        \\
        \cline{2-5}
        & \rule[0mm]{0mm}{4mm} & $\B \mu$ & $\B P$ & $p(\omega)$ \\
        \cline{2-5} 
        & \rule[-7mm]{0mm}{16mm} &
        ${
            \begin{bmatrix*}[r]
                0.424 \\
                0.429 \\
                -0.854 \\
            \end{bmatrix*}
        }$
        &
        ${
            \begin{bmatrix*}[r]
                0.118 & 0.037 & -0.156 \\
                0.037 & 0.099 & -0.137 \\
                -0.156 & -0.137 & 0.292 \\
            \end{bmatrix*}
        }$
        &
        0.650
        \\
        \hline
        $M$ & \rule[0mm]{0mm}{4mm} & $\B A$ & $\B \Gamma$ & $\B \Sigma$ \\
        \hline
        \multirow{7}{*}{2}
        & \rule[-7mm]{0mm}{16mm} &
        ${
            \begin{bmatrix*}[r]
                0.283 & 0.684 & 0.398 \\
                1.307 & 1.539 & 0.588 \\
                -0.912 & -1.556 & -0.279 \\
            \end{bmatrix*}
        }$ 
        &
        ${
            \begin{bmatrix*}[r]
                0.019 & -0.003 & -0.017 \\
                -0.003 & 0.059 & -0.058 \\
                -0.017 & -0.058 & 0.076 \\
            \end{bmatrix*}
        }$ 
        &
        0.000 
        \\
        \cline{2-5}
        & \rule[0mm]{0mm}{4mm} & $\B \mu$ & $\B P$ & $p(\omega)$ \\
        \cline{2-5} 
        & \rule[-7mm]{0mm}{16mm} &
        ${
            \begin{bmatrix*}[r]
                -0.068 \\
                -1.192 \\
                1.259 \\
            \end{bmatrix*}
        }$
        &
        ${
            \begin{bmatrix*}[r]
                0.042 & 0.023 & -0.064 \\
                0.023 & 0.589 & -0.612 \\
                -0.064 & -0.612 & 0.677 \\
            \end{bmatrix*}
        }$
        &
        0.350
        \\
        \hline
    \end{tabular}
    \label{population_pred_param}
\end{table}

\end{midpage}

\newpage

\begin{midpage}

\begin{figure}[H]

    \begin{tabular}{cc}

        \centering

        \begin{minipage}[c]{0.46\linewidth}
            \centering
            \includegraphics[width=75mm]{"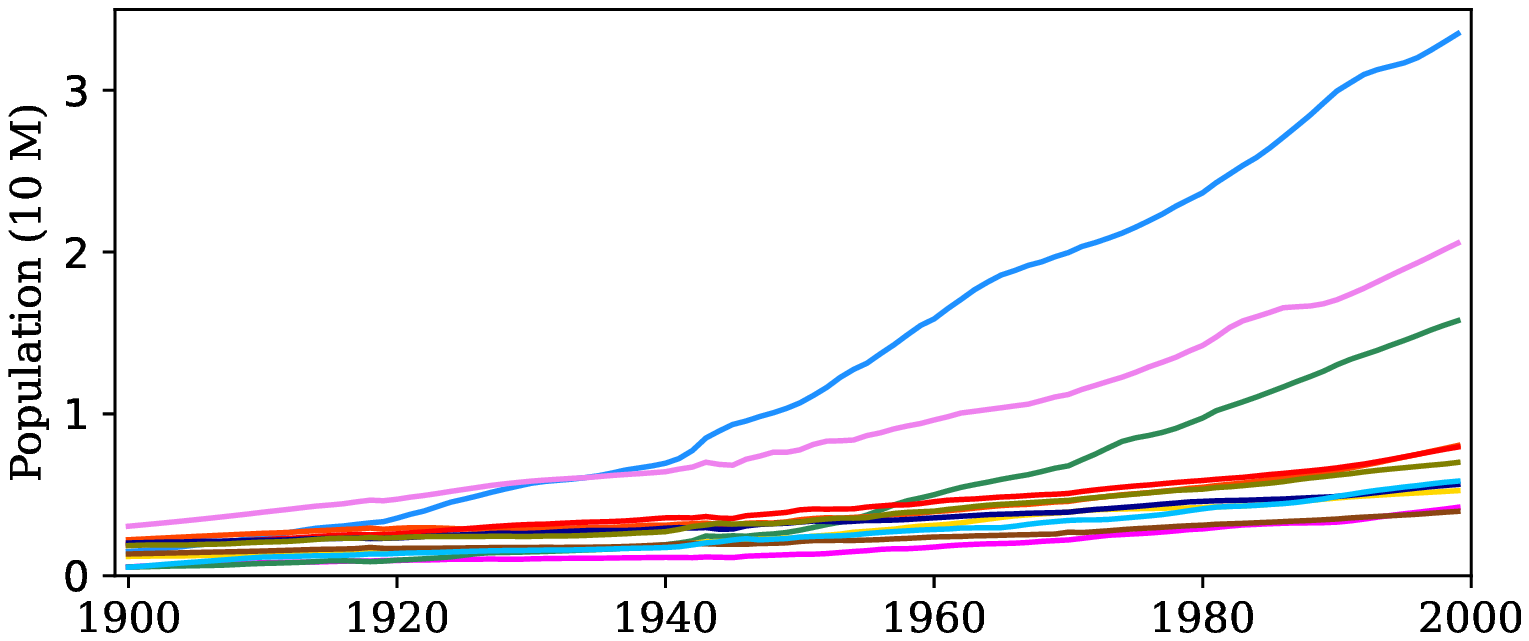"}
            \subcaption{Raw time series in group 1}
            \label{raw_group1}
            \vspace{5mm}
        \end{minipage} &

        \begin{minipage}[c]{0.46\linewidth}
            \centering
            \includegraphics[width=75mm]{"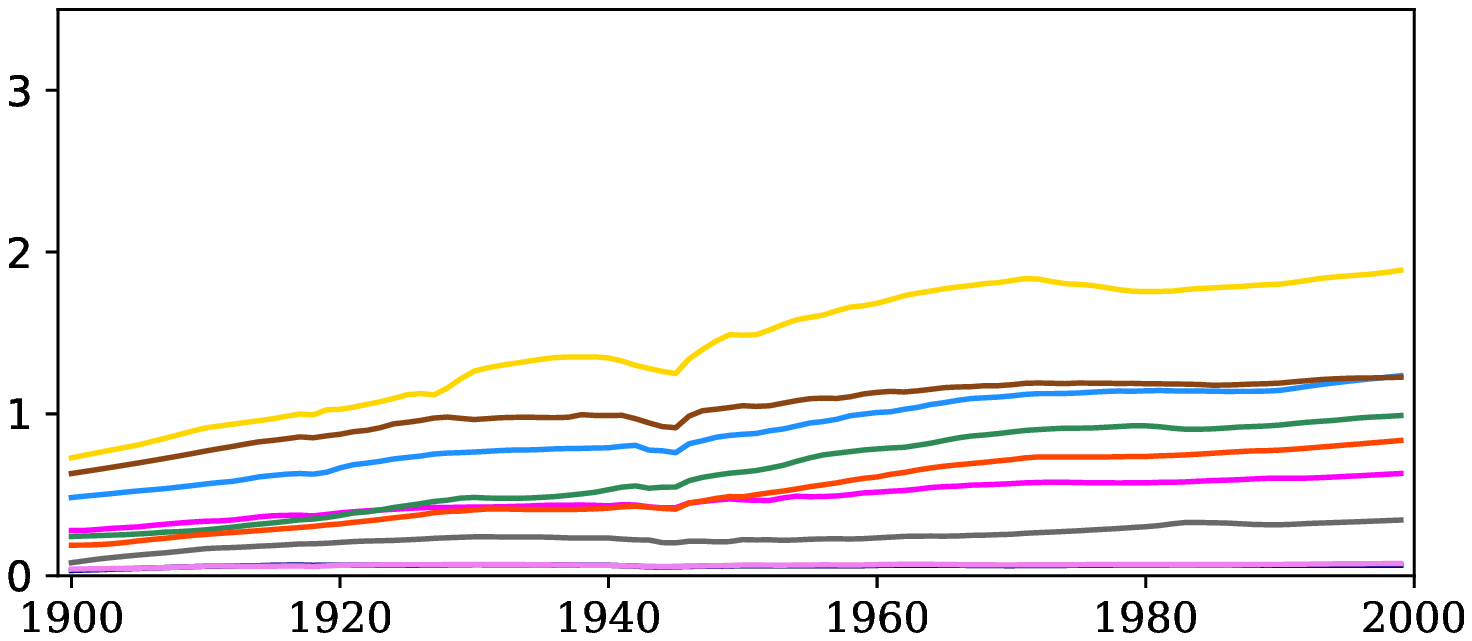"}
            \subcaption{Raw time series in group 2}
            \label{raw_group2}
            \vspace{5mm}
        \end{minipage} \\

        \begin{minipage}[c]{0.46\linewidth}
            \centering
            \includegraphics[width=75mm]{"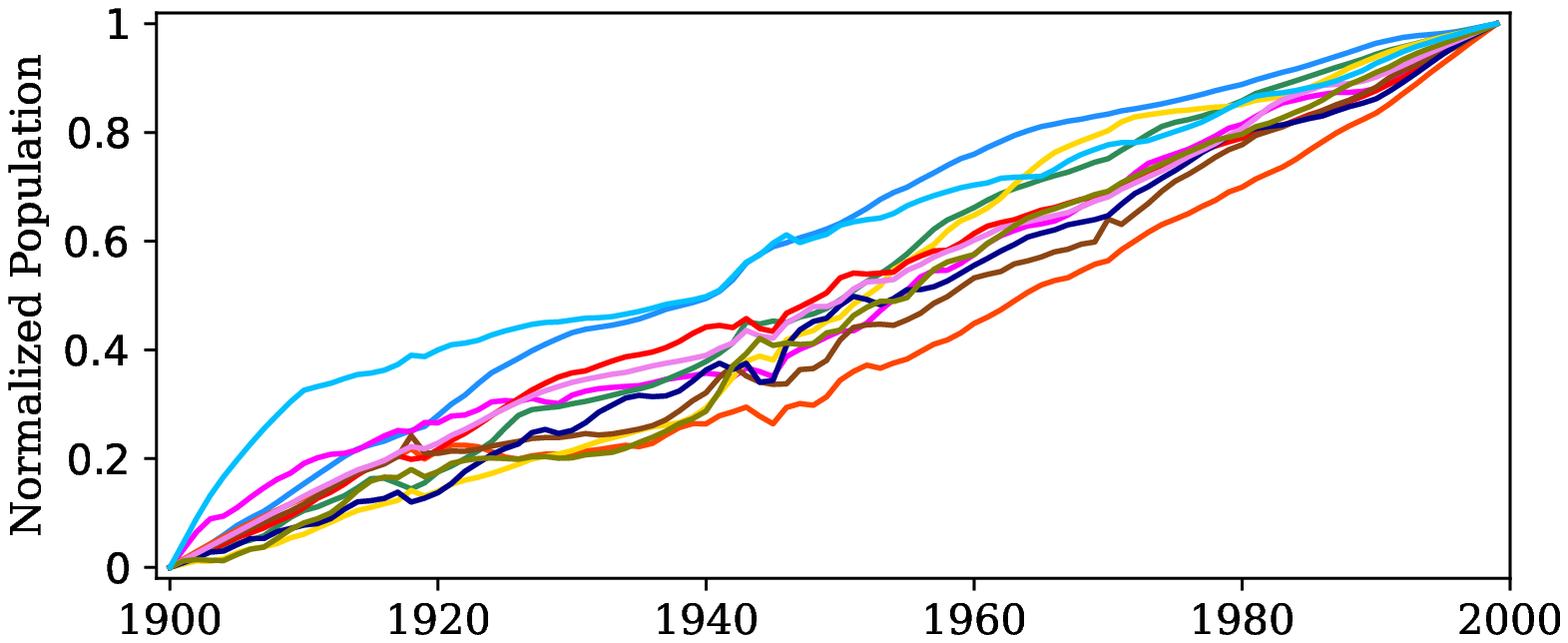"}
            \subcaption{Normalized time series in group 1}
            \vspace{5mm}
        \end{minipage} &

        \begin{minipage}[c]{0.46\linewidth}
            \centering
            \includegraphics[width=75mm]{"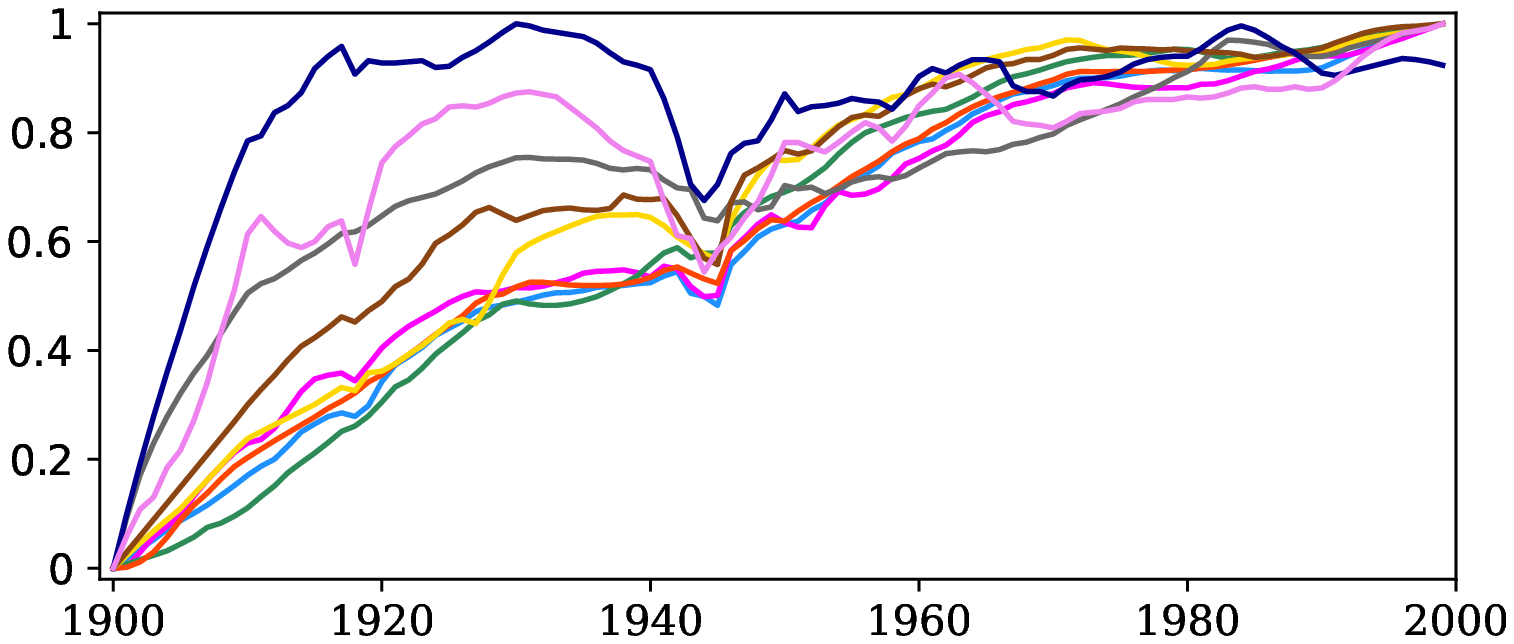"}
            \subcaption{Normalized time series in group 2}
            \vspace{5mm}
        \end{minipage} \\

        \begin{minipage}[c]{0.46\linewidth}
            \centering
            \includegraphics[width=75mm]{"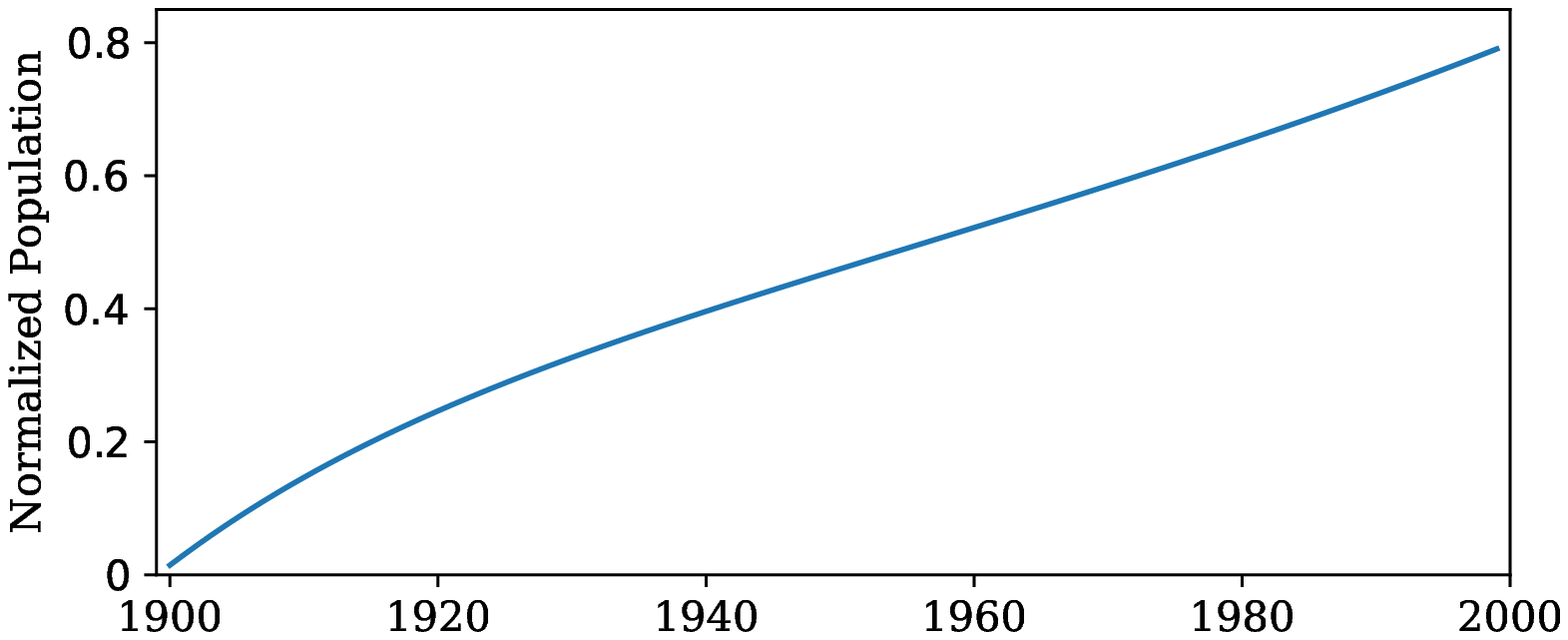"}
            \subcaption{Predicted time series without noise for cluster 1}
            \vspace{5mm}
        \end{minipage} &

        \begin{minipage}[c]{0.46\linewidth}
            \centering
            \includegraphics[width=75mm]{"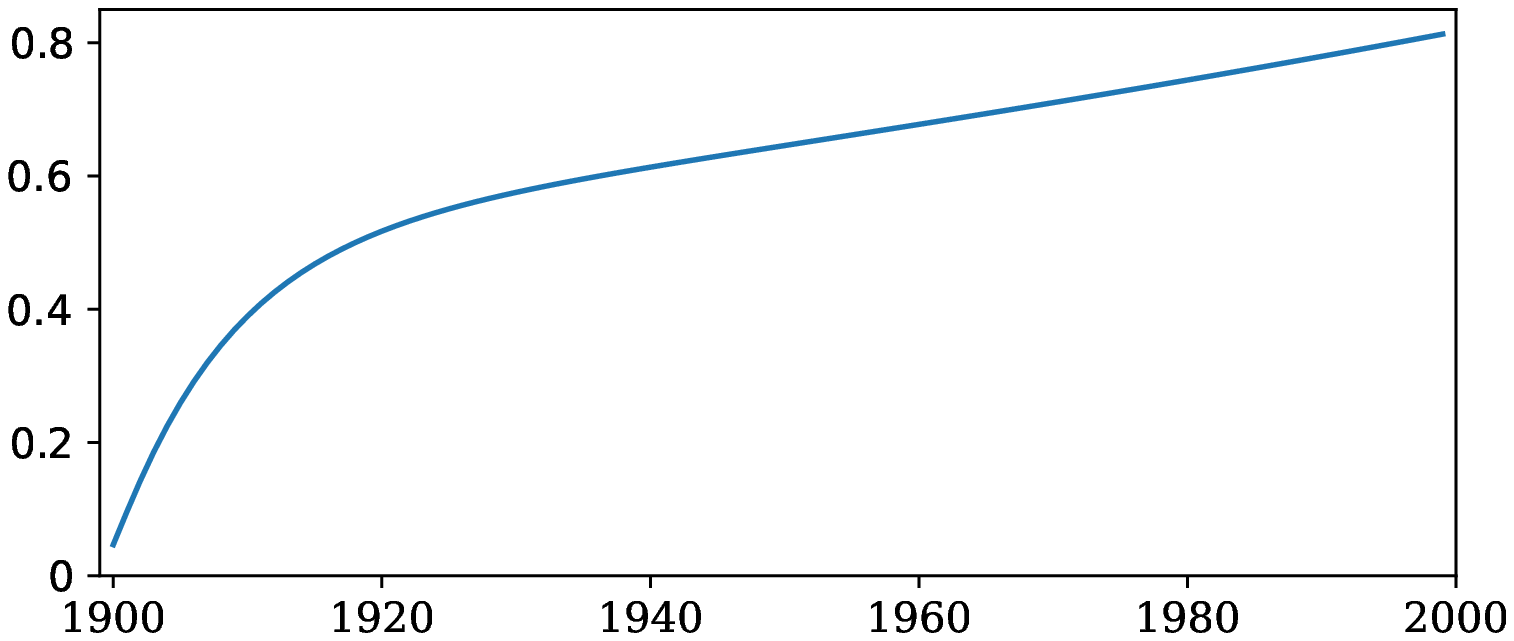"}
            \subcaption{Predicted time series without noise for cluster 2}
            \vspace{5mm}
        \end{minipage} \\

        \begin{minipage}[c]{0.46\linewidth}
            \centering
            \includegraphics[width=75mm]{"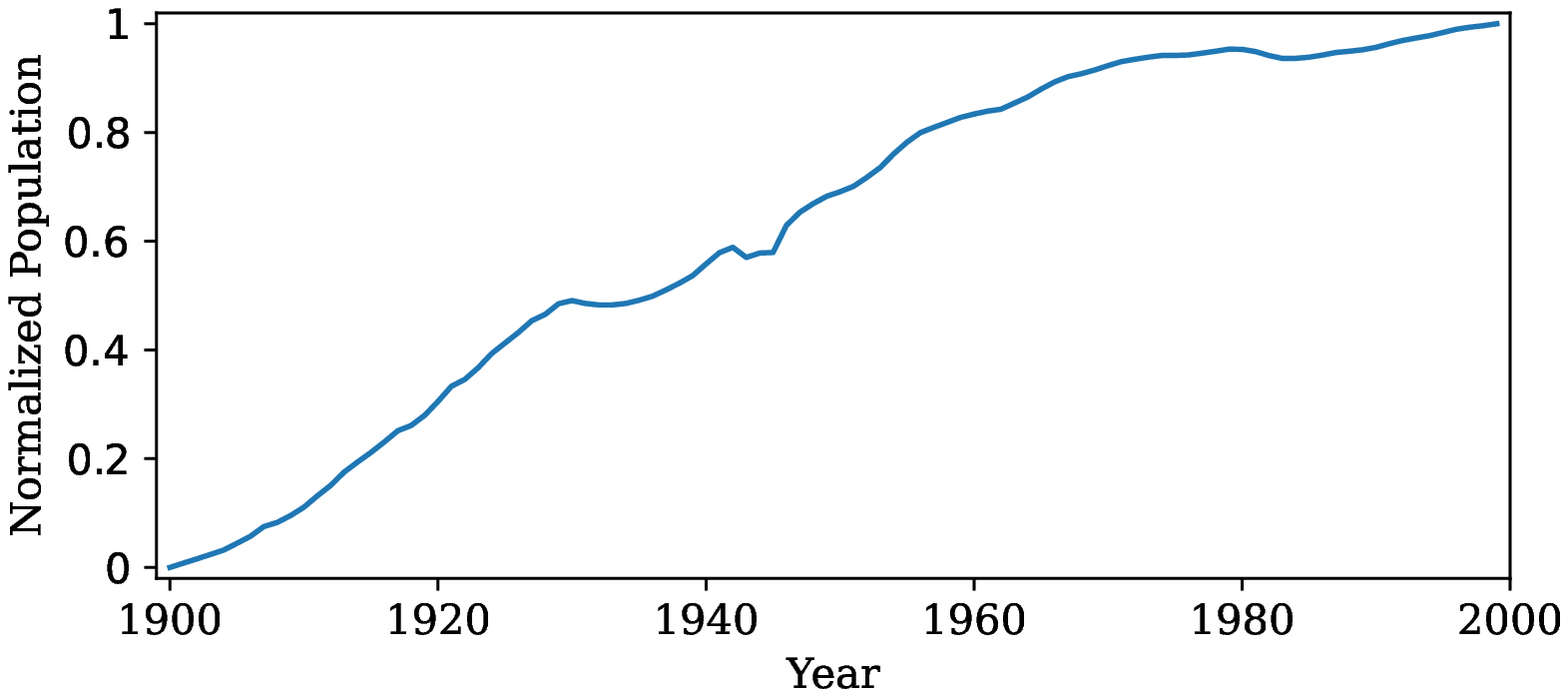"}
            \subcaption{Normalized population time series for Michigan}
        \end{minipage} &

        \begin{minipage}[c]{0.46\linewidth}
            \centering
            \includegraphics[width=75mm]{"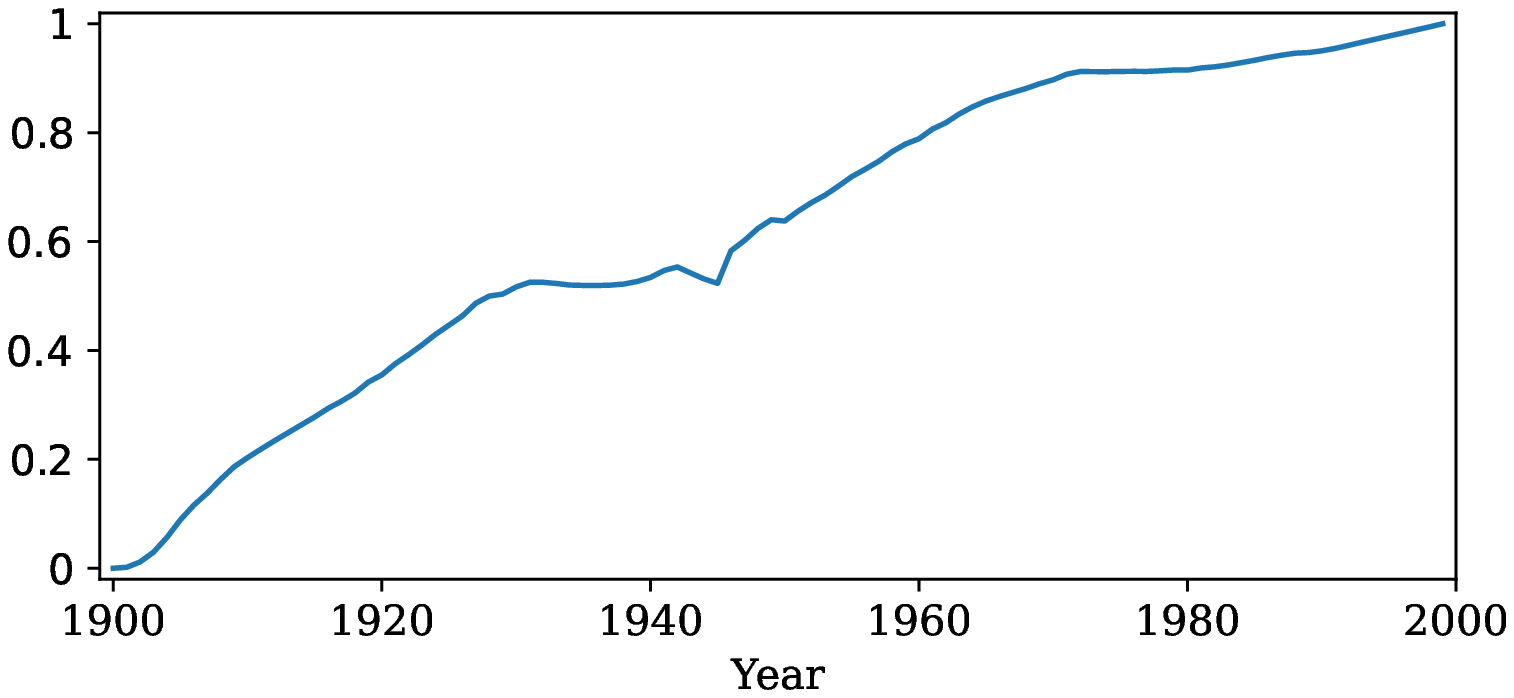"}
            \subcaption{Normalized population time series for New Jersey}
        \end{minipage} \\
    \end{tabular}
    \caption{
        Population time series. 
        (a) and (b) depict all time series in two groups. 
        Normalizing these time series yields (c) and (d).
        (e) and (f) depict the predicted time series obtained using the estimated parameters except $\B \Gamma$, $\B \Sigma$, and $\B P$. 
        In this experiment, clustering fails for two time series, (g) and (h).
        }
    \label{fig4}
\end{figure}

\end{midpage}

\newpage

\section{Conclusion}

In this paper, we proposed a novel model-based time series clustering method with mixtures of linear Gaussian state space models (MLGSSMs). 
As shown in Fig.~\ref*{fig1}, this method enables us not only to cluster a set of individual time series, but also to describe each cluster with an LGSSM\@. 
We can expect the method to provide accurate clustering results in many cases insofar as the LGSSMs have sufficient flexibility to describe the dynamics in various time series. 
In our method, the EM algorithm for MLGSSMs is used to estimate the model parameters.
Once the optimal parameter values are obtained, we assign each time series to a cluster such that the posterior probability of belonging to the cluster is a maximum. 
In addition, we can determine the hyperparameters such as the number of clusters using the BIC\@. 
Experiments on simulated datasets showed that clustering, parameter estimation, and model selection are performed properly. 
Furthermore, applications to real datasets indicated that the proposed method outperforms previous methods in terms of clustering accuracy.

The main contribution of this paper is that we propose a time series clustering method based on a highly flexible model. 
The flexibility of the MLGSSM allows us to perform accurate clustering even if a given dataset contains non-stationary time series.

As explained in Sec.~1, the model-based approach has the attractive advantage of being able to predict the future using the estimated model.
Our method is expected to be particularly useful for such prediction because of the flexibility of LGSSMs. 
Applying our method to forecasting problems offers an interesting opportunity for further investigation.

Several ideas to extend an MLGSSM to a more flexible model hold promise. 
The first is to incorporate exogenous variables into the LGSSM (\ref*{Eq:10}) as follows \cite{Chen1998}:
\begin{subequations}
    \renewcommand{\theequation}{
    \theparentequation-\alph{equation}  
    }
    \begin{align}
        \B x[t] &= \B A^{(k)} \B x[t-1] + \B B^{(k)} \B u[t] + \B w[t] ,  \\
        \B y[t] &=\B C^{(k)} \B x[t] + \B D^{(k)} \B u[t] + \B v[t] , \\
        \B x[1] &= \B \mu^{(k)} + \B u ,
    \end{align}
\end{subequations}
where $\B u[t]$, $\B B$, and $\B D$ are the $d_u$-dimensional input vector (corresponding to the exogenous variables), the ($d_x \times d_u$) input-to-state matrix, and the ($d_y \times d_u$) input-to-observation matrix, respectively. 
This will increase the predictive power of the model. 
The second is to remove the linear-Gaussian assumption as follows \cite{Kitagawa1987,TANIZAKI1998263}:
\begin{subequations}
    \renewcommand{\theequation}{
    \theparentequation-\alph{equation}  
    }
    \begin{align}
        \B x[t] &= F^{(k)}(\B x[t-1], \B w[t]) , \\
        \B y[t] &= H^{(k)}(\B x[t], \B v[t]) ,
    \end{align}
\end{subequations}
where $F$ and $H$ are parameterized nonlinear functions and $\B w_t$ and $\B v_t$ are random variables with certain non-Gaussian distributions. 
This will allow the models to capture more complicated dynamics. 
These extensions of MLGSSMs are left for future work.

\section*{Acknowledgements}

This work was supported by Osaka Gas Co., Ltd.

\appendix

\section{Derivation of ECDLL of MLGSSM}

We can obtain Eq.~(\ref*{Eq:16}) as below. 
As mentioned in Sec.~4, the ECDLL of an MLGSSM is defined as
\begin{align}
    \label{Eq:A1}
    \mathcal{Q}\bigl(\B \Theta \mid \B \Theta(s)\bigr)
    = \E_{\B Z,\,\B D_X \mid \B D_Y,\,\B \Theta(s)} \Bigl[ \,\log\,{p\bigl(\B Y_i,\,\B X_i \mid \B \theta^{(k)}\bigr)}\, \Bigr] .
\end{align}
Due to the relation
\begin{align}
    p\bigl(\B Z,\,\B D_X \mid \B D_Y,\,\B \Theta(s)\bigr) = p\bigl(\B Z \mid \B D_Y,\,\B \Theta(s)\bigr) p\bigl(\B D_X \mid \B Z,\,\B D_Y,\,\B \Theta(s)\bigr) ,
\end{align}
Eq.~(\ref*{Eq:A1}) can be rewritten as
\begin{align}
    \label{Eq:A2}
    \mathcal{Q}\bigl(\B \Theta \mid \B \Theta(s)\bigr)
    = \E_{\B Z \mid \B D_Y,\,\B \Theta(s)} \Bigl[ \,\E_{\B D_X \mid \B Z,\,\B D_Y,\,\B \Theta(s)} \Bigl[ \, \log\,{p\bigl(\B Y_i,\,\B X_i \mid \B \theta^{(k)}\bigr)} \,\Bigr] \,\Bigr].
\end{align}
By substituting Eq.~(\ref*{Eq:15}) into Eq.~(\ref*{Eq:A2}), we obtain
\begin{align}
    \label{Eq:A3}
    \mathcal{Q}\bigl(\B \Theta \mid \B \Theta(s)\bigr)
    &= \E_{\B Z \mid \B D_Y,\,\B \Theta(s)} \Bigl[ \,\E_{\B D_X \mid \B Z,\,\B D_Y,\,\B \Theta(s)} \Bigl[ \,\sum_{i=1}^{N} \log\,{p\bigl(\B Y_i,\,\B X_i \mid \omega^{(\B z_i)},\,\B \theta^{(\B z_i)}\bigr)}
     + \sum_{i=1}^{N} \log\,{p^{(\B z_i)}} \,\Bigr] \,\Bigr] \no \\
    &= \E_{\B Z \mid \B D_Y,\,\B \Theta(s)} \Bigl[ \,\sum_{i=1}^{N} \Bigl( \E_{\B X_i \mid \B Z,\,\B Y_i,\,\B \Theta(s)} \Bigl[ \,\log\,{p\bigl(\B Y_i,\,\B X_i \mid \omega^{(\B z_i)},\,\B \theta^{(\B z_i)}\bigr)} \,\Bigr] + \log\,{p^{(\B z_i)}} \Bigr) \,\Bigr] .
\end{align}
Because the posterior probability is defined as
\begin{align}
    p\bigl(\B Z \mid \B D_Y,\,\B \Theta(s)\bigr)
    \equiv p\bigl(\omega^{(\B z_1)},\;\dotsc\,,\;\omega^{(\B z_N)}\mid\B D_Y,\,\B \Theta(s)\bigr) 
    = \prod_{i=1}^{N} p\bigl(\omega^{(\B z_i)} \mid \B D_Y,\,\B \Theta(s)\bigr), 
\end{align}
we can rewrite Eq.~(\ref*{Eq:A3}) as
\begin{align}
    \mathcal{Q}\bigl(\B \Theta \mid \B \Theta(s)\bigr)
    =& \sum_{\B z_1 = 1}^{M} \cdots \sum_{\B z_N = 1}^{M}p\bigl(\omega^{(\B z_1)},\;\dotsc\,,\;\omega^{(\B z_N)}\mid\B D_Y,\,\B \Theta(s)\bigr) \no \\
    & \times \sum_{i=1}^{N} \Bigl( \,\E_{\B X_i \mid \B Z,\,\B Y_i,\,\B \Theta(s)} \Bigl[ \,\log\,{p\bigl(\B Y_i,\,\B X_i \mid \omega^{(\B z_i)},\,\B \theta^{(\B z_i)}\bigr)} \,\Bigr] + \log\,{p^{(\B z_i)}} \,\Bigr) \no \\
    =& \sum_{i=1}^{N} \sum_{\B z_1 = 1}^{M} \cdots \sum_{\B z_N = 1}^{M}p\bigl(\omega^{(\B z_1)},\;\dotsc\,,\;\omega^{(\B z_N)}\mid\B D_Y,\,\B \Theta(s)\bigr) \no \\
    & \times \Bigl( \,\E_{\B X_i \mid \B Z,\,\B Y_i,\,\B \Theta(s)} \Bigl[ \,\log\,{p\bigl(\B Y_i,\,\B X_i \mid \omega^{(\B z_i)},\,\B \theta^{(\B z_i)}\bigr)} \,\Bigr] + \log\,{p^{(\B z_i)}} \,\Bigr) .
\end{align}
Taking account of the constraint (\ref*{Eq:12}), we obtain
\begin{align}
    \mathcal{Q}\bigl(\B \Theta \mid \B \Theta(s)\bigr)
    =& \sum_{i=1}^{N} \sum_{\B z_1 = 1}^{M}p\bigl(\omega^{(\B z_1)}\mid\B D_Y,\,\B \Theta(s)\bigr) \cdots \sum_{\B z_N = 1}^{M}p\bigl(\omega^{(\B z_N)}\mid\B D_Y,\,\B \Theta(s)\bigr) \no \\
    & \times \,\E_{\B X_i \mid \B Z,\,\B Y_i,\,\B \Theta(s)} \Bigl[ \,\log\,{p\bigl(\B Y_i,\,\B X_i \mid \omega^{(\B z_i)},\,\B \theta^{(\B z_i)}\bigr)} + \log\,{p^{(\B z_i)}} \,\Bigr] \,\no \\
    =& \sum_{i=1}^{N} \sum_{\B z_i = 1}^{M}p\bigl(\omega^{(\B z_i)}\mid \B D_Y,\,\B \Theta(s)\bigr) \Bigl( \,\E_{\B X_i \mid \B Z,\,\B Y_i,\,\B \Theta(s)} \Bigl[ \,\log\,{p\bigl(\B Y_i,\,\B X_i \mid \omega^{(\B z_i)},\,\B \theta^{(\B z_i)}\bigr)} \,\Bigr] + \log\,{p^{(\B z_i)}} \,\Bigr) \no \\
    =& \sum_{i=1}^{N} \sum_{\B z_i = 1}^{M}p\bigl(\omega^{(\B z_i)}\mid \B Y_i,\,\B \Theta(s)\bigr) \Bigl( \,\E_{\B X_i \mid \B Z,\,\B Y_i,\,\B \Theta(s)} \Bigl[ \,\log\,{p\bigl(\B Y_i,\,\B X_i \mid \omega^{(\B z_i)},\,\B \theta^{(\B z_i)}\bigr)} \,\Bigr] + \log\,{p^{(\B z_i)}} \,\Bigr) \no \\
    =& \sum_{i=1}^{N} \sum_{k = 1}^{M} p\bigl(\omega^{(k)}\mid \B Y_i,\,\B \Theta(s)\bigr) \Bigl( \,Q_{i}^{(k)}\bigl(\B \theta^{(k)},\,\B \theta^{(k)} (s)\bigr) + \log\,{p^{(k)}} \,\Bigr).
\end{align}

\newpage

\section{Derivation of M-step of EM algorithm for MLGSSMs}

Equation (\ref*{Eq:20}) can be rewritten as
\begin{align}
    \frac{\partial}{\partial p^{(k)}} \sum_{i=1}^{N} \sum_{k = 1}^{M} p\bigl(\omega^{(k)} \mid \B Y_i,\,\B \Theta(s)\bigr) \log\,{p^{(k)}} - \lambda 
    = \frac{1}{p^{(k)}}\sum_{i=1}^N p\bigl(\omega^{(k)} \mid \B Y_i,\,\B \Theta(s)\bigr) - \lambda
    = 0 ,
\end{align}
that is,
\begin{align}
    \sum_{i = 1}^{N} p\bigl(\omega^{(k)} \mid \B Y_i,\,\B \Theta(s)\bigr) - \lambda p^{(k)} = 0.
\end{align}
Summing this equation for all $k$ yields
\begin{align}
    \sum_{k=1}^M \sum_{i=1}^N p\bigl(\omega^{(k)} \mid \B Y_i,\,\B \Theta(s)\bigr) - \lambda \sum_{k=1}^M p^{(k)} = 0 .
\end{align}
Taking account of the constraint (\ref*{Eq:12}), we obtain
\begin{align}
    \sum_{i=1}^N 1 - \lambda = 0 ,
\end{align}
that is,
\begin{align}
    \lambda = N .
\end{align}
Consequently, we obtain
\begin{align}
    \label{Eq:B6}
    \hat{p}^{(k)} = \frac{1}{N} \sum_{i=1}^N p\bigl(\omega^{(k)} \mid \B Y_i,\,\B \Theta(s)\bigr) .
\end{align}

Recall
\begin{align}
    \tag{\ref*{Eq:21}}
    \frac{\partial}{\partial \B \theta^{(k)}}  \mathcal{Q}\bigl(\B \Theta \mid \B \Theta(s)\bigr)
    = \sum_{i=1}^{N} p\bigl(\omega^{(k)} \mid \B Y_i,\,\B \Theta(s)\bigr) \frac{\partial}{\partial \B \theta^{(k)}} Q_{i}^{(k)}\bigl(\B \theta^{(k)},\,\B \theta^{(k)}(s)\bigr)
    =0 .
\end{align}
In this equation, $Q_{i}^{(k)}$ corresponds to the ECDLL (\ref*{Eq:2}) of a single LGSSM, and therefore can be expressed in the same form as Eq.~(\ref*{Eq:5}). 
By collecting terms containing $\B \mu^{(k)}$ and $\B P^{(k)}$, the equation corresponding to Eq.~(\ref*{Eq:5}) is written as follows:
\begin{align}
    Q_{i}^{(k)}\bigl(\B \theta^{(k)} , \B \theta^{(k)} (s)\bigr)
    =&\, \frac{1}{2}\log{\bigl|{\B P^{(k)}}^{-1}\bigr|} - \frac{1}{2} \Bigl\{ \,\tr{ \Bigl({\B P^{(k)}}^{-1}\E \Bigl[ \,\B x_{i}[1]\T{\B x_{i}[1]} \,\Bigr] \Bigr)} \no \\
    & - \tr{ \Bigl( {\B P^{(k)}}^{-1} \B \mu^{(k)}\E \Bigl[ \,\T{\B x_{i}[1]} \,\Bigr] \Bigr)} - \tr{ \Bigl( {\B P^{(k)}}^{-1} \E \Bigl[ \,\B x_{i}[1] \,\Bigr] \B \mu^{\T{(k)}} \Bigr)} \no \\
    & + \tr{ \Bigl({\B P^{(k)}}^{-1} \B \mu^{(k)}\B \mu^{\T{(k)}} \Bigr)}\,\Bigr\} \no \\
    & + \text{const}.
\end{align}
Substituting this equation into Eq.~(\ref*{Eq:21}) and maximizing $\mathcal{Q}$ with respect to $\B \mu^{(k)}$ and $\B P^{(k)}$ yield
\begin{align}
    \label{Eq:B8}
    \hat{\B \mu}^{(k)}
    = \dfrac{\sum\limits_{i=1}^{N} p\bigl(\omega^{(k)} \mid \B Y_i,\,\B \Theta(s)\bigr)\E \Bigl[ \,\B x_{i}[1] \,\Bigr]}{\sum\limits_{i=1}^{N} p\bigl(\omega^{(k)} \mid \B Y_i,\,\B \Theta(s)\bigr)} ,
\end{align}
and
\begin{align}
    \label{Eq:B9}
    \hat{\B P}^{(k)}
    =\dfrac{\B P^{\B '}}{\sum\limits_{i=1}^{N} p\bigl(\omega^{(k)} \mid \B Y_i,\,\B \Theta(s)\bigr)} ,
\end{align}
where
\begin{align}
    \B P^{\B '}
    = \sum_{i=1}^{N} p\bigl(\omega^{(k)} \mid \B Y_i,\,\B \Theta(s)\bigr)\Bigl( \,\E \Bigl[ \,\B x_{i}[1] \T{\B x_{i}[1]}\, \Bigr]-\hat{\B \mu}^{(k)} \T{\E \Bigl[ \,\B x_{i}[1] \,\Bigr]}
    - \E \Bigl[ \,\B x_{i}[1] \,\Bigr] \hat{\B \mu}^{\T{(k)}} + \hat{\B \mu}^{(k)} \hat{\B \mu}^{\T{(k)}} \,\Bigr) .
\end{align}

By collecting terms containing $\B A^{(k)}$ and $\B \Gamma^{(k)}$, $Q_{i}^{(k)}$ can also be expressed as
\begin{align}
    Q_{i}^{(k)}\bigl(\B \theta^{(k)},\,\B \theta^{(k)} (s)\bigr)
    =& - \frac{T-1}{2}\log{\bigl|\B \Gamma^{(k)}\bigr|} - \frac{1}{2} \sum_{t=2}^{T} \Bigl\{ \,\tr{\Bigl( {\B \Gamma^{(k)}}^{-1} \E \Bigl[ \,\B x_{i}[t] \B x_{i}\T{[t]} \,\Bigr] \Bigr)} \no \\
    & - \tr{\Bigl( {\B \Gamma^{(k)}}^{-1} \B A^{(k)} \E \Bigl[ \, \B x_{i}[t-1]\B x_{i}\T{[t]} \,\Bigr] \Bigr)} - \tr{\Bigl( {\B \Gamma^{(k)}}^{-1} \E \Bigl[ \,\B x_{i}[t]\B x_{i}\T{[t-1]} \,\Bigr]\B A^{\T{(k)}} \Bigr)} \no \\
    & + \tr{\Bigl( {\B \Gamma^{(k)}}^{-1} \B A^{(k)} \E \Bigl[ \, \B x_{i}[t-1]\B x_{i}\T{[t-1]} \,\Bigr] \B A^{\T{(k)}} \Bigr)} \,\Bigr\} \no \\
    & + \text{const} .
\end{align}
Substituting this equation into Eq.~(\ref*{Eq:21}) and maximizing $\mathcal{Q}$ with respect to $\B A^{(k)}$ and $\B \Gamma^{(k)}$ yield
\begin{align}
    \label{Eq:B12}
    \hat{\B A}^{(k)}
    =&\, \Bigl( \,\sum_{i=1}^{N}\sum_{t=2}^{T} p\bigl(\omega^{(k)} \mid \B Y_i,\,\B \Theta(s)\bigr)  \E \Bigl[ \,\B x_{i}[t] \B x_{i}\T{[t-1]} \,\Bigr] \,\Bigr) \no \\
    &\times \Bigl( \,\sum_{i=1}^{N}\sum_{t=2}^{T} p\bigl(\omega^{(k)} \mid \B Y_i,\,\B \Theta(s)\bigr)  \E \Bigl[ \,\B x_{i}[t-1] \B x_{i}\T{[t-1]} \,\Bigr] \,\Bigr)^{-1} ,
\end{align}
and
\begin{align}
    \label{Eq:B13}
    \hat{\B \Gamma}^{(k)}
    = \dfrac{\B \Gamma^{\B '}}{\sum\limits_{i=1}^{N} p\bigl(\omega^{(k)} \mid \B Y_i,\,\B \Theta(s)\bigr)\bigl(T-1\bigr)},
\end{align}
where
\begin{align}
    \B \Gamma^{\B '}
    =&\, \sum_{i=1}^{N}\sum_{t=2}^T p\bigl(\omega^{(k)} \mid \B Y_i,\,\B \Theta(s)\bigr) 
    \Bigl( \,
    \E \Bigl[ \,\B x_{i}[t] \B x_{i}\T{[t]} \,\Bigr] 
    - \hat{\B A}^{(k)} \E \Bigl[ \,\B x_{i}[t-1]\B x_{i}\T{[t]}
    \,\Bigr] \no \\
    & - \E \Bigl[ \,\B x_{i}[t]\B x_{i}\T{[t-1]} \,\Bigr] \hat{\B A}^{\T{(k)}} 
    + \hat{\B A}^{(k)} \E \Bigl[ \,\B x_{i}[t-1]\B x_{i}\T{[t-1]} \,\Bigr] \hat{\B A}^{\T{(k)}}
    \,\Bigr) .
\end{align}

Similarly, by collecting terms containing $\B C^{(k)}$ and $\B \Sigma^{(k)}$ in $Q_{i}^{(k)}$, we obtain
\begin{align}
    Q_{i}^{(k)}\bigl(\B \theta^{(k)} ,\, \B \theta^{(k)} (s)\bigr)
    =& - \frac{T}{2}\log{\bigl|\B \Sigma^{(k)}\bigr|} - \frac{1}{2} \sum_{t=1}^{T} \Bigl\{ \,\tr{\Bigl( {\B \Sigma^{(k)}}^{-1} \B y_{i}[t] \B y_{i}\T{[t]} \Bigr)} \no \\
    & - \tr{\Bigl( {\B \Sigma^{(k)}}^{-1} \B y_{i}[t] \E \Bigl[ \, \B x_{i}\T{[t]} \,\Bigr] \B C^{\T{(k)}} \Bigr)} - \tr{\Bigl( {\B \Sigma^{(k)}}^{-1} \B C^{(k)} \E \Bigl[ \,\B x_{i}[t] \,\Bigr] \B y\T{[t]} \Bigr)} \no \\ 
    & + \tr{\Bigl( {\B \Sigma^{(k)}}^{-1} \B C^{(k)} \E \Bigl[ \, \B x_{i}[t]\B x_{i}\T{[t]} \,\Bigr] \B C^{\T{(k)}} \Bigr)} \,\Bigr\} \no \\
    & + \text{const} .
\end{align}
Substituting this equation into Eq.~(\ref*{Eq:21}) and maximizing $\mathcal{Q}$ with respect to $\B C^{(k)}$ and $\B \Sigma^{(k)}$ yield
\begin{align}
    \label{Eq:B16}
    \hat{\B C}^{(k)} 
    =&\, \Bigl( \,\sum_{i=1}^{N} \sum_{t=1}^{T} p\bigl(\omega^{(k)} \mid \B Y_i,\,\B \Theta(s)\bigr) \E \Bigl[ \,\B x_{i}[t] \,\Bigr] \B y_{i}\T{[t]} \,\Bigr) \no \\
    & \times \Bigl( \,\sum_{i=1}^{N} \sum_{t=1}^{T} p\bigl(\omega^{(k)} \mid \B Y_i,\,\B \Theta(s)\bigr)\E \Bigl[ \,\B x_{i}[t] \B x_{i}\T{[t]} \,\Bigr] \,\Bigr)^{-1} ,
\end{align}
and
\begin{align}
    \label{Eq:B17}
    \hat{\B \Sigma}^{(k)}= \dfrac{\B \Sigma^{'}}{\sum\limits_{i=1}^{N} p\bigl(\omega^{(k)} \mid \B Y_i,\,\B \Theta(s)\bigr)T} ,
\end{align}
where
\begin{align}
    \B \Sigma^{'}
    =& \sum_{i=1}^{N} \sum_{t=1}^{T} p\bigl(\omega^{(k)} \mid \B Y_i,\,\B \Theta(s)\bigr)
    \Bigl( \,
    \B y_{i}[t] \B y_{i}\T{[t]} 
    - \B y_{i}[t] \E \Bigl[ \,\B x_{i}\T{[t]} \,\Bigr] \hat{\B C}^{\T{(k)}} \no \\
    & - \hat{\B C}^{(k)} \E \Bigl[ \,\B x_{i}[t] \,\Bigr]\B y_{i}\T{[t]}
    + \hat{\B C}^{(k)} \E \Bigl[ \,\B x_{i}[t]\B x_{i}\T{[t]} \,\Bigr] \hat{\B C}^{\T{(k)}}
    \,\Bigr) .
\end{align}

\bibliography{bibfile}

\begin{thebibliography}{10}
\expandafter\ifx\csname url\endcsname\relax
  \def\url#1{\texttt{#1}}\fi
\expandafter\ifx\csname urlprefix\endcsname\relax\def\urlprefix{URL }\fi
\expandafter\ifx\csname href\endcsname\relax
  \def\href#1#2{#2} \def\path#1{#1}\fi

\bibitem{Sangeeta2012}
S.~Rani, G.~Sikka, Recent techniques of clustering of time series data: A
  survey, International Journal of Computer Applications 52~(15) (2012) 1--9.
\newblock \href {https://doi.org/10.5120/8282-1278}
  {\path{doi:10.5120/8282-1278}}.

\bibitem{Axel2002}
A.~Wism^^c3^^bcller, O.~Lange, D.~R. Dersch, G.~L. Leinsinger, K.~Hahn,
  B.~P^^c3^^bctz, D.~Auer, Cluster analysis of biomedical image time-series,
  International Journal of Computer Vision 46~(2) (2002) 103--128.
\newblock \href {https://doi.org/10.1023/A:1013550313321}
  {\path{doi:10.1023/A:1013550313321}}.

\bibitem{Kurbalija2012}
V.~Kurbalija, C.~von Bernstorff, H.-D. Burkhard, J.~Nachtwei, M.~Ivanovi\'{c},
  L.~Fodor, Time-series mining in a psychological domain, in: Proceedings of
  the Fifth Balkan Conference in Informatics, 2012, pp. 58--63.
\newblock \href {https://doi.org/10.1145/2371316.2371328}
  {\path{doi:10.1145/2371316.2371328}}.

\bibitem{en6020579}
F.~Iglesias, W.~Kastner, Analysis of similarity measures in times series
  clustering for the discovery of building energy patterns, Energies 6~(2)
  (2013) 579--597.
\newblock \href {https://doi.org/10.3390/en6020579}
  {\path{doi:10.3390/en6020579}}.

\bibitem{Kumar2002}
M.~Kumar, N.~R. Patel, J.~Woo, Clustering seasonality patterns in the presence
  of errors, in: Proceedings of the Eighth ACM SIGKDD International Conference
  on Knowledge Discovery and Data Mining, 2002, pp. 557--563.
\newblock \href {https://doi.org/10.1145/775047.775129}
  {\path{doi:10.1145/775047.775129}}.

\bibitem{WARRENLIAO20051857}
T.~W. Liao, Clustering of time series data---a survey, Pattern Recognition
  38~(11) (2005) 1857--1874.
\newblock \href {https://doi.org/10.1016/j.patcog.2005.01.025}
  {\path{doi:10.1016/j.patcog.2005.01.025}}.

\bibitem{AGHABOZORGI201516}
S.~Aghabozorgi, A.~S. Shirkhorshidi, T.~Y. Wah, Time-series clustering -- {A}
  decade review, Information Systems 53 (2015) 16--38.
\newblock \href {https://doi.org/10.1016/j.is.2015.04.007}
  {\path{doi:10.1016/j.is.2015.04.007}}.

\bibitem{XIONG20041675}
Y.~Xiong, D.-Y. Yeung, Time series clustering with {ARMA} mixtures, Pattern
  Recognition 37~(8) (2004) 1675--1689.
\newblock \href {https://doi.org/10.1016/j.patcog.2003.12.018}
  {\path{doi:10.1016/j.patcog.2003.12.018}}.

\bibitem{4782736}
B.~V. Kini, C.~C. Sekhar, Bayesian mixture of {AR} models for time series
  clustering, Pattern Analysis and Applications 16~(2) (2013) 179--200.
\newblock \href {https://doi.org/10.1007/s10044-011-0247-5}
  {\path{doi:10.1007/s10044-011-0247-5}}.

\bibitem{Cen2000}
C.~Li, G.~Biswas, A {B}ayesian approach to temporal data clustering using
  hidden {M}arkov models, in: Proceedings of the Seventeenth International
  Conference on Machine Learning, 2000, pp. 543--550.

\bibitem{Kitagawa1996}
G.~Kitagawa, W.~Gersch, Smoothness Priors Analysis of Time Series, Springer,
  New York, 1996.
\newblock \href {https://doi.org/10.1007/978-1-4612-0761-0}
  {\path{doi:10.1007/978-1-4612-0761-0}}.

\bibitem{anderson1979}
B.~D.~O. Anderson, J.~B. Moore, Optimal Filtering, Prentice-Hall, New Jersey,
  1979.

\bibitem{mclachlan1988mixture}
G.~J. McLachlan, K.~E. Basford, Mixture Models: Inference and Applications to
  Clustering, Marcel Dekker, New York, 1988.

\bibitem{Dempster1977}
A.~P. Dempster, N.~M. Laird, D.~B. Rubin, Maximum likelihood from incomplete
  data via the {EM} algorithm, Journal of the Royal Statistical Society: Series
  B (Methodological) 39~(1) (1977) 1--22.
\newblock \href {https://doi.org/10.1111/j.2517-6161.1977.tb01600.x}
  {\path{doi:10.1111/j.2517-6161.1977.tb01600.x}}.

\bibitem{Richard1984}
R.~A. Redner, H.~F. Walker, Mixture densities, maximum likelihood and the {EM}
  algorithm, SIAM Review 26~(2) (1984) 195--239.
\newblock \href {https://doi.org/10.1137/1026034} {\path{doi:10.1137/1026034}}.

\bibitem{Agrawal1993}
R.~Agrawal, C.~Faloutsos, A.~Swami, Efficient similarity search in sequence
  databases, in: Foundations of Data Organization and Algorithms, 1993, pp.
  69--84.
\newblock \href {https://doi.org/10.1007/3-540-57301-1_5}
  {\path{doi:10.1007/3-540-57301-1_5}}.

\bibitem{795160}
Z.~R. Struzik, A.~Siebes, Measuring time series' similarity through large
  singular features revealed with wavelet transformation, in: Proceedings of
  the Tenth International Workshop on Database and Expert Systems Applications,
  1999, pp. 162--166.
\newblock \href {https://doi.org/10.1109/DEXA.1999.795160}
  {\path{doi:10.1109/DEXA.1999.795160}}.

\bibitem{Gavrilov}
M.~Gavrilov, D.~Anguelov, P.~Indyk, R.~Motwani, Mining the stock market
  (extended abstract): Which measure is best?, in: Proceedings of the Sixth ACM
  SIGKDD International Conference on Knowledge Discovery and Data Mining, 2000,
  pp. 487--496.
\newblock \href {https://doi.org/10.1145/347090.347189}
  {\path{doi:10.1145/347090.347189}}.

\bibitem{Sebastiani1999}
P.~Sebastiani, M.~Ramoni, P.~Cohen, J.~Warwick, J.~Davis, Discovering dynamics
  using {B}ayesian clustering, in: Advances in Intelligent Data Analysis, 1999,
  pp. 199--209.
\newblock \href {https://doi.org/10.1007/3-540-48412-4_17}
  {\path{doi:10.1007/3-540-48412-4_17}}.

\bibitem{Ramoni2000MultivariateCB}
M.~Ramoni, P.~Sebastiani, P.~R. Cohen, Multivariate clustering by dynamics, in:
  Proceedings of the Seventeenth National Conference on Artificial
  Intelligence, 2000, pp. 633--638.

\bibitem{Panuccio2002}
A.~Panuccio, M.~Bicego, V.~Murino, A hidden {M}arkov model-based approach to
  sequential data clustering, in: Structural, Syntactic, and Statistical
  Pattern Recognition, 2002, pp. 734--743.
\newblock \href {https://doi.org/10.1007/3-540-70659-3_77}
  {\path{doi:10.1007/3-540-70659-3_77}}.

\bibitem{Licen1999}
C.~Li, G.~Biswas, Temporal pattern generation using hidden {M}arkov model based
  unsupervised classification, in: Advances in Intelligent Data Analysis, 1999,
  pp. 245--256.
\newblock \href {https://doi.org/10.1007/3-540-48412-4_21}
  {\path{doi:10.1007/3-540-48412-4_21}}.

\bibitem{Li2002ApplyingTH}
C.~Li, G.~Biswas, Applying the hidden {M}arkov model methodology for
  unsupervised learning of temporal data, International Journal of
  Knowledge-based and Intelligent Engineering Systems 6 (2002) 152--160.

\bibitem{Piccolo1990}
D.~Piccolo, A distance measure for classifying {ARIMA} models, Journal of Time
  Series Analysis 11~(2) (1990) 153--164.
\newblock \href {https://doi.org/10.1111/j.1467-9892.1990.tb00048.x}
  {\path{doi:10.1111/j.1467-9892.1990.tb00048.x}}.

\bibitem{989529}
K.~Kalpakis, D.~Gada, V.~Puttagunta, Distance measures for effective clustering
  of {ARIMA} time-series, in: Proceedings of the 2001 IEEE International
  Conference on Data Mining, 2001, pp. 273--280.
\newblock \href {https://doi.org/10.1109/ICDM.2001.989529}
  {\path{doi:10.1109/ICDM.2001.989529}}.

\bibitem{18626}
L.~Rabiner, A tutorial on hidden {M}arkov models and selected applications in
  speech recognition, Proceedings of the IEEE 77~(2) (1989) 257--286.
\newblock \href {https://doi.org/10.1109/5.18626} {\path{doi:10.1109/5.18626}}.

\bibitem{Bishop2006}
C.~M. Bishop, Pattern Recognition and Machine Learning, Springer, New York,
  2006.

\bibitem{Beal}
M.~J. Beal, Variational algorithms for approximate {B}ayesian inference, Ph.D.
  thesis, University College London (2003).

\bibitem{Tyler2021}
T.~Roick, D.~Karlis, P.~D. McNicholas, Clustering discrete-valued time series,
  Journal of Advances in Data Analysis and Classification 15~(1) (2021)
  209--229.
\newblock \href {https://doi.org/10.1007/s11634-020-00395-7}
  {\path{doi:10.1007/s11634-020-00395-7}}.

\bibitem{LinBachelor}
A.~Lin, Model-based clustering of time series exhibiting nonlinear dynamics,
  Bachelor's thesis, Harvard College (2019).

\bibitem{pmlr-v89-lin19b}
A.~Lin, Y.~Zhang, J.~Heng, S.~A. Allsop, K.~M. Tye, P.~E. Jacob, D.~Ba,
  Clustering time series with nonlinear dynamics: A {B}ayesian non-parametric
  and particle-based approach, in: Proceedings of the Twenty-Second
  International Conference on Artificial Intelligence and Statistics, 2019, pp.
  2476--2484.

\bibitem{4383735}
S.~Chiappa, D.~Barber, Output grouping using {D}irichlet mixtures of linear
  {G}aussian state-space models, in: Proceedings of fifth International
  Symposium on Image and Signal Processing and Analysis, 2007, pp. 446--451.
\newblock \href {https://doi.org/10.1109/ISPA.2007.4383735}
  {\path{doi:10.1109/ISPA.2007.4383735}}.

\bibitem{NIPS2016_7d6044e9}
M.~J. Johnson, D.~K. Duvenaud, A.~Wiltschko, R.~P. Adams, S.~R. Datta,
  Composing graphical models with neural networks for structured
  representations and fast inference, in: Proceedings of the thirtieth
  International Conference on Neural Information Processing Systems, 2016, pp.
  2954--2962.

\bibitem{Ghahramani1996}
Z.~Ghahramani, G.~E. Hinton, Parameter estimation for linear dynamical systems,
  Technical Report CRG-TR-92-2, University of Toronto (1996).

\bibitem{Shumway1982}
R.~H. Shumway, D.~S. Stoffer, An approach to time series smoothing and
  forecasting using the {EM} algorithm, Journal of Time Series Analysis 3~(4)
  (1982) 253--264.
\newblock \href {https://doi.org/10.1111/j.1467-9892.1982.tb00349.x}
  {\path{doi:10.1111/j.1467-9892.1982.tb00349.x}}.

\bibitem{James1994}
J.~D. Hamilton, Time Series Analysis, Princeton University Press, New Jersey,
  1994.
\newblock \href {https://doi.org/10.2307/j.ctv14jx6sm}
  {\path{doi:10.2307/j.ctv14jx6sm}}.

\bibitem{Chen1998}
C.-T. Chen, Linear System Theory and Design, Oxford University Press, New York,
  2014.

\bibitem{Kitagawa1987}
G.~Kitagawa, Non-{G}aussian state-space modeling of nonstationary time series,
  Journal of the American Statistical Association 82~(400) (1987) 1032--1041.
\newblock \href {https://doi.org/10.2307/2289375} {\path{doi:10.2307/2289375}}.

\bibitem{TANIZAKI1998263}
H.~Tanizaki, R.~S. Mariano, Nonlinear and non-{G}aussian state-space modeling
  with {M}onte {C}arlo simulations, Journal of Econometrics 83~(1) (1998)
  263--290.
\newblock \href {https://doi.org/10.1016/S0304-4076(97)80226-6}
  {\path{doi:10.1016/S0304-4076(97)80226-6}}.

\end{thebibliography}

\end{document}